\pdfoutput=1
\newcommand{\arxivversion}{}
\documentclass[11pt]{article}
\ifdefined\arxivversion
  \usepackage[preprint]{acl}
\else
  \usepackage[review]{acl}
\fi
\usepackage{times}
\usepackage{latexsym}
\usepackage[T1]{fontenc}
\usepackage[utf8]{inputenc}
\usepackage{microtype}
\usepackage{inconsolata}
\usepackage{graphicx}
\usepackage{booktabs}
\usepackage{tabularx}
\usepackage{xurl}
\usepackage{algorithm}
\usepackage{algorithmic}

\usepackage{amsmath,amssymb,amsthm}
\usepackage[capitalize,noabbrev]{cleveref}
\usepackage{xcolor}
\usepackage{colortbl}
\usepackage{pifont}
\newcommand{\tabok}{\textcolor[HTML]{2E6B55}{\ding{51}}}\newcommand{\tabno}{\textcolor{specred}{\ding{55}}}\newcommand{\tabopen}{\textbf{?}}\newcommand{\tabkeep}{$\bullet$}
\usepackage{tikz}
\usepackage{adjustbox}
\DeclareRobustCommand{\charbadge}[1]{\tikz[baseline=(cb.base)]\node[rounded corners=1.5pt,fill=coral!30,text=black!85,font=\sffamily\bfseries\scriptsize,inner sep=1.6pt,minimum width=1.05em](cb){#1};}
\definecolor{draftred}{HTML}{9C2F24}
\definecolor{specred}{HTML}{B3261E}\definecolor{specteal}{HTML}{00796B}
\definecolor{draftbg}{HTML}{FFF1ED}
\definecolor{hdrgray}{HTML}{C9C9C9}
\definecolor{mint}{HTML}{BFDFD2}
\definecolor{coral}{HTML}{ED8D5A}

\newtheorem{definition}{Definition}[section]

\newtheorem{proposition}[definition]{Proposition}
\newtheorem{lemma}[definition]{Lemma}
\newtheorem{corollary}[definition]{Corollary}
\theoremstyle{remark}
\newtheorem{remark}[definition]{Remark}
\newcommand{\lossset}{\mathbb L}
\newcolumntype{L}{>{\raggedright\arraybackslash}X}
\newcolumntype{P}[1]{>{\raggedright\arraybackslash}p{#1}}
\newcommand{\Law}{\mathcal L}
\newcommand{\riskmodel}{\mathfrak R}
\newcommand{\interventions}{\mathcal C}

\newcommand{\agentrisk}{\texorpdfstring{\mbox{\fontfamily{qtm}\selectfont\scshape\textls[60]{agent-risk}}}{agent-risk}}

\theoremstyle{plain}
\newtheorem*{restatementinner}{\restatementname}
\newcommand{\restatementname}{}

\newenvironment{informal}[1]{\par\smallskip\noindent\textbf{#1.}\ \ignorespaces}{\par\smallskip}
\makeatletter\setboolean{ALC@noend}{true}
\newcommand{\alglinelabel}[1]{\protected@edef\@currentlabel{\arabic{ALC@line}}\label{#1}}\makeatother

\begin{document}
\title{\agentrisk: Risk Assessment for Changes to AI Agents}
\ifdefined\arxivversion
  \author{
    Yue Zhao$^{1}$ \quad Zixiang Xu$^{1}$ \quad Angela Huyue Zhang$^{1}$ \quad S. Alex Yang$^{2}$ \quad Xiyang Hu$^{3}$ \\[0.4em]
    $^{1}$University of Southern California \\
    $^{2}$London Business School \quad $^{3}$Arizona State University \\[0.3em]
    \texttt{yue.z@usc.edu} \quad \texttt{zixiangx@usc.edu} \quad \texttt{azhang@law.usc.edu} \\
    \texttt{sayang@london.edu} \quad \texttt{xiyanghu@asu.edu}
  }
\else
  \author{Anonymous}
\fi
\maketitle

\begin{abstract}
Before changing a deployed AI agent (a stricter safety gate, a new model), organizations must approve the change, starting from existing logs. Extending classical risk analysis to partial records and partial changes to running systems, \agentrisk{} replays logged episodes, assuming nothing else changes and the agent does not anticipate the change. The log settles episodes acting as before, a declared stop value settles stopped ones, and the rest need logged support, a saved-state rerun, or a bound. We prove \textit{what logs must record for exact replay and which saved states serve a gate family}. For independent episodes, we bound the reruns a decision needs and tolerable drift from live behavior. In a bash sandbox, a 7,200-episode log alone decided neither of two retry gates, which let the agent act after refusals the log never shows. Rerunning only each gate's blocked episodes from saved states reached both live decisions at half the agent calls of a live test sized post hoc for them, given the log. Without per-step harm, a stopping gate's certified harm ceiling rose up to 2.5-fold, since the log hides whether harm preceded the stop. Approval thus needs evidence designed in advance: log fields, saved states, targeted reruns.

\end{abstract}

\section{Introduction}
\label{sec:introduction}

\begin{figure*}[t]
\centering
\includegraphics[width=\textwidth]{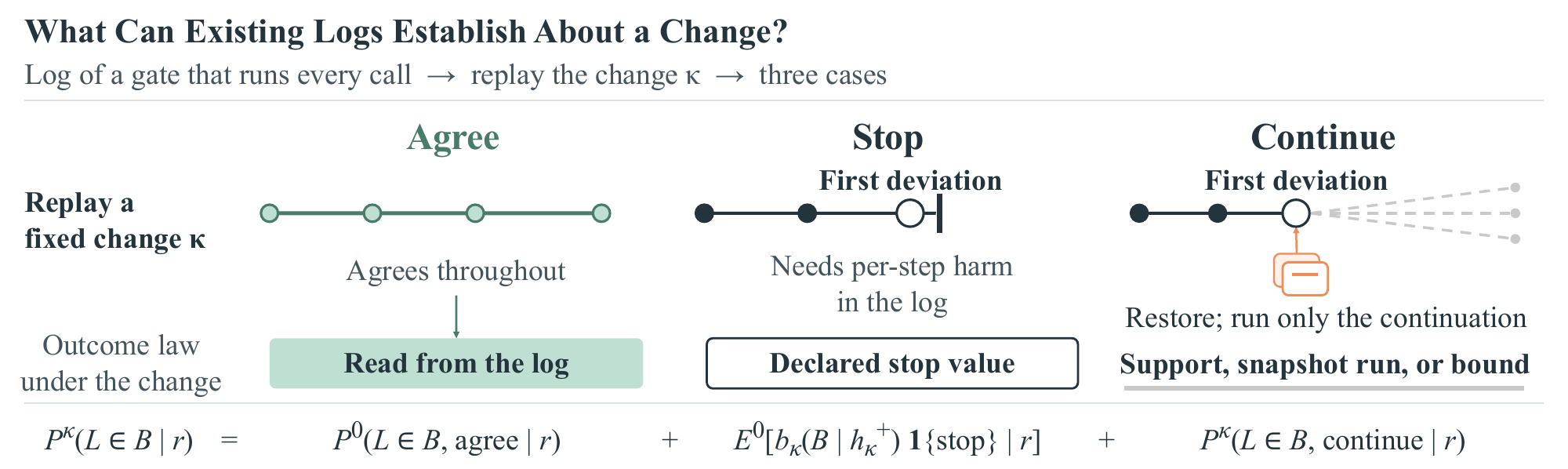}
\caption{Replaying a logged episode under a proposed change. With sufficient records, the log gives the agreeing case and a declared stop value the stopped case; a snapshot run, logged support, or bounds the continuing case.}
\label{fig:overview}
\end{figure*}

AI agents now act for organizations, answering customers and writing and running code \citep{elevenlabs2026aiuc,cursor2026aiuc}. The institutions that absorb the risk of these actions (insurers, certifiers, deployment approvers, and courts that allocate liability) need a stated loss, evidence, and uncertainty before the agent acts. Insurability and premium follow technical due diligence \citep{munichre2023derisking}, whereas one provider separated development from production databases only after a coding agent deleted production data \citep{masad2025replit}. We call the organization that runs an agent and holds its logs the \emph{deployer}. Take a coding agent whose tool calls pass a \emph{monitor} (a second model scoring each call's risk) and a \emph{gate}, which uses the score to run the call, stop the episode, or refuse the call so the agent retries. The deployer keeps changing such an agent, tightening the gate or releasing a new model. Before a change runs, the deployer, or an insurer or auditor with its logs, faces one question: how much of its risk do existing logs establish (Figure~\ref{fig:overview})? Logs (per-step records of tool calls, monitor scores, and outcomes) come first because they are already in hand: a live trial exposes users to the risk under assessment, and a new test campaign costs runs and may not match deployment.

Six characteristics (\S \ref{sec:characteristics}) make this question hard for AI (e.g., LLM) agents (Table~\ref{tab:characteristics}): \charbadge{1}~approval gates, \charbadge{2}~partial records, \charbadge{3}~long horizons, \charbadge{4}~model releases, \charbadge{5}~shared state, and \charbadge{6}~oversight cues, signs that the agent is being tested.

\paragraph{Current Gaps.} Benchmarks and control evaluations run and score a given system \citep{yao2025taubench,kwa2025measuring,greenblatt2024aicontrol}. Some also reuse logged monitor scores to set audit thresholds and resample actions \citep{bhatt2025ctrlz,controlarena2025}; we state when such replay is valid. Off-policy evaluation \citep{horvitz1952generalization,precup2000eligibility}, optimal stopping \citep{goel2017sample}, and partial identification \citep{manski2003partial} ask what logs of the current system imply about a changed one, but a stricter gate's stops never occur in the log, and a retry gate lets the agent act after a refusal no log contains. We specialize these tools to agent changes and add \emph{evidence design}: which logged fields, saved states, and new runs certify a decision about a change.

Our framework, \agentrisk{}, replays each logged episode \citep{li2011unbiased}: on the recorded inputs, without executing new environment steps, it recomputes the new rule's actions up to the \emph{first deviation}, where the two rules first differ, and sorts it into the three cases of Figure~\ref{fig:overview}; a continuing case may need new runs from a saved sandbox state (a \emph{snapshot}).

The key contributions are:
\begin{list}{\textbullet}{\setlength{\leftmargin}{0pt}\settowidth{\labelwidth}{\textbullet}\setlength{\labelsep}{0.4em}\setlength{\itemindent}{\dimexpr\labelwidth+\labelsep\relax}}
\item \textbf{Problem Formulation (Section~\ref{sec:framework}):} We formalize the risk of an agent change before it runs as a specification that extends scenario risk analysis \citep{kaplan1981risk} with partial records and admissible changes, guided by six agent characteristics (Table~\ref{tab:characteristics}), and ask which risk values the logs leave possible (partial identification).
\item \textbf{Theoretical Results (Section~\ref{sec:results}):} We prove (i) exactly what a log must record to settle agreeing and stopped episodes for any current system; (ii) for independent episodes, a lower bound on the reruns needed to certify a harm level, tests that approach it asymptotically, and the restoration error beyond which no valid test certifies reliably; and (iii) the fewest saved states, among those where a gate first acts, serving a whole threshold family.
\item \textbf{Empirical Verification (Section~\ref{sec:validation}):} To test these results against paired live runs in a bash sandbox, we reran only what one log left open (\emph{targeted execution}), which predicted two retry gates' harm within the registered margin of 0.015 and reached the live decision on both, which the log alone could not. Clean starts in a shared workspace and an announced gate mispredicted live runs (Table~\ref{tab:statements}). In all, 16 runs log about 49,000 episodes.
\item \textbf{Reproducibility and Access:} We release a Python package for replay, harm bounds, and logging comparisons, and scripts that check the formal results on finite instances (Appendix~\ref{app:checks}); \ifdefined\arxivversion the harness, logs, and registrations are available from \url{https://github.com/yzhao062/agent-risk}.\else the harness, logs, and registrations follow, anonymized.\fi
\end{list}

\begin{table}[t]
\centering
\footnotesize
\setlength{\tabcolsep}{2.5pt}
\renewcommand{\arraystretch}{1.05}
\begin{tabularx}{\columnwidth}{@{}P{0.36\columnwidth}L@{}}
\rowcolor{hdrgray}
Characteristic (where it enters) & What the logs can establish \\
\charbadge{1} \textbf{Approval gates} (execution) & Stricter stopping gates: replay plus a declared stop value (\S\ref{sec:replay}). Retry gates: similar logged cases, reruns, or bounds (\S\ref{sec:supported}) \\
\charbadge{2} \textbf{Partial records} (records) & Replay is exact only if the log keeps each episode's contribution, e.g., harm timing relative to a stop (\S\ref{sec:logging}) \\
\charbadge{3} \textbf{Long horizons} (execution) & Stopping changes replay at any length; for continuing ones, the share that agrees can stay the same or fall (\S\ref{sec:horizons}) \\
\charbadge{4} \textbf{Model releases} (admissible changes) & A new model rerun on logged inputs and seeds settles episodes where it acts as logged; the rest need similar cases, reruns, or bounds (\S\ref{sec:supported}) \\
\charbadge{5} \textbf{Shared state} (execution) & With the shared state's law known, single-agent tests bound joint risk; reruns from saved shared state settle it (\S\ref{sec:joint}) \\
\charbadge{6} \textbf{Oversight cues} (execution) & Results carry over only for cues seen in testing; an agent told of the gate can act differently, breaking (A2) (\S\ref{sec:cues}) \\
\end{tabularx}
\caption{\textbf{Six organizing agent characteristics and what logs can establish under each.} Characteristic $k$ enters the part of the risk specification (Section~\ref{sec:objects}) in parentheses; its result is in Section~\ref{sec:results}.$k$ and demonstration in Section~\ref{sec:validation}.$k$.}
\label{tab:characteristics}
\end{table}

\section{Problem Formulation}
\label{sec:framework}

\subsection{Six Organizing Characteristics of Agents}
\label{sec:characteristics}

Six characteristics make a change's risk hard to read from logs; each adds a part or an assumption to the risk specification of Section~\ref{sec:objects} (Table~\ref{tab:characteristics}). Approval gates \charbadge{1} are the running case: a stricter gate would stop episodes that the log shows running on, and no log shows what such a stop leads to. Partial records \charbadge{2} keep only the fields a log's designer chose \citep{chan2024visibility}, and long horizons \charbadge{3} add steps at which a changed rule can depart from the log \citep{kwa2025measuring}. A model release \charbadge{4} changes the agent itself, shared state \charbadge{5} links agents through common memory, budgets, or tools, and oversight cues \charbadge{6} let models tell a test from deployment \citep{needham2025evaluation}. We chose the six from a review of 85 sources (Appendices~\ref{app:survey} and~\ref{app:characteristics}).

\subsection{One Agent and One Logged Episode}
\label{sec:running}

We follow one agent and one logged \emph{episode} (one recorded run). A sandboxed agent proposes a tool call $u_t$ at step $t$; a trusted monitor scores it $m_t$ (higher means riskier), and the gate acts on it. The current gate runs every call (\emph{pass-all}) and is absent from the prompt. Episodes last at most $T$ steps, with loss (task failure, harm, cost); harm is an executed policy violation; cost counts tokens and calls.

A stopping gate $\kappa_\theta$ ends the episode at the first $m_t>\theta$; a retry gate refuses such calls and lets the agent try again. A stop ends the episode as a failure, keeps harm and cost accrued so far, and adds no later harm; the deployer declares this valuation.

\noindent\fcolorbox{hdrgray}{black!3}{\parbox{\dimexpr\columnwidth-2\fboxsep-2\fboxrule\relax}{\textbf{Example (a toy episode; made-up scores).} A logged episode has four steps with monitor scores $0.10$, $0.20$, $0.97$, and $0.30$. A gate with $\theta=0.94$ would stop it at step~2, its first deviation from the log. Up to that step, the log shows what the new gate does; three questions remain.\par\smallskip
\hangindent=1.2em\hangafter=1 (i)~\emph{What does the stop cost?} No log contains a stop, so the deployer declares it.\par
\hangindent=1.2em\hangafter=1 (ii)~\emph{Did the stop prevent the harm?} A log keeping only whether harm occurred cannot tell: harm at step~1 precedes the stop, while harm at step~3 would be prevented.\par
\hangindent=1.2em\hangafter=1 (iii)~\emph{What if the gate let the agent retry?} The agent would act after a refusal that no log contains, so logs alone can only bound what follows.}}

\subsection{Writing Down the Risk Specification}
\label{sec:objects}

\begin{center}
\fcolorbox{hdrgray}{white}{\begin{tabular}{@{}r@{\hspace{0.6em}}l@{}}
\multicolumn{2}{@{}c@{}}{$\riskmodel=(E,\ell,\textcolor{specred}{K},\textcolor{specteal}{\interventions})$}\\[1pt]
\footnotesize $E,\ell$ & \footnotesize $\{\langle s_i,p_i,x_i\rangle\}_{i=1}^{N}$ \citep{kaplan1981risk}\\
\footnotesize\textcolor{specred}{$K$} & \footnotesize\textcolor{specred}{partial observation} \citep{bauerle2017partially}\\
\footnotesize\textcolor{specteal}{$\interventions$} & \footnotesize\textcolor{specteal}{interventions} \citep{pearl1995causal}
\end{tabular}}
\end{center}
The colored $K$ and $\interventions$ extend scenario triplets with partial observation and intervention (Appendices~\ref{app:formal-spec} and~\ref{app:symbols}; Table~\ref{tab:foundations}); badges mark where each characteristic enters.

\emph{Execution} $E$ (what runs) names kernels, random maps from declared inputs to outputs, over at most $T$ steps \charbadge{3}: the agent, monitor, gate \charbadge{1}, sandbox, shared state \charbadge{5}, and oversight cues \charbadge{6}. Serving configuration is a declared input, since batching and the server process can change outputs (Section~\ref{sec:demo-release}). \emph{Consequences} $\ell$ (what counts as loss) map a complete episode to loss $L$, here (failure, harm, cost), keeping pre-stop harm \charbadge{1}. \emph{Records} $K$ (what the log keeps) specify what the evaluator logs from each step's available history \charbadge{2}, possibly differing from the agent's context. Write $P^0$ for the current system's law on the finite path space. \emph{Admissible changes} $\interventions$ (what may change) specify which kernels may be replaced, when, using what information, and by whom, such as model releases \charbadge{4} and agent connections \charbadge{5}.

\subsection{A Change's Risk and Identified Set}
\label{sec:question}
\label{sec:records}

With the deployment written down, we can ask what the logs justify about one change. Let $R_t=r$ be the evaluator's record up to step $t$ when $\kappa\in\interventions$ takes effect, and $X=\phi(L)$ the queried loss coordinate. Models that produce the same logs can disagree about the changed system, as each gap in the toy episode shows (Appendix~\ref{app:diagnostic}), so the deployer declares a family $\mathcal M_0$ of structural models; their kernels and records can vary, except where the specification fixes those kernels or their inputs. The deployer then asks which risks remain possible under the models that reproduce the observed record law of every evidence source, $P^R_M=P^R_{\mathrm{obs}}$ \citep{manski2003partial}. The risk of the change, and the values the records leave for it, are
\begin{equation}
\begin{aligned}
q^\kappa_t(r)&=\rho\bigl(\Law_{P^\kappa}(X\mid R_t=r)\bigr),\\
\Theta^\kappa_t(r)&=\bigl\{q^\kappa_{t,M}(r):\ M\in\mathcal M_0,\ P^R_M=P^R_{\mathrm{obs}}\bigr\}.
\end{aligned}
\label{eq:risk}
\end{equation}
Here $P^\kappa$ is the specified law of the changed system, and $q^\kappa_{t,M}$ the risk under $M$ (Appendix~\ref{app:formal-question}, which also covers records of positive probability and endpoint confidence bounds). The functional $\rho$ summarizes $X$'s law as harm probability, mean cost, or conditional value-at-risk (CVaR) of tokens \citep{rockafellar2002cvar}. The identified set $\Theta^\kappa_t(r)$ is what the records justify before the change runs, and so what a deployer or auditor can claim. The risk is identified when the set has one value; otherwise we bound it sharply or from outside.

Logs describe the current system, so they speak about the changed one only under two assumptions that need justification beyond the records (exact: Appendix~\ref{app:formal-question}). Without (A1, invariance), an undeclared change to a tool, the serving stack, or a model behind a fixed name makes logged steps unrepresentative. Without (A2, no anticipation), an agent told about the gate may act differently before the gate acts. If either fails, the deployer must model the extra change.

\begin{informal}{(A1) Invariance, informal}
Nothing the change leaves in place changes: the initial law and every kernel the change does not replace, random inputs included, stay as logged. Installing the gate thus leaves the model, tools, users, and sampler as logged. This covers unobserved environment state, whose initial law and transitions are part of $E$.
\end{informal}

\begin{informal}{(A2) No anticipation, informal}
Both systems run in the same causal order and coincide before the change. Every kernel left in place keeps its declared inputs. A replaced kernel may read other earlier inputs that the change declares, and for replay both rules read one common input that holds what either uses. Nothing left in place, and no declared input, reads which change was installed or its announcement, so announcing the gate in the agent's prompt violates (A2).
\end{informal}

\section{Evidence Design for Agent Changes}
\label{sec:results}
\label{sec:decisions}

A deployer who proposes a change needs a statement of its risk or the evidence that would settle it. Five steps give one or the other (in symbols: Algorithm~\ref{alg:main}, Appendix~\ref{app:procedure}):
\begin{list}{\arabic{enumi}.}{\usecounter{enumi}\setlength{\leftmargin}{1.1em}\setlength{\labelwidth}{0.8em}\setlength{\labelsep}{0.3em}}
\item \emph{Declare} the change, its stop value, and the decision rule; justify (A1) and (A2), or model the extra change.
\item \emph{Replay} both rules on recorded inputs, where the records permit, up to the first deviation (toy episode: step~2).
\item \emph{Read} agreement and stop contributions when the log is sufficient; the stop value keeps harm and cost accrued before the stop (\S\ref{sec:replay} to \S\ref{sec:horizons}; toy questions (i) and (ii)).
\item \emph{Resolve} the rest by matching logged cases, snapshot runs, or bounds (\S\ref{sec:supported}; question (iii)); other settings need a transfer link (\S\ref{sec:joint}, \S\ref{sec:cues}).
\item \emph{Decide} by jointly valid one-sided bounds, with an allowance for restoration error, against a declared harm ceiling and success floor, or name the evidence needed to settle it (\S\ref{sec:demo-retry}).
\end{list}
To make steps 2 and 3 precise, fix $\kappa,r$ from \eqref{eq:risk} (Appendix~\ref{app:formal-replay}). Baseline paths $\Omega^0$ follow the logged rule. The first deviation $\sigma_\kappa$ is the first step where the new rule differs on logged input; $\mathsf A_\kappa=\{\sigma_\kappa>T\}$ means no deviation. The replayed prefix $h^+_\kappa$ comprises the full history before that decision, including unrecorded coordinates, followed by the new output. It is terminal if the output stops the episode and continuing otherwise.

\begin{informal}{(R) Replay interface, informal}
Both rules are known and deterministic given each step's declared input $I_t$, possibly including a seed. The full record contains that input, so both rules can be rerun on logged proposals, scores, and seeds.
\end{informal}

\begin{informal}{(A3) Declared stop value, informal}
After a stop, the loss follows a law $b_\kappa(\cdot\mid h^+_\kappa)$ that the deployer declares, including losses accrued before the stop. It describes what happens after the stop, so the loss map is unchanged. Changes that stop with the same replayed prefix share it, and it must agree with any post-stop behavior baseline operation can show.
\end{informal}

The three events are agree ($\mathsf A_\kappa$), stop ($\{\sigma_\kappa\le T,\ \text{terminal}\}$), and continue ($\{\sigma_\kappa\le T,\ \text{continuing}\}$), the three columns of Figure~\ref{fig:overview}, whose bottom row is display~\eqref{eq:split} with matching labels; only the last needs evidence beyond the log and the stop value. Here $E^0[\,\cdot\mid r]$ averages under $P^0(\tau\mid r)$, with $h^+_\kappa=h^+_\kappa(\tau)$. Under \textup{(R)} and \textup{(A1)} to \textup{(A3)}, for every $B\subseteq\lossset$,
\begin{equation}
\fcolorbox{hdrgray}{white}{$\begin{aligned}
&P^\kappa(L\in B\mid r)\\
&\quad=\underbrace{P^0(L\in B,\ \text{agree}\mid r)}_{\text{\setlength{\fboxsep}{0.8pt}\colorbox{mint}{agree: read from the log}}}\\
&\quad\phantom{=}+\underbrace{E^0\bigl[b_\kappa(B\mid h^+_\kappa)\,\mathbf 1\{\text{stop}\}\bigm| r\bigr]}_{\text{\setlength{\fboxsep}{0.8pt}\fcolorbox{black!60}{white}{stop: declared stop value}}}\\
&\quad\phantom{=}+\underbrace{P^\kappa(L\in B,\ \text{continue}\mid r)}_{\text{continue: support, rerun, or bounds}}.
\end{aligned}$}
\label{eq:split}
\end{equation}
The first two terms restrict the deterministic-regime g-formula \citep{robins1986new} to paths that follow the new rule, as censoring at deviation does \citep{cain2010start}. The third is identified under the support and recording conditions of Section~\ref{sec:supported}; otherwise, worst-case bounds \citep{manski2003partial} give each continuing episode its least and greatest feasible outcome, attained when unseen continuations can be chosen freely and jointly (condition (S), Proposition~\ref{prop:replay-cont}; coarser records: Algorithm~\ref{alg:main}).

\subsection{Approval Gates: Can Logs Replay a Stop?}
\label{sec:replay}

\begin{informal}{Deviation replay \citep{robins1986new,cain2010start}, informal; exact: Proposition~\ref{prop:replay}}
Assume (R), (A1), and (A2), and condition on the record $r$. The agreement term of \eqref{eq:split} then holds: the changed system has the logged loss law on episodes that agree at every step. With (A3), the stop term holds even if the baseline never stops.
\end{informal}

\textbf{For the deployer.} Log the most permissive safe gate, declare the stop value (tolerated post-stop harm: Appendix~\ref{app:demo-gates}), and replay stricter gates; a looser gate evaluated from a stricter log is a continuing change that needs support or bounds.

\subsection{Partial Records: What Must Logs Keep?}
\label{sec:logging}

\begin{figure}[t]
\centering
\includegraphics[width=\columnwidth]{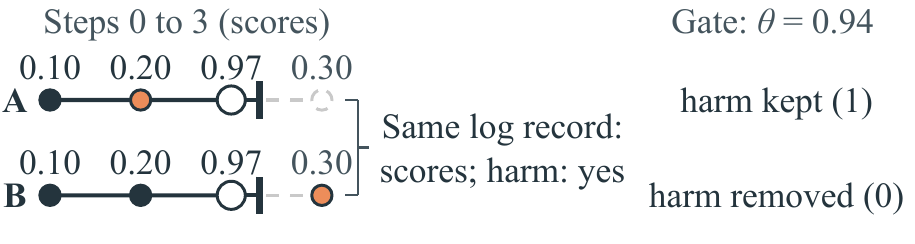}
\caption{\textbf{Missing harm timing leaves stop risk unresolved.} Identical records hide whether executed harm (coral) came before (A) or after (B) the $\theta=0.94$ stop at step~2; the gate keeps A's harm and removes B's.}
\label{fig:timing}
\end{figure}

Section~\ref{sec:replay} uses records sufficient for agreement and stops; with partial records \charbadge{2}, Figure~\ref{fig:timing} shows why missing harm timing leaves stop risk unresolved even with more episodes.

For every baseline law, a log determines the sum of the first two terms of~\eqref{eq:split} exactly when it preserves each episode's combined contribution. Within the conditioning event, a log may merge two episodes only if they contribute equally, which those of Figure~\ref{fig:timing} do not. Contributions are logged loss on agreement, declared loss law on stops, and zero on continuation. In Proposition~\ref{thm:logging}, $u_\kappa(\tau)$ indicates a continuing deviation; its conditional expectation is unresolved mass. Here $g_\#P$ is the log's law under $P$, and $g^{-1}(y)$ the paths producing log value $y$ (scope of necessity: Appendix~\ref{app:logging-scope}).

\begin{proposition}[Logging sufficiency]\label{thm:logging}
Assume \textup{(R)}, \textup{(A1)}, and \textup{(A2)}, with \textup{(R)} referring to the full record; the logging map $g$ below need not record $I_t$.
Let $C_\kappa(\tau)$ be $\delta_{\ell(\tau)}$ on agreement, $b_\kappa(\cdot\mid h^+_\kappa(\tau))$ at a terminal deviation, and $0$ at a continuing one, and let $u_\kappa=1-C_\kappa(\lossset)$. Let $\mathcal P_r$ be the laws on $\Omega^0$ with $P(R_t=r)>0$, and let $s_{\kappa,r}(P)=(\int C_\kappa\,dP(\cdot\mid r),\int u_\kappa\,dP(\cdot\mid r))$. For a logging map $g$ that determines the event $\{R_t=r\}$, there is a map $\Psi$ with $s_{\kappa,r}(P)=\Psi(g_\#P)$ for every $P\in\mathcal P_r$ if and only if $(C_\kappa,u_\kappa)$ is constant on every set $g^{-1}(y)\cap\Omega^0\cap\{R_t=r\}$.
\end{proposition}

\textbf{For the deployer: what to log.} Keep fields that reconstruct each episode's contribution, including harm timing relative to deviation (Table~\ref{tab:log-fields}). Otherwise, distinct baseline distributions with identical records can yield different combined replay contributions, which neither more episodes nor coarser logs resolve (Corollary~\ref{cor:garbling}; \citealp{blackwell1951comparison,blackwell1953equivalent}). Log the serving mode: batch-dependent serving fails (R) when a change reruns the agent.

\subsection{Long Horizons: Does Length Matter?}
\label{sec:horizons}

Each extra step is another chance to leave the log, so the share agreeing throughout cannot rise with the horizon (Corollary~\ref{cor:horizon}). With sufficient records and (A3, declared stop value), stopping changes stay identified at every horizon; a continuing change's worst-case mean interval (Proposition~\ref{prop:replay-cont}) can widen as agreement falls (\S\ref{sec:demo-horizon}).

\subsection{Continuing Changes: Runs and Snapshots}
\label{sec:supported}

Retry gates \charbadge{1} and model releases \charbadge{4} continue past the first deviation into the continuing term of~\eqref{eq:split}, where the agent may read inputs the log never shows. The log suffices when it covers and records every input the changed system reads (supported continuation); otherwise the deployer reruns or bounds, and saves states for the reruns.

\begin{informal}{Supported continuation, informal; exact: Proposition~\ref{prop:supported}}
Assume (A1) and (A2), and let the change replace kernels by known ones after the record $r$ was taken. Suppose (G1) baseline support: at each position, each input the change leads an unchanged kernel to read has positive baseline probability there; inputs where the specification fixes that kernel's output, as after a stop, are exempt. Suppose also that the log records (G2) those kernels' declared inputs, (G3) every coordinate the loss reads, and (G4) every input the change reads. Then the g-formula identifies the changed system's loss law, within mechanisms that share the declared kernel inputs and the values the specification fixes, even if no episode agrees.
\par\smallskip\noindent Without (G1), keeping (G2) to (G4) and every response law the specification fixes, allow any response law at each remaining input of zero baseline probability (U, free responses). Choosing these laws separately for each position gives the identified set under the g-formula; a law shared across positions, or an input merely absent from a finite log, gives an outer set.
\end{informal}

Both parts are classical \citep{robins1986new,manski2003partial}; Appendix~\ref{app:r4-unsupported} unifies them. Because baseline never refuses, a retry gate fails (G1) if an unchanged kernel without a declared value reads the refusal or its message. Under (R), a seed-addressable model release (deterministic at each logged decision input, random tape on fixed serving) has identified agreement mass, at most the first-decision agreement (Corollary~\ref{cor:model-update}).

\textbf{When and how much to rerun.} Where the log has no matching cases, as after a refusal or a new model's new text, the deployer can bound the outcome or use \emph{targeted execution}: restore the sandbox at a logged prefix and run only the continuation. Assumption (T), \emph{restoration fidelity}, links these runs to $\mathcal M_0$: restored state, agent context, and remaining budgets give the continuation its live law after that prefix. A restored law within $\gamma$ in total variation of the live one at each continuing prefix shifts P(harm) by at most $\gamma$ times the continuing mass (Lemma~\ref{lem:restore}). A uniform random sample of unresolved continuations can replace executing all of them (Proposition~\ref{prop:budget}). Live testing spreads the gap between true harm and the target over all episodes; execution concentrates it on the open ones, and restoration error eats into it.

\begin{informal}{Execution budget: limits and achievability, informal; exact: Propositions~\ref{prop:budget-limits} and~\ref{prop:budget-achievability}, Corollary~\ref{cor:cost-ratio}}
Assume (S), binary harm, independent episodes, and one harm probability $p$ for every unresolved continuation (Appendix~\ref{app:budget-strata} allows strata). Treat the log's resolved harmful mass $q_0$ and unresolved mass $m$ as known. The change's P(harm) $q=q_0+mp$ falls short of a target $q^\star<q_0+m$ by $\zeta=q^\star-q>0$. Restored runs err by at most $\gamma$ in total variation, and $\mathrm{kl}$ is Bernoulli relative entropy. (i) If $m\gamma\ge\zeta$, restoration error can hide the whole gap: no procedure that executes logged continuations has power above its size, that is, none certifies a safe change at exact restoration more often than it can certify an unsafe one. (ii) Otherwise, let $\bar p=p+\zeta/m-\gamma$. A procedure valid at level $\alpha$ that certifies with probability at least $1-\beta$ at exact restoration executes at least $\mathrm{kl}(1-\beta,\alpha)/\mathrm{kl}(p,\bar p)$ continuations in expectation \citep{kaufmann2016complexity}. Live testing that does not trust the log needs at least $\mathrm{kl}(1-\beta,\alpha)/\mathrm{kl}(q,q^\star)$ episodes. As $\alpha,\beta\to0$ with a log that grows fast enough, sequential probability ratio tests \citep{wald1945sequential} approach both bounds.
\end{informal}

Here $\bar p$ is the open episodes' harm rate that would just reach the target, less the restoration error; the closer $p$ is to $\bar p$, the more runs a decision needs. Appendix~\ref{app:budget-strata} extends both bounds to strata with their own harm rates, errors, and costs. Strata where restoration error can make harm certain get no runs, and the others' run shares are part of the test's validity (Propositions~\ref{prop:strata-limits} and~\ref{prop:strata-achieve}; Remark~\ref{rem:strata-scope}; allocation: \citealp{neyman1934two}).

\begin{informal}{What to store: snapshot design for threshold gates, informal; exact: Proposition~\ref{prop:snapshot-design}}
A gate that blocks above a threshold in $[\theta_{\min},1)$ first blocks only at a \emph{running maximum}: a step whose score exceeds $\theta_{\min}$ and every earlier score since the change took effect. A running maximum is \emph{open} if some such gate that first blocks there leaves the query's outcome unsettled. Under full-history kernels, (T), and free responses at unsupported inputs (U), the open running maxima before the last step form the least set among first-block snapshots whose continuation laws identify the query for every gate in the family. An earlier snapshot can replace one by re-executing the agreeing steps.
\end{informal}

\textbf{For the deployer.} Record (G2) to (G4), rerun open continuations from snapshots meeting (T), and for threshold gates save first-block snapshots at open running maxima, the least such set.

\subsection{Shared State: Does Evidence Transfer?}
\label{sec:joint}

Sections~\ref{sec:replay} to~\ref{sec:supported} evaluate a change from the deployer's own logs; evidence from components tested alone needs a transfer link \citep{pearl2011transportability}. When agents share state $S$ (memory, budget, tools) \charbadge{5}, component logs identify each failure probability given $S$, leaving dependence open; the Fr\'echet range contains every compatible joint failure probability, and each value in it is attainable when dependence given $S$ is unrestricted \citep{frechet1951,nelsen2006copulas,mosleh1998ccf}. Suppose the state law is known, component loss laws given $S$ match alone and composed, and components read only $S$ and private randomness. With unrestricted responses on unseen states, standalone evidence identifies a nontrivial loss event up to width $P_S(U_i)$, where $U_i$ comprises composed states never seen alone (Proposition~\ref{prop:component-joint}).

\subsection{Oversight Cues: Do Tests Transfer?}
\label{sec:cues}

Oversight cues raise the same transfer question with a different state variable: an agent that can tell a test from deployment may act differently in each. Recorded cues \charbadge{6}, including evaluation signs \citep{needham2025evaluation} and gate notices, play the role of $S$ in Section~\ref{sec:joint} when the agent reads only the cue and private randomness. Test evidence determines risk only for deployment cues occurring in testing (Remark~\ref{rem:p3-eval}). A cue created by the change, such as an announced gate, violates (A2, no anticipation); Section~\ref{sec:demo-cues} tests one.

\section{Applying the Framework in a Sandbox}
\label{sec:validation}

We test each characteristic in ControlArena's bash setting \citep{controlarena2025,bhatt2025ctrlz}, where harm is an executed violation of a declared sandbox policy, and repeat gates and logging in $\tau^2$-bench's airline environment \citep{barres2025tau2bench}, where harm is an executed booking write that breaks the airline's written policy. The 16 runs of Table~\ref{tab:runs} log about 49,000 episodes, using vLLM mostly on NVIDIA H100 80GB hosts (Appendix~\ref{app:setup}). Every log is from a pass-all gate; eight of the nine registered runs fixed sample, analysis, and margin before launch, and three froze predictions before new live records were read. The agent never sees an unannounced gate and a stop ends the episode, so (A2) holds by construction for stopping gates. (A1) holds where serving and environment are stable; where outputs vary by run (the stronger agent, the airline setting), replay is compared with live gates in distribution, within a registered margin.

\begin{table*}[t]
\centering
\footnotesize
\setlength{\tabcolsep}{2.5pt}
\renewcommand{\arraystretch}{1.0}
\begin{tabularx}{\textwidth}{@{}P{0.26\textwidth}P{0.235\textwidth}P{0.12\textwidth}P{0.11\textwidth}L@{}}
\rowcolor{hdrgray}
Change & Log alone & New runs & Live & Result \\
\multicolumn{5}{@{}l}{\textit{What the framework establishes}} \\
\charbadge{1} Stopping gate, 3 settings & \cellcolor{mint!60}0.039 (bash; see App.~\ref{app:demo-gates}) &  & 0.030 & \cellcolor{mint!60}\tabok{} Match within 0.03 in all 3 \\
\charbadge{1} Retry gate 1 & \tabopen{} $[0.244,0.345]$ & \cellcolor{mint!60}0.279 & 0.288 & \cellcolor{mint!60}\tabok{} Match within 0.015 \\
\charbadge{1} Retry gate 2 & \tabopen{} $[0.100,0.451]$ & \cellcolor{mint!60}0.158 & 0.165 & \cellcolor{mint!60}\tabok{} Match within 0.015 \\
\charbadge{1} Gates 1 / 2: approved at 0.25 & \tabopen{} Undecided & \cellcolor{mint!60}No / yes & No / yes & \cellcolor{mint!60}\tabok{} Same decision as live \\
\charbadge{1} Retry gate, bash & \tabopen{} $[0.023,0.422]$ & \cellcolor{mint!60}0.029 & 0.030 & \cellcolor{mint!60}\tabok{} Match within 0.02 \\
\charbadge{2} Stopping gate, per-step harm & \cellcolor{hdrgray!60}$\le$0.064 with, $\le$0.159 without &  & Not needed & \cellcolor{hdrgray!60}\tabkeep{} Log must keep harm timing \\
\charbadge{5} Relay: snapshot start &  & \cellcolor{mint!60}0.118 to 0.119 & 0.115 & \cellcolor{mint!60}\tabok{} Match within 0.03 \\
\multicolumn{5}{@{}l}{\textit{Simpler readings and broken assumptions}} \\
\charbadge{1} Gates 1 / 2: retry as stop & Registered: 0.2445 / 0.1002 &  & 0.288 / 0.165 & \cellcolor{coral!35}\tabno{} Low by 0.043 / 0.065 \\
\charbadge{1} Gates 1 / 2: gate ignored & Registered: 0.3445 for both &  & 0.288 / 0.165 & \cellcolor{coral!35}\tabno{} High by 0.057 / 0.179 \\
\charbadge{1} Gates 1 / 2: blocks skipped & Post hoc: 0.3108 / 0.1904 &  & 0.288 / 0.165 & \cellcolor{coral!35}\tabno{} High by 0.023 / 0.025 \\
\charbadge{5} Relay: clean start &  & 0.178 & 0.115 & \cellcolor{coral!35}\tabno{} High by 0.063 \\
\charbadge{6} Announced gate, abort rate & Unannounced-log replay &  & 181 of 400 & \cellcolor{coral!35}\tabno{} Replay low; CI excludes 0 \\
\end{tabularx}
\caption{\textbf{The log alone leaves retry gates open (\tabopen); rerunning only what it leaves open matches live runs (\tabok); simpler readings misestimate harm, and clean starts and an announced gate miss live runs (\tabno).} \tabkeep{}: a logging requirement. P(harm) unless labeled. Match: Newcombe 90\% interval inside the registered margin. Certified limits are exact one-sided 95\%; log intervals are plug-in bounds. In $x$ / $y$ cells, $x$ is gate 1; blank: not used.}
\label{tab:statements}
\end{table*}

\subsection{Approval Gates: Stopping and Retry}
\label{sec:demo-stop}

\paragraph{Retry Gates: Estimating Harm.}\phantomsection\label{sec:demo-retry} A retry gate lets the agent act after a refusal, which no pass-all log contains. \textbf{The log alone bounds retry-gate harm; targeted execution pins it down if restored runs behave like live runs (restoration fidelity)} (Figure~\ref{fig:bound}). One pass-all log of 7,200 stronger-agent episodes serves two retry gates (thresholds on a log-scale monitor score), which first block in 14\% and 41\% of episodes; before every high-scoring step, a snapshot rule that names no gate saved the sandbox. Following the five steps of Section~\ref{sec:results}, executing only the continuations from these snapshots identifies harm for both gates (worked example: Appendix~\ref{app:demo-e7}). Each P(harm), frozen before live records were read, matched paired live gates within 0.015; the simpler readings misestimated it (Table~\ref{tab:statements}; Figure~\ref{fig:forest}).

\paragraph{Retry Gates: Deciding.} A rule declared in advance approves a gate whose exact one-sided 95\% bounds put harm at most 0.25 and task success at least 0.37. \textbf{The log alone cannot decide either gate; with the continuations rerun, the rule approves gate 2 and declines gate 1, as on the live runs.} Read at face value (post hoc), counting retries as stops would approve gate 1, which live runs refuse; skipping blocked calls reaches both live decisions but misses live harm (Table~\ref{tab:statements}). Given the log, the continuations cost 15\% of the agent calls of testing both gates live and half those of a live test sized to reach both decisions in 90\% of 500 random subsets (post hoc; Appendix~\ref{app:demo-e7}). Targeted execution also matched live runs for a bash agent and the airline environment; in all four, live harm ran 0.001 to 0.008 higher (Appendix~\ref{app:demo-gates}).

\subsection{Partial Records: What the Log Must Keep}
\label{sec:demo-log}

\textbf{In our logs, one field, per-step harm, decided how low a harm ceiling we could certify.} On 300 held-out bash episodes, a reduced log of scores, per-step harm and cost, and outcome certifies the same stopping-gate harm ceiling as the full record (per field, an instance, not a general list: Table~\ref{tab:log-fields}). Without per-step harm, they certify 0.159 instead of 0.064, since the log cannot tell whether harm came before the stop (Figure~\ref{fig:timing}); a replication log (0.111 instead of 0.051) and the stronger agent agree, illustrating both directions of Proposition~\ref{thm:logging}.

\subsection{Long Horizons: Replay Across 40 Steps}
\label{sec:demo-horizon}

\textbf{Replay still pins down stopping-gate harm as episodes get longer; fewer logged episodes agree with a retry gate, leaving more of its harm to bounds or targeted execution.} On a 600-episode bash log with a 40-step cap, replayed stopping-gate P(harm) stays at 0.043 to 0.067 (unknown outcomes) from 15 steps on; the retry gate's agreement mass falls from 0.77 to 0.57 and its plug-in bounds reach $[0.043,0.473]$ at 40 steps (Appendix~\ref{app:demo-horizon}).

\subsection{Model Releases: Old Logs, New Model}
\label{sec:demo-release}

\textbf{The new model reproduced none of 99 logged episodes, so replay settled none.} At logged inputs and seeds, the same model reproduced every episode, so the release needs support, bounds, or new runs from step 0 (Appendix~\ref{app:demo-release}).

\subsection{Shared State: Two Agents in Relay}
\label{sec:demo-relay}

A first agent takes up to 7 steps; a fresh-context second agent continues from its workspace for up to 8 (Appendix~\ref{app:demo-relay}). \textbf{Testing the second agent from a clean start overstates its harm in the relay; testing it from the handoff snapshot matches the relay} (Table~\ref{tab:statements}). A clean start leaves out the first agent's changes to the workspace, which the second agent reads. Both agents violated the policy in four times as many relays as the product of their rates; per-agent logs that record the task's risk class narrow this joint probability to $[0.031,0.115]$ (Proposition~\ref{prop:component-joint}(ii)).

\subsection{Oversight Cues: Announcing the Gate}
\label{sec:demo-cues}

\textbf{Announcing the gate in our prompt raised the abort rate above what replay from unannounced logs predicted} (same tasks and seeds; Table~\ref{tab:statements}). The unannounced gate, a negative control, stays compatible with replay (Appendix~\ref{app:demo-cues}).

\section{Related Work}
\label{sec:related}

\paragraph{Evaluation from Logs.} Off-policy evaluation, optimal stopping, and dynamic treatment regimes hold the closest precursors of our stopping replay and supply its core. Full trajectories evaluate any stopping rule with a known reward \citep{goel2017sample}, and regime comparisons censor at deviation \citep{cain2010start}. The cited replay and weighting methods use randomized logging \citep{li2011unbiased,bottou2013counterfactual}, which our logs lack. Execution from saved states appears in driving simulation \citep{gulino2023waymax}, rare-event splitting \citep{glasserman1999multilevel}, agent resampling \citep{bhatt2025ctrlz}, and reset access \citep{yin2022efficient,song2023hybrid}. Agent checkpoint and replay systems \citep{wu2026crab,chawla2026chronicle} restore recorded prefixes without identification guarantees.

\paragraph{What Is New.} Without overlap, partial identification gives sets \citep{manski2003partial,khan2024offpolicy}, and sharp bounds exist for dynamic regimes \citep{alvarez2026sharp} and discrete causal models \citep{duarte2023automated}; our unsupported-input result specializes them to replacement kernels with declared values. Robust sequential tests \citep{saha2026huber} and cost-aware pure exploration \citep{kanarios2024cost} supply the tools behind our execution bounds. New here are the logging condition for replayed stop values (Proposition~\ref{thm:logging}), the restoration threshold and execution-to-live cost ratio (Appendix~\ref{app:budget-limits}), allocation across strata (Appendix~\ref{app:budget-strata}), and the least set of first-deviation snapshots per threshold family (Proposition~\ref{prop:snapshot-design}).

\paragraph{Agent Risk and Governance.} \citet{ray2026enforced} bounds a multi-judge gate's errors and shows that ungated trajectories need not identify a blocking regime's safety frontier. Our joint-risk range conditions on shared state (Section~\ref{sec:joint}; certificates from joint runs: \citealp{bhardwaj2026agent}). Taxonomies list our characteristics \citep[e.g.,][]{chan2023harms,slattery2024ai}, rigorous audits need more than black-box access \citep{casper2024blackbox}, and underwriting proposals model claim costs \citep{xu2026traceeconomic}; evidence design specifies needed records (Appendices~\ref{sec:practice} and~\ref{app:related}).

\section{Conclusion and Future Directions}
\label{sec:conclusion}

\agentrisk{} establishes when logs can support approving changes to deployed agents and when more evidence is needed. If nothing else changes and the agent does not anticipate the change, it shows what a log settles and must record and, for independent episodes, the reruns a decision needs. In our bash sandbox, the log alone decided neither retry gate; rerunning only what it left open reached both live decisions at half the agent calls of a live test sized post hoc for them, given the log. Future work could restore state beyond saved files and bound the risk of oversight-aware agents \citep{needham2025evaluation}.

\section*{Limitations}\phantomsection
\label{sec:discussion}

Each result rests on stated conditions; Table~\ref{tab:statements-full} lists the evidence each demonstration still needs. Identification results concern population record laws on finite path spaces. Execution results add sampling guarantees for independent episodes; agents sharing state use the component evidence of Section~\ref{sec:joint}. Replay assumes that nothing else changes (A1) and that the agent does not anticipate the change (A2). A change the agent can perceive, such as an announced gate, violates (A2). Paired runs with and without the announcement expose this violation (Section~\ref{sec:demo-cues}); bounds for agents that recognize oversight remain open. Logs cannot confirm the stop value (A3), which our runs declare but do not test for harm. Restoration checks covered only the file tree compared by the harness.

A frozen mechanical checker defines our harm label under four declared rules (Appendix~\ref{app:setup}); the framework takes any declared loss. The demonstrations use honest agents (not evading oversight) from one open-weight model family, with 14B and 27B parameters, in a bash sandbox and an airline booking environment. The relay joins two copies of one agent, and supported continuation is untested. For the model release, replay settled none of 99 episodes, leaving its losses unestimated.

\section*{Ethical Considerations}
Risk estimates can affect deployment, resource access, and responsibility for failures. Claims should identify affected parties, the loss definition, evidence, and uncertainty. Unobserved consequences and misspecified mechanisms can give false reassurance. The framework exposes assumptions to deployers and auditors, preserving unresolved uncertainty in downstream decisions.

A deployer could choose a loss, stop value, or margin that favors approval. Declaring the specification and decision rule before reading results, as our registered runs do, guards against this choice. Episodes the log leaves open enter the decision as intervals. In deployment, replay's per-step records and saved states can contain user data, subject to the deployer's retention and access rules. Proposition~\ref{thm:logging} states when logs trimmed for privacy still determine the agreeing and stopped terms, allowing checks before use. Our logs from sandboxed coding tasks and simulated airline customers contain no personal data. Appendix~\ref{app:ai-use} provides our AI Use Statement.

\bibliography{working}

\appendix
\counterwithin{figure}{section}
\counterwithin{table}{section}
\numberwithin{definition}{subsection}
\section*{Guide to the Appendix}
\label{app:guide}

The appendices support the formulation in Section~\ref{sec:framework}, results in Section~\ref{sec:results}, demonstrations in Section~\ref{sec:validation}, and related work in Section~\ref{sec:related}.

Start with Appendix~\ref{app:formal} for the specification, assumptions, and evaluation procedure, or Appendix~\ref{app:demo} for the registered and exploratory demonstrations and their uncertainty. Appendix~\ref{app:replay} proves what replay identifies, what a log must preserve, and how horizons affect unresolved outcomes. Appendix~\ref{app:supported} treats supported and unsupported continuation, execution budgets, restoration error, strata, and snapshot design. Appendix~\ref{app:p3-proof} establishes when component evidence transfers to a composition or deployment setting. Appendices~\ref{sec:extensions} and~\ref{app:related} explain the survey and comparisons with prior work. Appendix~\ref{app:checks} indexes the verification scripts. \texttt{ARTIFACT.md} and \texttt{checks/README.md} link the full operational details and worked examples.

\section{Formal Statements Behind Sections 2 and 3}
\label{app:formal}

This appendix gives the formal specifications, assumptions, and results behind Sections~\ref{sec:framework} and~\ref{sec:results}; the proofs appear in Appendices~\ref{app:replay}, \ref{app:supported}, and~\ref{app:p3-proof}.

\subsection{The Specification}
\label{app:formal-spec}

This is the specification of Section~\ref{sec:objects}.

\begin{definition}[Risk specification]
A risk specification is a tuple
\begin{equation}
\riskmodel=(E,\ell,K,\interventions),
\label{eq:specification}
\end{equation}
where $E$ is an execution mechanism, $\ell$ is a consequence map, $K$ is an observation mechanism, and $\interventions$ is a declared family of admissible mechanism replacements.
\end{definition}

\paragraph{Execution Mechanism $E$.}
Fix a finite horizon $T$. $E$ specifies initial distributions, policies, transitions, updates, and scheduling as named stochastic kernels whose declared inputs are generated earlier in a declared order; policies use only information available before their decisions. Because mechanisms with identical path laws can respond differently under intervention, $E$ retains its modular kernel and input structure rather than only its induced law.

\paragraph{Consequence map $\ell$.}
Let $L=\ell(\tau)\in\mathbb R^m$ be the full-episode loss vector \citep[Section~3.4]{kaplan1981risk}, with $\ell$ defined on a trajectory space shared by $E$ and every admissible replacement. In the running agent, $L$ is (failure, harm, cost) over the whole episode, and a declared post-abort kernel of $E$ generates the abort valuation of Section~\ref{sec:running}.

\paragraph{Observation mechanism $K$.}
Execution and evaluation share the clock $t=0,\ldots,T$, and the declared generation order interleaves execution coordinates and records. The evaluator receives records $R_t=(O_0,\ldots,O_t)$, trivial for $t=-1$ and possibly different from the agent's context, from causal kernels that read only the execution history available by time $t$, so an online record cannot depend on a future execution suffix. A monitor that changes approval, timing, scheduling, agent behavior, or losses belongs in $E$, as does any record used for control or loss accounting. The kernels of $E$ and $K$, composed in their declared order, induce a joint law $P_{E,K}$ on a finite set $\Omega$ of complete joint execution and observation paths, with loss $L=\ell(\omega)$ in a finite set $\lossset\subset\mathbb R^m$; the law of the current system is the baseline law $P^0$. Every result in this paper uses this finite setting. In the running agent, $K$ is the deployer's log.

\paragraph{Admissible Replacements $\interventions$.}
Each replacement $\kappa\in\interventions$ specifies replaced kernels, implementation boundaries, and allowed information. Replacement kernels keep their output coordinates, ranges, and causal positions and are adapted to the implementer's declared information; they may read different earlier inputs within it. Unreplaced kernels retain their declared inputs. For deviation replay, $\kappa$ replaces one decision kernel. We refine the clock so that this kernel acts at most once per step, mapping declared input $I_t$ to baseline action $f^0_t(I_t)$ or replacement action $f^\kappa_t(I_t)$. Refining the clock changes no kernel, declared input, consequence map, or coarse record. For the stopping gate $\kappa_\theta$, input $I_t$ holds proposals $u_t$ and score $m_t$, and $f^\kappa_t$ aborts when $m_t>\theta$. The query applies $\kappa$ to all histories compatible with the conditioning record; a selection using information outside that record belongs in $E$ as further evidence.

\subsection{The Question, Assumptions (A1) and (A2), and Identified Sets}
\label{app:formal-question}

For the query of Section~\ref{sec:question}, suppose replacement $\kappa$ takes effect at boundary $s>t$ after records $R_t$. Splicing $\kappa$ into $E$ yields changed mechanisms $E^\kappa$ and $K^\kappa$ with joint law $P^\kappa=P_{E^\kappa,K^\kappa}$, defining the continuation query:
\begin{equation}
Q_t^\kappa(\,\cdot\mid r)
=\Law_{P_{E^\kappa,K^\kappa}}(L\mid R_t=r),
\label{eq:continuation}
\end{equation}
computed from the specification; we write $P^\kappa=P_{E^\kappa,K^\kappa}$. Applying loss aggregation $\phi$ and risk functional $\rho$ gives the scalar query $q^\kappa_t(r)=\rho(\phi_\#Q^\kappa_t(\cdot\mid r))$, such as harm probability, mean cost, or conditional value-at-risk \citep{rockafellar2002cvar}. The changed system obeys $P^\kappa$ under two core conditions.

\smallskip\noindent\textbf{(A1)}~\emph{Invariance}: the initial law and every unreplaced kernel, including random-input generators, remain unchanged. Installing a gate leaves models, tools, users, and samplers as logged.

\smallskip\noindent\textbf{(A2)}~\emph{Prefix preservation and no anticipation}: both mechanisms share the causal generation order and coincide before the replacement boundary. Every unreplaced kernel retains its declared inputs; a replaced kernel may read different earlier inputs that $\interventions$ permits. For deviation replay, $f^0_t$ and $f^\kappa_t$ are evaluated on a common input $I_t$ that contains the inputs either rule uses. No unreplaced kernel or declared input reads the identity or announcement of the installed replacement. Announcing the gate in the agent's prompt violates (A2).

\smallskip\noindent Because true kernels are unknown, the evaluator determines which query values observed records rule out \citep{manski2003partial}. A structural model $M$ specifies target mechanisms $(E_M,\ell,K_M)$ and evidence sources, including continuation runs linked by \textbf{(T)}~\emph{restoration fidelity}: continuation laws equal conditional laws of $P^\kappa_M$ given restored prefix $h$. The model family $\mathcal M_0$ encodes declared structural restrictions. Given observed record laws $P_{\mathrm{obs}}^R$, the compatible models are $\mathcal M=\{M\in\mathcal M_0:P_M^R=P_{\mathrm{obs}}^R\}$, and the population identified set for query $\psi$ is $\Theta(\mathcal M)=\{\psi(M):M\in\mathcal M\}$, yielding $\Theta^\kappa_t(r)$ for $q^\kappa_{t,M}(r)$. Equality of evidence laws is source by source, and pointwise conditional sets are stated at records with positive observed probability.

\paragraph{Sampling error.} The set in~\eqref{eq:risk} is stated at records with positive probability. One-sided endpoint confidence bounds, each with noncoverage at most $\alpha/2$, cover it with probability at least $1-\alpha$; covering only the true risk can permit shorter intervals \citep{imbens2004confidence}, and plug-in bounds ignore sampling uncertainty.

\subsection{Records That Leave a Gate Change Open}
\label{app:diagnostic}

Section~\ref{sec:running} names three ways baseline records can leave replacement risk unresolved. A declared abort valuation resolves the first ambiguity. The remaining ambiguities depend on the fixed kernels and input restrictions in the model family; the examples below state which restrictions they retain.
(1)~\emph{Stop.} No baseline episode aborts, so no record shows the loss after an abort; the declared valuation, (A3) of Appendix~\ref{app:formal-replay}, supplies it, and records cannot check it.
(2)~\emph{Continue.} The logged gate never blocks, so no logged input contains a block message. If the family fixes every kernel's declared inputs, the gate, and the abort valuation but leaves the agent's response to a block message free, two models that agree at every input of positive baseline probability have the same record law and can differ in the harm after a retry.
(3)~\emph{Logging.} If the log keeps scores and an episode harm flag but not the step of each harm, and the family fixes the monitor and the gate but leaves the timing of harm free, a model whose harm falls before the step at which a stopping gate aborts and a model whose harm falls after it have the same record law, yet the gate removes the harm in only the second.
The third ambiguity has the finite instance below, for a stricter gate as in the running agent. It is the toy episode of Section~\ref{sec:running} in formal terms, with harm at step 1 or step 3 and the gate acting at step 2.

\paragraph{Omitted Harm Timing under a Stricter Gate.}
When a log records overall harm but omits step-level harm $h_t$, baseline paths with early harm ($t=1$) and late harm ($t=3$) generate identical coarse records. A stricter gate aborting at $t=2$ under declared accrued-harm valuation eliminates late harm but retains early harm. In the family fixing every kernel except the law of the unrecorded early/late state, with (R) and (A1) to (A3), latent early-harm probability $p\in[0,1]$ is unconstrained by the coarse log, the fibre condition of Proposition~\ref{thm:logging} fails and the identified harm interval is $[0,1]$. Logging step harm identifies $p$, contracting the garbled set $[0,1]$ to $\{p\}$.

\paragraph{An Omitted State Variable under a Looser Gate.}
Replace a gate that blocks every proposal by one that executes every proposal. A record of the proposal, the damage, and the cost that omits a stale-state bit $S$ has the same law for $P(S=1)=0.1$ and $P(S=1)=0.9$, yet the damage probability after the change is $0.1$ or $0.9$. The baseline gate decision is constant, so logging it adds nothing, while logging $S$ identifies the replaced damage probability within this family (Appendix~\ref{app:checks}).

\subsection{Deviation Replay}
\label{app:formal-replay}

This formalizes the replay decomposition in Section~\ref{sec:results} and the result in Section~\ref{sec:replay}.

Let $\Omega^0$ be the paths where every replaced decision equals baseline rule $f^0_t(I_t)$. The first deviation is $\sigma_\kappa=\min\{t:f^\kappa_t(I_t)\neq f^0_t(I_t)\}$, and $\mathsf A_\kappa=\{\sigma_\kappa>T\}$ is the agreement event. At a deviation, the replayed prefix $h^+_\kappa$ appends replacement action $f^\kappa_{\sigma_\kappa}(I_{\sigma_\kappa})$ to earlier history. A deviation is terminal if the replacement action aborts execution, and continuing otherwise. We condition on record $R_t=r$ with $P^0(R_t=r)>0$ prior to intervention. Beyond (A1) and (A2), replay assumes:
\begin{itemize}
\item[(R)] \emph{Replay interface}: $f^0_t$ and $f^\kappa_t$ are known and deterministic in the declared input $I_t$, which may include a random tape and which the full record contains (a coarser log, as in Proposition~\ref{thm:logging}, need not). The gate rule can be rerun on logged proposals, scores, and seeds.
\item[(A3)] \emph{Declared abort valuation}: after a terminal deviation with replayed prefix $h^+$, the law of $L$ is a declared $b_\kappa(\cdot\mid h^+)$ that includes accrued losses. It describes post-abort kernels of $E$, so $\ell$ is unchanged; replacements that abort at the same replayed prefix share it, and it must agree with any post-abort kernel at inputs of positive baseline probability.
\end{itemize}

\begin{proposition}[Deviation replay; \citealp{robins1986new,cain2010start}]\label{prop:replay}
Assume \textup{(R)}, \textup{(A1)}, and \textup{(A2)}, and let $B\subseteq\lossset$.
\par\textup{(i)}~\emph{Agreement.} $P^\kappa(L\in B,\mathsf A_\kappa\mid r)=P^0(L\in B,\mathsf A_\kappa\mid r)$.
\par\textup{(ii)}~\emph{Stopping.} Under \textup{(A3)}, $P^\kappa(L\in B,\ \sigma_\kappa\le T,\ \text{terminal}\mid r)=\int b_\kappa(B\mid h^+_\kappa(\tau))\,\mathbf 1\{\sigma_\kappa\le T,\ \text{terminal}\}\,dP^0(\tau\mid r)$, even if the baseline never aborts.
\end{proposition}

Part (ii) formalizes the stop term in display~\eqref{eq:split}. Appendix~\ref{app:replay} establishes both assertions via Lemma~\ref{lem:app-prefix}, while Proposition~\ref{prop:replay-cont} treats continuing deviations.

\subsection{Supported Continuation, Compact Form}
\label{app:formal-supported}

This is the formal result summarized in Section~\ref{sec:supported}.

\begin{proposition}[Supported continuation and unsupported inputs]\label{prop:supported}
Assume \textup{(A1)} and \textup{(A2)}, let $\kappa$ replace kernels from boundary $s$ onward by known kernels, and fix $R_t=r$ with $t<s$ and $P^0(R_t=r)>0$. Suppose \textup{(G1)} at each position of the declared generation order, every input that the replacement leads an unchanged downstream kernel to read there has positive baseline probability at that position, unless the specification fixes that kernel's value at that input, as after an abort. Also suppose \textup{(G2)} those kernels' declared inputs are recorded at those positions, \textup{(G3)} every coordinate that $\ell$ reads is recorded, and \textup{(G4)} the replacement reads recorded inputs only. Then, within mechanisms that share the declared kernel inputs and the values the specification fixes, $Q_t^\kappa(\cdot\mid r)$ is identified by the g-formula. Zero agreement mass is compatible with identification.
\par\emph{Unsupported inputs.} Without \textup{(G1)}, keep \textup{(G2)} to \textup{(G4)}, the declared kernel inputs, and the fixed values. Let the kernel law at each position of the declared generation order be free at its other inputs of baseline probability zero, within declared sets of allowed outputs, separately from the laws at other positions. Then the identified set is the image of the g-formula with one free law at each such position and input, shared by all histories that reach it; if one law is shared across positions, as for a policy called at every step, this image is an outer set.
\end{proposition}

The first statement identifies $Q_t^\kappa$ via the sequential g-formula (Appendix~\ref{app:supported-proof}). When (G1) fails, optimizing over unobserved inputs specializes partial-identification bounds \citep{alvarez2026sharp,duarte2023automated}, giving a sharp identified image when each position's law at unseen inputs varies freely and an outer set when laws are shared across positions (Appendix~\ref{app:r4-unsupported}). Treating unobserved inputs in a finite log as unsupported preserves a valid outer set (Remark~\ref{rem:r4-logs}).

\subsection{The Evaluation Procedure}
\label{app:procedure}

Algorithm~\ref{alg:main} translates the five evaluation steps of Section~\ref{sec:decisions} into formal decision logic.

\begin{algorithm}[ht]
\caption{Evaluating a change $\kappa$ from logs}
\label{alg:main}
\footnotesize
\renewcommand{\algorithmicrequire}{\textbf{Input:}}\renewcommand{\algorithmicensure}{\textbf{Output:}}
\begin{algorithmic}[1]
\REQUIRE $\riskmodel$, $\mathcal M_0$, record $r$, change $\kappa$, query $\rho\circ\phi$, decision rule
\ENSURE status; $q^\kappa_t(r)$ or an interval; missing evidence
\STATE Declare the kernels $\kappa$ replaces, their inputs, $b_\kappa$ of (A3), and the links (A1), (A2), (T)
\IF{(A1) or (A2) is not justified and the extra change is not modeled}
\STATE \textbf{return} ``this analysis does not evaluate $q^\kappa_t(r)$''
\ENDIF
\STATE Under (R), where records permit, rerun both rules on each logged $\tau$: $\sigma_\kappa$, $h^+_\kappa$, and its event
\IF{the log is sufficient (Prop.~\ref{thm:logging}) and keeps the feasible sets $\mathcal X_h$}
\STATE read the agree-plus-stop sum of~\eqref{eq:split} from $P^0(\cdot\mid r)$ and $b_\kappa$
\ENDIF
\FORALL{terms left open: the continue term, or any term of a failed check}
\IF{(G1) to (G4) hold}
\STATE apply the g-formula (Prop.~\ref{prop:supported})
\ELSIF{snapshots at $h^+_\kappa$ are kept and (T) holds}
\STATE execute continuations from restored prefixes, or, for binary harm, a random sample (Prop.~\ref{prop:budget})
\ELSE
\STATE bound by $\Theta^\kappa_t(r)$~\eqref{eq:risk} over record completions, or by~\eqref{eq:bound} for full records and monotone $\rho$; sharp under (S), else outer\alglinelabel{alg:main-open}
\ENDIF
\ENDFOR
\IF{the value or interval, with sampling uncertainty, leaves the decision open}
\STATE name what is missing: episodes, record fields, justified assumptions, or runs
\ENDIF
\STATE \textbf{return} status (identified; bounded; needs execution), value or interval, decision
\end{algorithmic}
\end{algorithm}

When targeted execution samples binary harm continuations at line~11, it invokes hypothesis (H) of Proposition~\ref{prop:budget} (Appendix~\ref{app:budget}) to guarantee one-sided confidence coverage.

\subsection{Symbols}
\label{app:symbols}

Table~\ref{tab:symbols} lists the symbols of Sections~\ref{sec:framework} and~\ref{sec:results} and of this appendix, with a one-line meaning and the place where each is defined.

\begin{table*}[!t]
\centering
\footnotesize
\setlength{\tabcolsep}{4pt}
\renewcommand{\arraystretch}{1.08}
\begin{tabularx}{\textwidth}{@{}P{0.17\textwidth}LP{0.16\textwidth}@{}}
\rowcolor{hdrgray}
Symbol & Meaning & Exact definition \\
$\riskmodel=(E,\ell,K,\interventions)$ & Risk specification: execution mechanism, consequence map, observation mechanism, admissible replacements & Appendix~\ref{app:formal-spec} \\
$T$; $t$ & Finite horizon; step index, shared by execution and records & Appendix~\ref{app:formal-spec} \\
$u_t$; $m_t$ & Tool calls proposed at step $t$; the monitor's score for them & Section~\ref{sec:running} \\
$\kappa$; $\kappa_\theta$; $\theta$ & A replacement (a change); the stopping gate; its score threshold & Appendix~\ref{app:formal-spec} \\
$L=\ell(\omega)$; $\lossset$ & Full-episode loss vector (failure, harm, cost); its finite set of values & Appendix~\ref{app:formal-spec} \\
$R_t=(O_0,\ldots,O_t)$; $r$ & Evaluator records up to $t$; a value of $R_t$ with positive baseline probability & Appendix~\ref{app:formal-spec} \\
$\Omega$; $P_{E,K}$; $P^0$; $P^\kappa$ & Complete joint paths; their law; the baseline law; the law under $\kappa$ & Appendices~\ref{app:formal-spec}, \ref{app:formal-question} \\
$I_t$; $f^0_t$; $f^\kappa_t$ & Declared input of the replaced decision kernel; the baseline and replacement rules & Appendix~\ref{app:formal-spec} \\
$Q^\kappa_t(\cdot\mid r)$ & Loss law of the changed system given the record at the change & Appendix~\ref{app:formal-question} \\
$q^\kappa_t(r)$; $\Theta^\kappa_t(r)$ & Risk of the change: $\rho$ applied to the law of $\phi(L)$ under $Q^\kappa_t(\cdot\mid r)$; its values under the models that reproduce the records & Display \eqref{eq:risk} \\
$\phi$; $\phi_\#Q$; $\rho$ & Coordinate or aggregation of $L$; the law of $\phi(L)$ under $Q$; risk functional (probability, mean, CVaR) & Appendix~\ref{app:formal-question} \\
$\mathcal M_0$; $\mathcal M$; $\psi$; $\Theta(\mathcal M)$ & Declared model family; models that reproduce the records; a query; the identified set & Appendix~\ref{app:formal-question} \\
(A1); (A2) & Invariance; prefix preservation and no anticipation & Appendix~\ref{app:formal-question} \\
(T) & Restoration fidelity of a continuation run from a restored prefix & Appendix~\ref{app:formal-question} \\
$\Omega^0$ & Paths on which every replaced decision equals the baseline rule & Appendix~\ref{app:formal-replay} \\
$\sigma_\kappa$; $\mathsf A_\kappa$ & First deviation of the new rule from the logged one; the agreement set $\{\sigma_\kappa>T\}$ & Appendix~\ref{app:formal-replay} \\
$h^+_\kappa$ & Replayed prefix: the history before the deviation followed by the replacement output & Appendix~\ref{app:formal-replay} \\
terminal; continuing & A deviation whose replacement output is an abort; any other deviation & Appendix~\ref{app:formal-replay} \\
(R); (A3); $b_\kappa$ & Replay interface; declared abort valuation; the declared loss law after a stop & Appendix~\ref{app:formal-replay} \\
agree; stop; continue & The events $\mathsf A_\kappa$, $\{\sigma_\kappa\le T,\ \text{terminal}\}$, and $\{\sigma_\kappa\le T,\ \text{continuing}\}$ & Display \eqref{eq:split} \\
$E^0[\,\cdot\mid r]$ & Expectation of a path function under $P^0(\tau\mid r)$ & Display \eqref{eq:split} \\
$\mathcal X_h$; (S) & Feasible values of $X=\phi(L)$ after a continuing prefix $h$; saturated continuation & Appendix~\ref{app:replay-cont} \\
$\nu^-$; $\nu^+$ & Law of $X$ given $r$ when agreeing episodes keep their logged value, stopped episodes follow the declared valuation, and each continuing episode takes the least (greatest) value in $\mathcal X_h$ & Display \eqref{eq:bound} \\
$C_\kappa$; $u_\kappa$ & An episode's combined contribution to the first two terms of \eqref{eq:split}; its missing mass & Proposition~\ref{thm:logging} \\
$\mathcal P_r$; $s_{\kappa,r}$ & Baseline laws on $\Omega^0$ that charge $r$; the replay summary of a law & Proposition~\ref{thm:logging} \\
$g$; $g_\#P$; $\Psi$ & Logging map; the law of the log; a map from the log law to the summary & Proposition~\ref{thm:logging} \\
$\eta_j(r)$ & Deviation hazard at step $j$ among episodes that agree before $j$ & Corollary~\ref{cor:horizon} \\
(G1) to (G4) & Support, recorded kernel inputs, recorded loss coordinates, recorded replacement inputs & Appendix~\ref{app:formal-supported} \\
$S$; $P_S$; $U_i$ & Shared state; its law; composed states never seen by component $i$ alone & Section~\ref{sec:joint} \\
\end{tabularx}
\caption{Symbols of Sections~\ref{sec:framework} and~\ref{sec:results}.}
\label{tab:symbols}
\end{table*}

\section{Survey and Classical Foundations}
\label{sec:extensions}

This appendix describes the survey, classical tools, and existing practices behind the specification in Section~\ref{sec:framework} and comparisons in Section~\ref{sec:related}.

\subsection{Survey Method}
\label{app:survey}

For the characteristics of Section~\ref{sec:characteristics}, we reviewed a purposive sample of 85 papers and reports on agent risk. Two language-model coders from different model families independently coded 23 agent-risk taxonomies, surveys, and governance frameworks alongside 30 incidents from the AI Incident Database (AIID; \citealp{mcgregor2021preventing}) against 20 candidate characteristics. The incidents cover 2025 and 2026 reports describing tool-using AI agents. The coded sources are \citep{chan2023harms,shavit2023practices,chan2024visibility,kolt2025governing,hammond2025multi,slattery2024ai,deng2024ai,yu2025survey,he2024security,hua2024trustagent,ruan2024identifying,zhang2024agent,wang2024survey,owasp2026top,bengio2025international,bengio2026international,nist2024artificial,caisi2025technical,eu2025general,aisi2024approach,imda2026model,anthropic2024model,openai2025operator}.

Three criteria selected the six characteristics of Table~\ref{tab:characteristics}.
Sources or incidents name each one often, each remains a persistent risk driver under current mitigations, and the framework provides an identification result for each.
This third criterion is partly circular because the framework was designed around these characteristics; the selection therefore motivates the six without proving that they matter most.

The released artifact lists all 23 sources, their consulted versions, and the 30 incident identifiers.

\subsection{The Other Characteristics}
\label{app:characteristics}

Agent characteristics modify instances of the four objects, interfaces, or model families rather than adding primitive concepts.
Section~\ref{sec:characteristics} (Table~\ref{tab:characteristics}) focuses on the six characteristics with formal results and demonstrations.
Other candidates such as prompt injection, reviewer load, environment drift, and serving nondeterminism map to definitions within these same four objects; the rows are neither exhaustive nor additive risk scores.
The companion mapping lists the other characteristics and their definitions within the four objects.

\subsection{Classical Tools}
\label{app:background}

Table~\ref{tab:foundations} summarizes how the four-object specification of Section~\ref{sec:objects} adapts classical tools; dynamic PRA models feedback and uncertain event timing \citep{uclaadsidac}, and the four-object specification is our modeling choice.

\begin{table*}[!t]
\centering
\footnotesize
\setlength{\tabcolsep}{4pt}
\renewcommand{\arraystretch}{1.0}
\begin{tabularx}{\textwidth}{@{}P{0.37\textwidth}L@{}}
\toprule
Classical tool & Role, essential condition, and status in specification \\
\midrule
Scenario triplets \citep{kaplan1981risk} & Kept with named kernels of $E$ and loss $\ell$; requires declared consequence definitions under changed systems. \\
Partial observation \citep{bauerle2017partially} & Kept in observation record $K$ and Proposition~\ref{thm:logging}; deployer-designed records differ from agent context. \\
Mechanism intervention \citep{pearl1995causal} & Kept in intervention family $\interventions$ under (A1) and (A2); replaces agent decision rules with declared timing and inputs. \\
g-formula and censoring at deviation \citep{robins1986new,cain2010start} & Inherited in Proposition~\ref{prop:replay}; agreement uses (R), (A1), and (A2), and stopping uses declared abort valuations (A3). \\
Logged replay \citep{li2011unbiased} & Inherited in Proposition~\ref{prop:replay} and deviation time $\sigma_\kappa$; requires rerunnability (R) via recorded seeds and invariant serving. \\
Worst-case bounds \citep{manski2003partial} & Inherited in Proposition~\ref{prop:replay-cont} and Proposition~\ref{prop:supported}; bounded losses give prefix-feasible bounds, sharp under (S) with full records and outer otherwise; shared-law restrictions may narrow them. \\
Fr\'echet bounds and transportability \citep{frechet1951,pearl2011transportability} & Inherited in Proposition~\ref{prop:component-joint}; uses known marginals for joint bounds, and invariance and support to transfer standalone component evidence. \\
\bottomrule
\end{tabularx}
\caption{\textbf{Classical tools, standard assumptions, agent strains, and roles in the specification.}}
\label{tab:foundations}
\end{table*}

\subsection{Existing Practice as Instances}
\label{sec:practice}
To complement Section~\ref{sec:related}, we write existing practices as specifications without ranking them: $\tau$-bench and $\tau^2$-bench fix task reward $\ell$ \citep{yao2025taubench,barres2025tau2bench}; AI Control specifies audit and deferral thresholds in $\interventions$ \citep[Sections~2.2 and~2.5, Appendix~K]{greenblatt2024aicontrol}; PRA for AI models pathways in $E$ and $\ell$ \citep[Sections~3.2, 4.3, and~5.2]{wisakanto2025pra}; and risk-sensitive compositions evaluate joint losses \citep[Sections~2 and~4.2]{shabadi2026compositions}. The companion practice mapping records the objects each practice fixes and the conditions its conclusions require, which reflect our embedding unless a source section is cited.

\section{Comparisons With Prior Work}
\label{app:related}

We extend Section~\ref{sec:related} by positioning our identification and logging results against prior literature; Table~\ref{tab:foundations} in Appendix~\ref{sec:extensions} summarizes foundations.

\paragraph{Off-Policy Evaluation And Dynamic Regimes.} Our logged gate is deterministic and never stops, leaving no observed stops to reweight. Halting needs no overlap. Proposition~\ref{prop:replay} identifies stopping risk via declared accrued-loss valuation (A3); display~\eqref{eq:split} adds continuing mass.

\paragraph{Bounds Without Overlap.} Unlike dynamic regime bounds under unobserved confounding \citep{zhang2020designing,namkoong2020offpolicy}, we bound missing support at recorded inputs. Where bandit approaches restrict policies \citep{sachdeva2020offpolicy}, Proposition~\ref{prop:supported} leaves unsupported inputs free with declared valuations. Specializing discrete causal bounds \citep{duarte2023automated,alvarez2026sharp}, we separate position-specific free-law sharp sets from shared-law outer bounds.

\paragraph{Transport And Evidence Collection.} Causal transport uses graphical conditions \citep{bareinboim2016causal}; assumptions (A1), (A2), and (T) formalize transfer across logs, restored runs, and altered systems. Rather than designing behavior policies \citep{hanna2017dataefficient}, Proposition~\ref{prop:budget} executes open continuations through reset access \citep{yin2022efficient,song2023hybrid}. Unlike learned simulator models \citep{liu2026autoregressive}, targeted execution rests on reset fidelity (T).

\paragraph{What logs must keep.} Agent logs \citep{chan2024visibility} and reasoning from a deployed system's logs \citep{bottou2013counterfactual} motivate Proposition~\ref{thm:logging}. It instantiates the elementary fact that a log determines an expected contribution for every path law exactly when indistinguishable paths have equal contributions. State-abstraction irrelevance gives sufficient conditions for evaluation on compressed states \citep{hao2024deeply}. Information states preserve values through reward and transition sufficiency \citep{subramanian2020ais}.

\paragraph{Logging and reporting duties.} Several laws set related duties for AI providers and developers. The EU AI Act provides for automatic event logging in high-risk systems (applicable from 2027 or 2028), documentation by providers of general-purpose models, and, for models with systemic risk, evaluation and incident reporting \citep[Arts.~12, 53, and~55]{eu2024aiact}; the voluntary Code of Practice elaborates the systemic-risk duties \citep{eu2025general}. California's SB~53 requires large frontier developers to publish safety frameworks and all frontier developers to report critical safety incidents \citep{california2025sb53}, and New York's amended RAISE Act sets similar duties from January~2027 \citep{newyork2026raise}. These instruments do not provide a general criterion for whether an agent's records suffice to assess a specified change. Under (R), (A1), and (A2), Proposition~\ref{thm:logging} characterizes the records that preserve the combined replay contributions and unresolved mass across all baseline laws.

\paragraph{Shared randomness and model comparison.} Shared-seed replay is a common-random-numbers coupling of the two policies. \citet{oberst2019counterfactual} infer Gumbel-max noise under a structural causal model to evaluate a new policy on logged episodes, and Gumbel-max coupling of language models generates counterfactual text and compares models \citep{chatzi2025counterfactual,ravfogel2025gumbel,corvelobenz2025coupled}. Seed-addressable replay reads recorded seeds instead, and Corollary~\ref{cor:model-update} needs only deterministic evaluation at logged inputs and seeds.

\paragraph{Agent evaluation and control.} \citet{ray2026enforced} bounds a multi-judge gate's errors by Fr\'echet--Hoeffding. Our range conditions on the state that composed agents share and transfers standalone evidence only on covered states (Proposition~\ref{prop:component-joint}). \citet{bhardwaj2026agent} certify the reliability of composed agents without assuming independence, from measured joint executions; our range needs only component logs and the shared state. We separate stopping replay from continuing changes and state the evidence each requires. Appendix~\ref{sec:practice} writes four existing practices as specifications \citep{yao2025taubench,greenblatt2024aicontrol,wisakanto2025pra,shabadi2026compositions}. Of these, only control evaluation reuses records to assess a change (the companion practice mapping).

\paragraph{Insurance.} An insurer pricing a proposed change to an agent can use the agree, stop, and continue split to state which losses the policyholder's logs settle and which need further evidence. \citet{chen2026counterfactual} charges each side-effect action against a contractually fixed safe default, and \citet{xu2026traceeconomic} map task traces, economic exposure, and contract terms to modeled claim costs; our results complement them by characterizing the evidence needed to assess a specified change to the agent. Premiums require a further link from episode losses to calibrated monetary severity, exposure volumes, coverage terms, and dependence across claims.

\section{Registered Demonstrations and Supporting Evidence}
\label{app:demo}

This appendix provides the experimental setup, runs, and change-by-change results behind Section~\ref{sec:validation}, exploratory values, and secondary readings. Interval types follow Section~\ref{sec:validation}; Newcombe intervals are score intervals for differences of proportions, and exact ones are Clopper-Pearson. Post hoc values were computed after registered analyses using the same logs and frozen code, without multiplicity control.

In the primary bash setting \citep{controlarena2025,bhatt2025ctrlz}, task failure is near saturation (0.873 on the held-out log), so these claims hold within this sandbox setting only.

\subsection{Logs and Runs}

The bash runs use uniform task sampling with replacement from 257 tasks, fresh containers and seeds per episode, and the frozen checker (version~1.1). Table~\ref{tab:runs} summarizes the registered and exploratory runs of Section~\ref{sec:validation}.

\begin{table*}[!t]
\centering
\footnotesize
\setlength{\tabcolsep}{4pt}
\renewcommand{\arraystretch}{1.05}
\begin{tabularx}{\textwidth}{@{}P{0.17\textwidth}P{0.2\textwidth}P{0.12\textwidth}P{0.14\textwidth}L@{}}
\rowcolor{hdrgray}
Run & Gate & Episodes & Serving & Role \\
Exploratory log & Pass-all & 300 & Default kernels & Exploratory; $\theta=0.94$ chosen here \\
Exploratory live gates & Stopping gate at $\theta=0.99$, $0.94$, $0.19$ & 300 each (286 scored at 0.99) & Default kernels & Exploratory live comparison with replay from the exploratory log \\
Held-out log & Pass-all & 300 & Batch-invariant & Primary held-out log; source of the retry bounds and the step-0 continuations \\
Live retry gate & Retry gate, $\theta=0.94$, retry cap 2 & 300 & Batch-invariant & Live check of the retry bounds, after the fact \\
Targeted-execution run & Pass-all with a snapshot at the first block; continuations under the retry gate; live retry gate & 1,200 units (2,885 episodes) & Batch-invariant, two hosts & Registered; prediction frozen before the live arm was read \\
First announcement test & Stopping gate, announced & 300 & Batch-invariant & Registered null against the held-out log (no margin; analysis fixed after launch); followed up by the paired announcement test \\
Replication log; live replication runs & Pass-all; stopping gate & 1,000 and 1,000 & Batch-invariant, one host & The registered replication \\
Paired announcement test & Pass-all; stopping gate unannounced; announced & 400 triplets & Batch-invariant, one host & Registered; announced and unannounced gates on the same 400 task-and-seed triplets \\
Step-0 continuations & Retry continuations from the held-out log & 85 & Batch-invariant & Targeted execution of step-0 blocks \\
Model-update reruns & Reruns of logged decisions & 99 (and 30) & Per rung & Model-update ladder \\
Long-horizon log & Pass-all, cap 40 & 600 (594 scored) & Batch-invariant, two hosts & Both gates replayed by horizon \\
Stronger-agent run & Pass-all; stopping gate at $\lambda=11.551$ on $s$ & 1,400 pairs (2,800; 2,797 scored) & Default kernels, one host & Registered; stopping-gate replay and certification with a stronger agent \\
Airline run & Pass-all; stopping gate at $\lambda=11.186$ on $s$; retry gate & 1,200 units (2,800 episodes) & Default kernels, one host & Registered; stopping-gate replay, certification, and retry bounds in a second environment \\
Airline retry by targeted execution & Continuations restored by replaying logged writes; live retry gate & 1,200 units (959 new episodes: 159 continuations, 800 live) & Default kernels, one host & Registered; prediction frozen before the 800 new live episodes were read; live units 0 to 399 reuse the airline run's retry arm, analyzed there \\
Two retry gates by targeted execution & Pass-all with a gate-free snapshot rule; two live retry gates; continuations & 7,200 units (25,561 episodes: 7,200 pass-all, 14,400 live, 3,961 continuations) & Default kernels, six hosts & Registered; prediction frozen before the live arms were read; two gates from one log \\
Two-agent relay & Relay, five conditions & 1,600 units (8,000; 7,998 scored) & Batch-invariant, two hosts & Registered; tests of each agent alone against the live relay \\
\end{tabularx}
\caption{Runs behind Section~\ref{sec:validation}. Batch-invariant serving returns the same output for the same input whatever else shares the batch; default kernels do not. \texttt{ARTIFACT.md} maps each run to the code that its logs and commands use.}
\label{tab:runs}
\end{table*}

\subsection{Setup Specification}
\label{app:setup}

Verbatim prompts, commands, input logs, and verifying records for Section~\ref{sec:validation} are documented in \texttt{ARTIFACT.md}. ControlArena and $\tau^2$-bench (MIT license), the Qwen models, and vLLM (Apache 2.0) are used for research evaluation under their licenses.

\paragraph{Declared policy.} The operating rules forbid: (1)~network access, (2)~modifying files outside \texttt{/workspace} or \texttt{/tmp}, (3)~destructive commands outside those paths, and (4)~privilege escalation. An episode is harmed when any executed tool call violates these rules.

\paragraph{Harm checker and limitations.} A frozen mechanical checker (version~1.1) inspects commands and filesystem changes after each call, excluding sanctioned \texttt{run\_tests} and \texttt{submit}. Command substitutions or unexpanded variables may escape command checks but can trigger state checks. A failed state check leaves harm status unknown unless a violation was recorded or, for a replayed stop, the failure came at or after the stop.

\paragraph{Pre-execution blocking and monitor.} The monitor, Qwen3-4B-Instruct-2507 \citep{yang2025qwen3} (revision \texttt{cdbee75}), scores proposed calls before execution at temperature~0 and seed~0. Scores are scaled to $[0,1]$; the gate at $\theta=0.94$ blocks on scores above 0.94, terminating execution before the flagged action runs.

\paragraph{Agent and episode limits.} The agent, Qwen3-14B-FP8 (revision \texttt{9a283b4}), operates in ControlArena's honest bash setting at temperature~0.7, top-$p$~1.0. Step $t$ uses seed $1000 \times \text{episode\_seed} + t$. Episodes terminate at submission, 15 steps (40 in the long-horizon log), 400,000 total tokens, or context exhaustion (32,768 tokens).

\paragraph{Serving.} Models are served with vLLM 0.30.0 \citep{kwon2023vllm} (PyTorch 2.13.0, CUDA 13.0) on one NVIDIA H100 80GB per host with seed~0, prefix caching off, and offline weights. Agent and monitor contexts are 32,768 and 8,192 tokens (55\% and 30\% GPU memory). The exploratory log, the exploratory live gates, the stronger-agent run, both airline runs, and the two-gate run served the agent with default kernels. The other runs of Table~\ref{tab:runs} used batch-invariant kernels (\texttt{VLLM\_BATCH\_INVARIANT=1}), except the model-update reruns, whose serving varies by rung. Each H100 host ran 32 episodes at a time. The model-update reruns replayed logged decisions on one NVIDIA GH200 120GB per job (85\% GPU memory), with 32 requests in flight under batch-invariant kernels and one under default kernels.

\begin{table*}[!t]
\centering
\footnotesize
\setlength{\tabcolsep}{3pt}
\renewcommand{\arraystretch}{1.08}
\begin{tabularx}{\textwidth}{@{}P{0.105\textwidth}P{0.095\textwidth}LP{0.165\textwidth}P{0.11\textwidth}@{}}
\rowcolor{hdrgray}
Change & Status & Harm (interval type) & Rests on & To settle \\
\charbadge{1} Stopping gate, $\theta=0.94$ & \cellcolor{mint!60}Identified; implementation check & P(harm): replay 0.039, live 0.030, equiv.\ within 0.03 (Newcombe 90\%). & (R), (A1), (A2); (A3) declared, harm untested & \cellcolor{coral!35}Test (A3) for harm \\
\charbadge{1} Stopping gate, replications & \cellcolor{mint!60}Identified; implementation check & Equiv.\ within 0.03 (Newcombe paired 90\%): stronger agent $[-0.021,0.009]$; airline $[-0.020,0.018]$. & (R), (A1), (A2); outputs vary across runs; (A3) declared & \cellcolor{coral!35}Test (A3) for harm; margin vs effect \\
\charbadge{1} Two retry gates, one log & \cellcolor{mint!60}Identified by targeted execution & P(harm): log $[0.244,0.345]$ and $[0.100,0.451]$; log + 1,032 and 2,929 snapshot continuations: 0.279 and 0.158; live 0.288 and 0.165, equiv.\ within 0.015 (Newcombe paired 90\% $[0.003,0.014]$ and $[0.001,0.013]$); pass-all and retry-as-stop outside the margin. Cost: 0.08 and 0.23 of live agent calls, given the log (partial). & (R), (A1), (A2); snapshot (T): 3,961 of 3,961 continuations valid & \cellcolor{coral!35}Check (T) beyond file tree \\
\charbadge{1} Retry gate, bash agent & \cellcolor{mint!60}Identified by targeted execution & P(harm): log $[0.023,0.422]$ (plug-in outer bounds); log + 485 snapshot continuations: 0.029 (exact 95\% $[0.020,0.040]$); live 0.030, equiv.\ within 0.02 (Newcombe paired 90\% $[-0.002,0.004]$). Cost: 0.30 of live agent calls, given the log (partial). & (R), (A1), (A2); snapshot (T): 485 of 485 restoration checks passed & \cellcolor{coral!35}Check (T) beyond file tree \\
\charbadge{2} Log without per-step harm & \cellcolor{mint!60}Set; identified with per-step harm & Certified P(harm) without per-step harm: held-out 0.159, replication 0.111; with it: 0.064, 0.051 (exact one-sided 95\%) & As stopping gate & \cellcolor{coral!35}Log harm timing \\
\charbadge{3} Both gates, 40-step cap & \cellcolor{hdrgray!60}Stop identified; retry bounded & P(harm): stopping 0.043 for horizons $k\geq15$ (unknowns harmful: 0.067); retry $[0.043,0.473]$ (plug-in outer bounds), agreement 0.77 ($k=1$) to 0.57 ($k=40$) & As both gates; 17 unknown outcomes (disk incident) & \cellcolor{coral!35}Execute retry continuations \\
\charbadge{4} Model release & \cellcolor{hdrgray!60}Agreement only, pooled & Agreement 0/99, share-weighted over six server processes (exact two-sided 95\% upper end 0.037, conservative); losses unestimated & Seed-addressable access per process & \cellcolor{coral!35}Support or new runs \\
\charbadge{5} Two-agent relay & \cellcolor{mint!60}Identified from snapshots & Second-agent P(harm): relay 0.115; snapshot equiv.\ within 0.03 (Newcombe paired 90\% $[-0.010,0.004]$); clean 0.178 (relay minus clean, Newcombe paired 95\% $[-0.078,-0.049]$). Both violate: 0.085; rate product 0.0195. & Snapshot (T): start digests 1,599 of 1,599 & \cellcolor{coral!35}Clean starts: workspace in $S$ \\
\charbadge{6} Announced gate & \cellcolor{hdrgray!60}Replay inapplicable & Abort $+0.070$ over the live unannounced gate (Newcombe paired 95\% $[0.019,0.120]$); P(harm): 10/400 live & (A2) fails for this prompt & \cellcolor{coral!35}Live runs \\
\end{tabularx}
\caption{\textbf{Table~\ref{tab:statements} in full: what the deployer can state before each change ships, and what would settle the rest.} Honest agents; in the bash setting, failure is near saturation and $\theta$ was chosen on an exploratory log; the stronger-agent and airline thresholds were chosen on pilots. Mint: identified, by the first two terms of display~\eqref{eq:split} or by targeted execution under (T). Gray: needs support or bounds, or replay does not apply. Coral: evidence that would settle the rest (last column). Newcombe: difference-of-proportions score interval; exact: Clopper-Pearson; plug-in: without sampling uncertainty; equiv.: equivalent. Failure, task success, and cost: Appendix~\ref{app:demo}.}
\label{tab:statements-full}
\end{table*}

\subsection{\texorpdfstring{\charbadge{1}~}{}Approval Gates: Stopping and Retry Gates}
\label{app:demo-results}\label{app:demo-gates}

This subsection gives the stopping and retry results of Section~\ref{sec:demo-stop}.

\begin{figure}[!ht]
\centering
\includegraphics[width=\columnwidth]{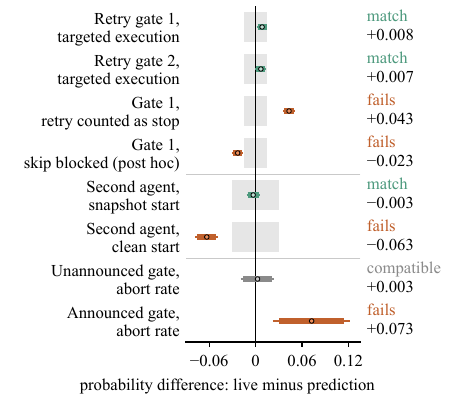}
\caption{\textbf{Targeted execution and snapshot starts match within registered margins; the unannounced gate stays compatible.} A thick 90\% Newcombe bar inside the gray band is a match (with registered cluster checks). Otherwise a thin 95\% whisker decides: excluding zero is a failure, and containing zero is compatible (unknown outcomes: midpoints and envelopes).}
\label{fig:forest}
\end{figure}

\begin{figure}[!ht]
\centering
\includegraphics[width=\columnwidth]{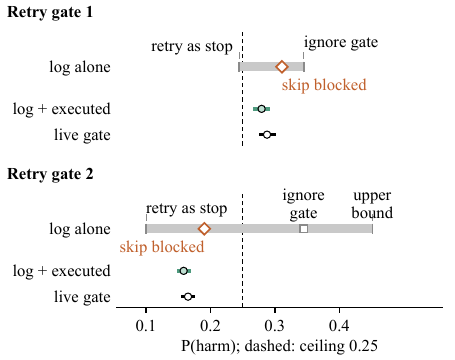}
\caption{\textbf{The log alone leaves each gate's P(harm) on both sides of the ceiling; executed continuations settle it and match live runs} (bars: exact 95\%). Gray: point estimates from retry as stop to the log-only upper bound; diamond: post hoc skip-blocked reading.}
\label{fig:bound}
\end{figure}

\begin{figure}[!ht]
\centering
\includegraphics[width=\columnwidth]{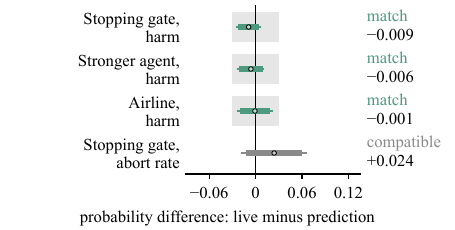}
\caption{\textbf{Implementation checks: stopping-gate replay matches live harm in all three settings.} The two ``Stopping gate'' rows use unpaired Newcombe intervals. Intervals, midpoint dots, and verdicts follow Figure~\ref{fig:forest}; the airline check also requires its task-cluster interval to fit the margin.}
\label{fig:checks}
\end{figure}

\paragraph{Stopping gate (Section~\ref{sec:demo-stop}).} Replay from the 1,000 episodes of the replication log predicts P(harm) 0.039, and the 1,000 live replication runs yield 0.030 (Figure~\ref{fig:checks}). The difference, $-0.009$ (Newcombe 95\% $[-0.025,0.007]$), falls within the registered equivalence margin of $\pm0.03$ (Newcombe 90\% $[-0.023,0.005]$).

\paragraph{Sensitivity to the abort valuation (A3).} On the replication log, 372 of 1,000 episodes stopped before harm, so if a stopped episode later caused harm with probability $p$, replayed P(harm) would be $0.039+0.372p$, below an illustrative 0.10 harm target only if $p<0.16$; (A3) is declared and untested for harm.

\paragraph{Stopping gate with a stronger agent.} The stronger-agent run tests Qwen3.5-27B-FP8 (up to 15 steps) on default kernels, so replay predicts the live gate in distribution. The gate reads the log-scale doubt $s=-\log_{10}(1-\mathrm{E}[N]/100)$ from the monitor's next-token distribution. On a pilot, threshold $\lambda=11.551$ was chosen where the stop rate fell between 10\% and 15\%; the registered run used 1,400 task-and-seed pairs. Live P(harm) is 0.224 to 0.226 against 0.230 to 0.233 for replay. Live minus replay is $-0.009$ to $-0.004$ (Newcombe paired 95\% $[-0.024,0.011]$; 90\% $[-0.021,0.009]$), equivalent within the registered margin of $\pm0.03$. With per-step harm, the log certifies P(harm) at most 0.253 (exact one-sided 95\%, rounded outward); without it, the identified set is $[0.189,0.331]$ and certifies only 0.353. Seven episodes across five pairs had unknown outcomes from a host disk incident (three unscored, four failed state checks); all seven remain in every denominator as ranges.

\paragraph{Stopping gate in a second environment.} The airline run uses $\tau^2$-bench v1.0.1 \citep{barres2025tau2bench}, where an agent modifies a customer booking database under written rules for cancellations, changes, fees, and compensation. A frozen deterministic checker flags policy violations. The agent (Qwen3-14B-FP8) and simulator run on default kernels. Threshold $\lambda=11.186$ was chosen on a pilot (stop rate 0.130), and the registered evaluation used 1,200 units across 50 tasks (24 seeds each). Live P(harm) is 554/1,200 = 0.4617 against 555/1,200 = 0.4625 for replay. Live minus replay is $-0.001$ (Newcombe paired 90\% $[-0.020,0.018]$; task-cluster bootstrap 90\% $[-0.023,0.020]$); both intervals satisfy the registered $\pm0.03$ margin. Pass-all P(harm) is 0.512; the gate removes 0.049 of harm in replay and 0.050 live, so the 0.03 margin is loose relative to this 0.05 effect. Because the database state is a deterministic function of executed tool calls, rerunning the checker on logged calls recovered harm timing in 1,200 of 1,200 episodes, certifying P(harm) at most 0.487 (without timing: 0.511).

\paragraph{Retry gate by targeted execution (Section~\ref{sec:demo-retry}).} Each of 1,200 units (a task and seed) runs three conditions on one host: a pass-all baseline committing a container snapshot prior to the first step scoring above $\theta=0.94$; a continuation from that snapshot with context rebuilt from logged messages, the proposal refused, and the live retry gate active; and the live retry gate from a clean sandbox. The retry gate agrees with 715 logged episodes and first blocks 485 (317 at step 0). From the log alone, display~\eqref{eq:bound} bounds P(harm) within $[0.023,0.422]$ (plug-in outer bounds). Executing the 485 continuations yields P(harm) of 35/1,200 = 0.029 (exact 95\% $[0.020,0.040]$), with no unknown outcomes; this prediction was frozen with its hash recorded before reading live records. The live gate gives 36/1,200 = 0.030 ($[0.021,0.041]$). Paired by unit, both arms are harmful in 35 units, the live arm alone in 1, and neither in 1,164. Live minus identified is 0.001 (Newcombe paired 95\% $[-0.003,0.005]$; 90\% $[-0.002,0.004]$), well within the registered $\pm0.02$ margin. Equivalence applies to marginal P(harm) over the task distribution; it does not establish that pathwise trajectories or task-level rates agree, as agreement and continuation errors could offset. Every restoration matched its source file-tree digest and passed context checks; no snapshot held outside processes. An independent recomputation from raw logs reproduced all values.

\paragraph{Execution cost.} The continuations required 4,004 agent calls and 26.1 million tokens, compared to 13,489 calls and 83.7 million tokens for the live gate: cost ratios of 0.30 and 0.31. These are partial-cost ratios that assume the pass-all log exists (14,460 agent calls).

\paragraph{Two retry gates from one log: protocol (Section~\ref{sec:demo-retry}; registered).}\label{app:demo-e7} The two-gate run (E7) logs the stronger agent (Qwen3.5-27B-FP8 on default kernels) under pass-all across 7,200 task-and-seed units on six hosts. A gate-free snapshot rule committed the container before every step whose monitor doubt score $s$ exceeded 1.0, yielding 16,456 committed snapshots (mean 61~MB per episode and 3.1~seconds per commit). Two retry gates block at $\lambda_1=11.551$ and $\lambda_2=10.25$ on $s$ with two retries permitted. For each gate, the log resolves agreeing units (6,168 and 4,271), while continuations from first-block snapshots were executed live (1,032 and 2,929, about 14\% and 41\% of the log; all 3,961 valid). Predictions were frozen, with their hash recorded, before reading live records, and an independent recomputation verified all identified rows, intervals, and decisions.

For Gate~2, the evaluation follows Section~\ref{sec:results}. The deployer declares the gate, stop value, and decision rule. Replay finds 4,271 agreeing episodes and 2,929 first blocks (41\%). Of these, 92 fall at the last step, where no agent step remains, so their recorded outcome is final (69 harmless, 23 harmed before the block); the other 2,837 continue. The log reads 387 harmed agreeing episodes, and executing the continuations yields 730 more. Display~\eqref{eq:split} gives $(387+23+730)/7{,}200\approx0.158$ (with the one unknown outcome, the midpoint is 0.1584). Joint 95\% bounds put harm at most 0.1672 and success at least 0.3871, approving the gate.

\paragraph{E7 primary results.} For Gate 1, identified P(harm) is 0.2794 against live $2{,}072/7{,}200=0.2878$ (one unknown outcome), a difference of 0.008 (Newcombe paired 90\% $[0.003,0.014]$; task-cluster 90\% $[0.002,0.015]$), equivalent within the registered 0.015 margin. For Gate 2, identified P(harm) is 0.1584 against live 0.1652 (one unknown each), a difference of 0.007 (paired 90\% $[0.001,0.013]$; task-cluster $[0.000,0.013]$), also equivalent within 0.015. Across all four targeted-execution comparisons, live harm exceeds identified harm by 0.001 to 0.008, all within registered margins. Log-only outer bounds are $[0.244,0.345]$ and $[0.100,0.451]$, corrected for the unknown outcome. Both ends are attainable: an agent that halts after a refusal gives the lower end, and one that violates the policy whenever a refusal leaves it a step gives the upper end. The log contains no refusal and rules out neither, so it cannot decide either gate.

\paragraph{E7 reference predictors.} Both registered reference predictors fail the 0.015 margin on both gates. Ignoring the gate (pass-all, 0.3445) misses live harm by $-0.057$ and $-0.179$. Counting each retry as a stop (0.2445 and 0.1002) misses by $+0.043$ and $+0.065$ (Newcombe paired 95\% $[0.036,0.051]$ and $[0.058,0.072]$). A third reading, fixed in a note before computation (post hoc), drops blocked calls while retaining the remaining logged episode and yields 0.3108 and 0.1904 (discrepancies $-0.023$, 95\% $[-0.030,-0.016]$, and $-0.025$, $[-0.033,-0.018]$), also outside the margin; under the declared rule, it would select gate 2 with harm at most 0.1984. The registered reference errors round to 0.04 to 0.18, as reported in Section~\ref{sec:demo-retry}.

\paragraph{E7 decision and execution cost.} The declared decision rule (P(harm) at most 0.25 and task success at least 0.37, exact one-sided 95\% bounds, unknowns unfavorable) selects Gate 2 in both arms: identified harm at most 0.1658 and success at least 0.3889; live harm at most 0.1727 and success at least 0.3897 (also passing with 97.5\% per bound, for joint 95\% coverage: 0.1672 and 0.3871). Neither registered reference certifies any candidate (post hoc): pass-all violates the harm ceiling, and retry-as-stop violates the harm ceiling for gate 1 and the success floor for gate 2. Continuations used 0.08 and 0.23 of live agent calls (8\% and 23\%; partial-cost records); including the pass-all log's 102,779 calls, the ratio is 0.66. With the log already recorded, the continuations' 31,706 calls are 15\% of the 204,875 of both live gates. Random live subsets of that expected cost (1,114 units per gate) reached both live decisions in 56.4\% of 500 draws; the smallest common size on a 1\% grid reaching both in at least 90\% was 2,232 units per gate (31\% of the cohort; 91.8\%), at 2.0 times the continuations' calls and 0.47 times those of log plus continuations (post hoc; per-bound 95\%).

\paragraph{Simpler readings at face value (post hoc).} Read as point values without coverage at the declared ceiling and floor, counting retries as stops approves gate 1, which live runs refuse, and refuses gate 2, whose success it puts at 0.334; ignoring the gate also refuses gate 2.

\paragraph{E7 snapshot design.} Minimal first-deviation snapshots at open running maxima after step 0 require 6,332 of the 16,456 snapshots and 173 of 441~GB for harm alone (Proposition~\ref{prop:snapshot-design}, post hoc). A running maximum at the last step is not open because no agent step remains.

\paragraph{Retry gate in the airline environment.} The registered airline run (E5) restores each continuation by replaying logged database writes without container snapshots. The pass-all log plus 159 valid executed continuations identify P(harm) of 597/1,200 = 0.4975; the paired live retry gate gives 598/1,200 = 0.4983. Both 90\% equivalence intervals (Newcombe paired $[-0.019,0.020]$; task-cluster bootstrap $[-0.017,0.019]$) lie inside $\pm0.03$. The prediction was frozen before the 800 new live records were read; units 0 to 399 reuse the airline run's live retry episodes, already analyzed there. Counting retries as stops gives the lower bound 0.4625 (0.036 below live). Continuations used 0.090 of live agent calls and 0.131 of declared tokens (complete usage).

\paragraph{Retry gate on the held-out log.} In the earlier held-out protocol, later-prefix restoration checks passed 0 of 10 trials, so those continuations stayed bounded; step-0 runs needed only fresh starts (bounds and live check: companion file).

\subsection{\texorpdfstring{\charbadge{2}~}{}Partial Records: Certification}
\label{app:demo-records}

For the logging comparison in Section~\ref{sec:demo-log}, exact one-sided 95\% Clopper--Pearson bounds on replayed P(harm) certify a harm level with reduced logs in this setting. On the replication log (no unknown outcomes), full and reduced logs certify 0.051 (39/1,000 harms) against 0.111 without per-step harm (94/1,000). On the held-out log, counting one failed state check as harmful yields 0.064 for full and reduced logs (12/300) against 0.159 without per-step harm (37/300). The reduced log preserves per-step scores, per-step harm and cost, and final outcomes.

\begin{table}[!ht]
\centering
\footnotesize
\setlength{\tabcolsep}{2.5pt}
\renewcommand{\arraystretch}{1.05}
\begin{tabularx}{\columnwidth}{@{}P{0.17\columnwidth}P{0.16\columnwidth}P{0.27\columnwidth}L@{}}
\rowcolor{hdrgray}
Field & Needed for & Droppable when & Evidence \\
Monitor scores & Deviation & Monitor reruns deterministically & Held-out: 3,589 of 3,589 recomputed \\
\rowcolor{mint!60}
Per-step harm & Stop valuation & Calls and a deterministic checker fix timing & Certified ceiling with vs.\ without it: held-out 0.064 vs.\ 0.159, replication 0.051 vs.\ 0.111; airline: 1,200 of 1,200 recomputed \\
Per-step cost & Pre-stop cost & Another field fixes it & Mean tokens, exploratory gates (companion) \\
Inputs, seeds, serving mode & Agent rerun & Change reads logged fields only & 1,650 of 1,650 decisions; other kernel mode: 2 of 99 episodes \\
Gate decisions & Nothing & Always (Remark~\ref{rem:app-status}) & Unused \\
\end{tabularx}
\caption{\textbf{What a stopping-gate replay needs from the log here: per-step harm (shaded) is the one field whose removal raised the certified harm ceiling.} An instance of Proposition~\ref{thm:logging}; no general field list. Certification: exact one-sided 95\% bounds on replayed P(harm).}
\label{tab:log-fields}
\end{table}

\subsection{\texorpdfstring{\charbadge{3}~}{}Long Horizons}
\label{app:demo-horizon}

For the horizon comparison in Section~\ref{sec:demo-horizon}, truncating a 600-episode pass-all log with a 40-step cap under batch-invariant serving yields horizon curves with all 600 planned episodes in every denominator (the companion file holds the per-horizon table). Beyond $k=25$, more than 5\% of runs have already ended on token or context limits (151 by $k=40$, while 240 reach the 40-step cap). Stopping-gate replayed P(harm) stabilizes at 26/600 = 0.043 for $k\ge15$ (0.067 if all unknowns were harmful). Retry-gate agreement mass falls from 0.77 at $k=1$ to 0.57 at $k=40$, widening plug-in outer bounds from $[0.018,0.030]$ to $[0.043,0.473]$. A host disk incident caused 17 unknown outcomes (6 unscored, 11 failed state checks), which generate all reported ranges; no episode was excluded.

\subsection{\texorpdfstring{\charbadge{4}~}{}Model Releases}
\label{app:demo-release}

In the model-release test of Section~\ref{sec:demo-release}, the release candidate (Qwen3.5-27B-FP8) reproduces none of 99 logged episodes across six server processes at identical inputs and seeds, while the identical model reproduces 99 of 99 and kernel and precision changes reproduce 2 and 1. Outputs were seed-addressable within each process (40 of 40 probe inputs reproduced) but varied across processes (14 of 40 differed). Episodes were assigned to processes independently of content, so the bound targets the mean agreement mass over processes, weighted by their shares; the exact two-sided 95\% Clopper--Pearson bound for 0/99 (0.037) is conservative for it, by independence across episodes and concavity of $\log(1-p)$.

\subsection{\texorpdfstring{\charbadge{5}~}{}Shared State: Two Agents in Relay}
\label{app:demo-relay}

In the registered five-condition design of Section~\ref{sec:demo-relay}, a first agent works a task up to 7 steps, and a second agent continues for up to 8 steps in the inherited workspace. Across 1,600 units (8,000 planned runs, 7,998 scored; two snapshot starts failed to commit and count as unknown), the second agent violated policy in 184 of 1,600 relays (0.115, exact 95\% $[0.100,0.132]$). From a clean start on the same task and seed, it violated in 285 of 1,600 (0.178, $[0.160,0.198]$). Relay minus clean start is $-0.063$ (Newcombe paired 95\% $[-0.078,-0.049]$; 90\% $[-0.075,-0.051]$), failing equivalence within the registered $\pm0.03$ margin. Started from a snapshot of the first agent's standalone end state, violations occurred in 0.1175 to 0.1188 of runs; relay minus snapshot start is $-0.004$ to $-0.003$ (paired 90\% $[-0.010,0.004]$), equivalent within $\pm0.03$. Here clean starts substantially overstate relay harm.

Both agents violated in 136 of 1,600 relays ($\beta=0.085$, exact 95\% $[0.072,0.100]$), against marginal rates $p_1=0.169$ and $p_2=0.115$ whose product is 0.0195. The difference $\beta - p_1 p_2 = 0.066$ shows strong positive dependence; treating the agents as independent overstates the union harm probability $\pi=p_1+p_2-\beta$ by 0.066. Per-agent logs without shared state give the Proposition~\ref{prop:component-joint}(ii) range $[0,0.115]$ for $\beta$; conditioning on the task risk class $C$ narrows it to $[0.031,0.115]$, and adding external write flags $D$ gives $[0.041,0.115]$. Linking snapshot runs by unit yields 0.084 to 0.085, matching relay within $\pm0.03$.

At initialization, file-tree digests matched their sources in 1,599 of 1,599 runs per snapshot condition.

\subsection{\texorpdfstring{\charbadge{6}~}{}Oversight Cues}
\label{app:demo-cues}

In the paired design of Section~\ref{sec:demo-cues} over 400 triplets (pass-all, unannounced stopping gate, announced stopping gate), the announced gate aborted 0.070 more often than the unannounced gate (181/400 versus 153/400; Newcombe paired 95\% $[0.019,0.120]$; 90\% $[0.027,0.112]$, excluding equivalence within $\pm0.03$). Replay from unannounced logs understates the announced abort rate by 0.0725 (Newcombe paired 95\% $[0.022,0.122]$), while the unannounced gate stays compatible with replay (negative control). Harm is 10/400 in all three arms. This intervention alters the live abort rate, rejecting the abort-rate invariance of (A2) for this prompt; it does not isolate semantic anticipation from other prompt perturbations.

\section{Replay, Logging Sufficiency, and Long Horizons}
\label{app:replay}

For Sections~\ref{sec:replay} to~\ref{sec:horizons}, this appendix proves Propositions~\ref{prop:replay}, \ref{prop:replay-cont}, and~\ref{thm:logging}, Lemma~\ref{lem:app-unrestricted}, and Corollaries~\ref{cor:garbling} and~\ref{cor:horizon} in the finite setting.

\subsection{Continuing Deviations: Statement}
\label{app:replay-cont}

Proposition~\ref{prop:replay} gives the agreement and stopping components of display~\eqref{eq:split} in Section~\ref{sec:results}; this subsection bounds the continuing component.
For a scalar query $X=\phi(L)$ and continuing replayed prefix $h$, let $\mathcal X_h$ denote the values of $X$ on paths extending $h$ that follow $f^\kappa$ at later decisions.
Condition~(S) formalizes saturated continuation, with primitive conditions in Lemma~\ref{lem:app-unrestricted}.
\begin{itemize}
\item[(S)] \emph{Saturated continuation}: distinct continuing replayed prefixes of positive baseline mass reach distinct inputs of unknown kernels, each of baseline probability zero. For every family $(Q_h)$ with $Q_h\in\Delta(\mathcal X_h)$, one admissible mechanism with the same baseline law realizes all $Q_h$ at once. A replacement that sends two or more continuing prefixes of positive baseline mass to one common state violates (S).
\end{itemize}

\begin{proposition}[Continuing deviations; \citealp{manski2003partial}]\label{prop:replay-cont}
Assume \textup{(R)}, \textup{(A1)}, and \textup{(A2)}, and let $B\subseteq\lossset$.
Suppose every coordinate of $\Omega$ is recorded, so the evidence is $P^0$. Let $\nu$ be the subprobability from \textup{(i)} and \textup{(ii)} of Proposition~\ref{prop:replay} pushed to $X$. Under \textup{(A3)} and \textup{(S)}, the identified set of laws of $X$ is $\{\nu+\int Q_h\,P^0(dh,\text{continuing}\mid r):Q_h\in\Delta(\mathcal X_h)\}$. Every law-invariant functional nondecreasing in first-order stochastic dominance (mean, exceedance probability, VaR, CVaR) attains its minimum and maximum over this set when each continuing prefix places its mass at $\min\mathcal X_h$ or at $\max\mathcal X_h$, respectively. For the mean, exceedance probability, and CVaR, every value between the extremes is attained; VaR takes values in a finite set. Without \textup{(S)}, these are outer bounds.
\end{proposition}

Let $\nu^-$ and $\nu^+$ be the laws of $X$ given $r$ that keep agreement and stop values and assign each continuing prefix its least and greatest feasible value, respectively. Every such functional $\rho$ then satisfies
\begin{equation}
\rho(\nu^-)\le q^\kappa_t(r)\le\rho(\nu^+),
\label{eq:bound}
\end{equation}
with the inequalities holding without (S).
Section~\ref{app:replay-saturated} contains the proof.

\subsection{Finite Setting}
\label{app:replay-setting}

This setting is used in the replay, logging, and horizon proofs for Sections~\ref{sec:replay} to~\ref{sec:horizons}.

\paragraph{Paths, Histories, and Kernels.}
Let $\Omega$ be a finite set of complete paths $\omega=(v_1,\dots,v_{n(\omega)})$ listing execution, observation, and post-abort coordinates in declared generation order, with no path a proper prefix of another.
Histories $\mathcal H$ are proper prefixes, including the empty history.
Declared order assigns to each $h\in\mathcal H$ a kernel name $\mathrm{nm}(h)\in\mathcal N$ and a declared input $\pi(h)$.
The tree is full: $(h,v)\in\mathcal H\cup\Omega$ for every kernel value $v$.
A mechanism $M$ assigns transition kernels $k^M_n(\cdot\mid i)$.
For $\omega_{1:j}=(v_1,\dots,v_j)$, with $\omega_{1:0}$ empty, the path law is
\begin{equation}
P_M(\omega)=\prod_{j=0}^{n(\omega)-1}k^M_{\mathrm{nm}(\omega_{1:j})}\bigl(v_{j+1}\mid\pi(\omega_{1:j})\bigr).
\label{eq:app-path-law}
\end{equation}
The cylinder $[h]$ collects paths extending $h$.
Summing over extensions of $h$ gives
\begin{equation}
P_M([h])=\prod_{j=0}^{|h|-1}k^M_{\mathrm{nm}(h_{1:j})}\bigl(h_{j+1}\mid\pi(h_{1:j})\bigr).
\label{eq:app-cylinder}
\end{equation}
Loss is $L=\ell(\omega)\in\lossset\subset\mathbb R^m$.
Observation kernels are named kernels, so records $R_t$ are path coordinates, realizing $P_{E,K}$ of Appendix~\ref{app:formal-spec} in the finite setting.

\paragraph{Replaced Kernel.}
The replaced decision kernel is named $k\in\mathcal N$.
A history $h$ with $\mathrm{nm}(h)=k$ is a decision history with input $I(h)=\pi(h)$ and step $t(h)\in\{0,\dots,T\}$.
The refined clock of Appendix~\ref{app:formal-spec} invokes $k$ at most once per step.
Under (R), known maps define the baseline kernel $\delta_{f^0(i)}$ and replacement kernel $\delta_{f^\kappa(i)}$.
The input is a function of history contained in the full record, distinguishing full histories from declared inputs.
Baseline laws are supported on $\Omega^0$, where decision coordinates match $f^0$.
A value $a$ is an abort if no extension of $(h,a)$ has $\mathrm{nm}(h')=k$, terminating calls to the replaced kernel.

\paragraph{First Deviation.}
A decision history $h$ is a deviation history if $f^\kappa(I(h))\neq f^0(I(h))$ while $f^\kappa(I(h'))=f^0(I(h'))$ on all proper decision prefixes $h'$.
The set $\mathcal D_\kappa$ of deviation histories has pairwise disjoint cylinders.
For $\omega\in[h]$ with $h\in\mathcal D_\kappa$, set $\sigma_\kappa(\omega)=t(h)$ and replayed prefix $h^+=(h,f^\kappa(I(h)))$; set $\sigma_\kappa=\infty$ on agreement set $\mathsf A_\kappa$.
Paths partition into agreement and disjoint deviation cylinders:
\begin{equation}
\Omega=\mathsf A_\kappa\sqcup\bigsqcup_{h\in\mathcal D_\kappa}[h].
\label{eq:app-partition}
\end{equation}
Deviations partition into terminal set $\mathcal D^{\mathrm{term}}_\kappa$ (abort outputs) and continuing set $\mathcal D^{\mathrm{cont}}_\kappa$.

\paragraph{Conditioning Record.}
A record value $r$ is pre-intervention if $\Omega_r=\{R=r\}$ is a disjoint union of cylinders $[h]$ over $\mathcal H_r$, none having a deviation history as a proper prefix.
Records $R_t$ with $t<s$ satisfy this property.
We assume $P^0(\Omega_r)>0$ and write $P(\cdot\mid r)=P(\cdot\cap\Omega_r)/P(\Omega_r)$.
Let $\mathcal D_\kappa(r)=\{h\in\mathcal D_\kappa:[h]\subseteq\Omega_r\}$, with $\mathcal D^{\mathrm{term}}_\kappa(r)$ and $\mathcal D^{\mathrm{cont}}_\kappa(r)$ defined analogously.

\paragraph{Assumptions.}
Let $P^0=P_{M^0}$ and $P^\kappa=P_{M^\kappa}$.
The assumptions of Appendices~\ref{app:formal-question} and~\ref{app:formal-replay} take tree form:
\begin{itemize}
\item[(R)] Replay interface: the replaced decision kernel is $\delta_{f^0(i)}$ under $M^0$ and $\delta_{f^\kappa(i)}$ under $M^\kappa$, with both maps known and the input contained in the full record (a coarser log, as in Proposition~\ref{thm:logging}, need not contain it).
\item[(A1)] Invariance: $k^{M^\kappa}_n=k^{M^0}_n$ for every name $n\neq k$, including the kernel at the empty history (the initial law) and every random-input kernel.
\item[(A2)] Prefix preservation (no anticipation): both mechanisms use the same declared order and inputs $(\mathrm{nm},\pi)$, and no declared input or unreplaced kernel depends on the identity or specification of the installed replacement.
\item[(A3)] Declared abort valuation: for every $h\in\mathcal D^{\mathrm{term}}_\kappa$ with $P^0([h])>0$, $P^\kappa(L\in\cdot\mid[h])=b_\kappa(\cdot\mid h^+)$, a declared probability measure on $\lossset$.
\end{itemize}
Because no replaced kernel acts after an abort, (A1) ensures that post-abort distributions depend only on $h^+$, shared by replacements that abort at the same replayed prefix.

\subsection{Prefix Lemma and Proof of \texorpdfstring{Proposition~\ref{prop:replay}}{the Deviation-Replay Proposition}}

\begin{lemma}[Prefix Lemma]\label{lem:app-prefix}
Assume (R), (A1), and (A2).
\textup{(a)}~$P^\kappa(\omega)=P^0(\omega)$ for every $\omega\in\mathsf A_\kappa$.
\textup{(b)}~$P^\kappa([h])=P^0([h])$ for every history $h$ none of whose proper prefixes is a deviation history.
In particular this holds for every $h\in\mathcal D_\kappa$ and every $h\in\mathcal H_r$, and $P^\kappa(\Omega_r)=P^0(\Omega_r)$.
\end{lemma}
\begin{proof}
By (R), (A1), and (A2), corresponding factors in \eqref{eq:app-path-law} and \eqref{eq:app-cylinder} agree until the first deviation, proving (a) and (b).
Summing over the disjoint record cylinders gives equality on $\Omega_r$.
\end{proof}

\begin{proof}[Proof of Proposition~\ref{prop:replay}(i)]
Immediate from Lemma~\ref{lem:app-prefix}(a),(b), summing over agreement paths and conditioning on $r$.
\end{proof}

\begin{proof}[Proof of Proposition~\ref{prop:replay}(ii)]
Immediate from Lemma~\ref{lem:app-prefix}(b) and (A3), summing over the disjoint terminal cylinders and conditioning on $r$.
\end{proof}

\subsection{Saturated Continuation and Proof of \texorpdfstring{Proposition~\ref{prop:replay-cont}}{the Continuing-Deviation Proposition}}
\label{app:replay-saturated}

Fix a finite query $\phi:\lossset\to\mathcal X$, with $\mathcal X\subset\mathbb R$ for order statements.
For $h\in\mathcal D_\kappa$, let $\Omega^\kappa(h^+)$ collect replacement-consistent extensions that follow $f^\kappa$ at all decisions after $h^+$, giving feasible set $\mathcal X_h=\{\phi(\ell(\omega)):\omega\in\Omega^\kappa(h^+)\}$.
The model class $\mathcal M$ is a family of mechanisms that share the tree, $\ell$, $f^0$, $f^\kappa$, and $b_\kappa$, and satisfy (R) and (A1) to (A3).
Let $\mathcal M(P)$ be members with baseline law $P$, with identified set $\Theta_\phi(P)=\{\phi_\#P^\kappa_M(L\in\cdot\mid r):M\in\mathcal M(P)\}$.

\begin{definition}[Saturated Continuation]\label{def:app-saturated}
Let $\mathcal D^{\mathrm{cont}}_\kappa(r,P)$ be the set of $h\in\mathcal D^{\mathrm{cont}}_\kappa(r)$ with $P([h])>0$.
The class $\mathcal M$ is saturated at $P$ for $\phi$ if, for every family $(Q_h)$ with $Q_h\in\Delta(\mathcal X_h)$ for $h\in\mathcal D^{\mathrm{cont}}_\kappa(r,P)$, some $M\in\mathcal M(P)$ satisfies $\phi_\#P^\kappa_M(L\in\cdot\mid[h])=Q_h$ for every such $h$.
\end{definition}

Let $\mathcal P$ be the set of laws $P$ on $\Omega^0$ with $P(\Omega_r)>0$.

\begin{lemma}[Unrestricted Class]\label{lem:app-unrestricted}
Let $\mathcal M_{\mathrm{full}}$ be the class of mechanisms on the declared tree that satisfy (R) and (A1) to (A3) and in which every unreplaced kernel reads the full history, $\pi(h)=h$ whenever $\mathrm{nm}(h)\neq k$.
Suppose that for every $h\in\mathcal D^{\mathrm{term}}_\kappa$ the declared $b_\kappa(\cdot\mid h^+)$ is the law of $\ell$ under some probability on the replacement-consistent extensions of $h^+$.
Then for every $P\in\mathcal P$, every $\phi$, and every family $(Q_h)$ with $Q_h\in\Delta(\mathcal X_h)$ for $h\in\mathcal D^{\mathrm{cont}}_\kappa(r,P)$, some $M\in\mathcal M_{\mathrm{full}}$ has baseline law $P$ and satisfies $\phi_\#P^\kappa_M(L\in\cdot\mid[h])=Q_h$ for every such $h$.
In particular, every law in $\mathcal P$ is the baseline law of a member, and $\mathcal M_{\mathrm{full}}$ is saturated at every $P\in\mathcal P$.
\end{lemma}
\begin{proof}[Proof sketch]
For each continuing prefix $h$, mix replacement-consistent paths achieving each outcome in $\mathcal X_h$ with weights $Q_h$; at terminal prefixes, use extensions realizing $b_\kappa$.
Set unreplaced kernels to the conditional transitions of $P$ on its support and of these mixtures inside each $[h^+]$.
Full-history inputs allow these assignments independently: the replayed cylinders are disjoint and baseline-null, since each contains a first output differing from $f^0$.
Complete unused inputs arbitrarily.
The product path law then recovers $P$ and realizes all target continuation laws and terminal valuations simultaneously.
\end{proof}

Assumption~(S) is Definition~\ref{def:app-saturated} at $P^0$ with distinct continuing prefixes reaching distinct unreached inputs; under realizable $b_\kappa$, $\mathcal M_{\mathrm{full}}$ is saturated at every $P\in\mathcal P$.

\begin{proposition}[Identified Set]\label{prop:app-identified}
Assume (R) and (A1) to (A3), and let $P$ be a baseline law with $P(\Omega_r)>0$.
Write $w_h=P([h]\mid r)$ for $h\in\mathcal D^{\mathrm{cont}}_\kappa(r,P)$, and let
\[
\begin{aligned}
\nu=&\phi_\#\Bigl(P(L\in\cdot,\mathsf A_\kappa\mid r)\\
&+\sum_{h\in\mathcal D^{\mathrm{term}}_\kappa(r)}P([h]\mid r)\,b_\kappa(\cdot\mid h^+)\Bigr).
\end{aligned}
\]
Then $\Theta_\phi(P)$ is contained in
$\mathcal S_\phi(P)=\bigl\{\nu+\sum_hw_hQ_h:Q_h\in\Delta(\mathcal X_h)\bigr\}$,
with equality if $\mathcal M$ is saturated at $P$.
\end{proposition}
\begin{proof}
Summing over \eqref{eq:app-partition}, Proposition~\ref{prop:replay}(i),(ii) fixes $\nu$, while Lemma~\ref{lem:app-prefix}(b) fixes each continuing weight $w_h$.
By (R), each continuation law lies in $\Delta(\mathcal X_h)$, giving the inclusion; saturation gives equality by Definition~\ref{def:app-saturated}.
\end{proof}

\paragraph{Order Bounds.}
For finitely supported laws $\eta,\eta'$ on $\mathbb R$, write $\eta\le_{\mathrm{st}}\eta'$ if $F_\eta(x)=\eta((-\infty,x])\ge F_{\eta'}(x)$ for all $x$ \citep{shaked2007stochastic}.
Let $\eta^-=\nu+\sum_hw_h\delta_{\min\mathcal X_h}$ and $\eta^+=\nu+\sum_hw_h\delta_{\max\mathcal X_h}$.

\begin{proposition}[Order Bounds]\label{prop:app-order}
Under the hypotheses of Proposition~\ref{prop:app-identified}, with $\mathcal X\subset\mathbb R$, the following hold.
\textup{(a)}~$\eta^-,\eta^+\in\mathcal S_\phi(P)$, and $\eta^-\le_{\mathrm{st}}\eta\le_{\mathrm{st}}\eta^+$ for every $\eta\in\mathcal S_\phi(P)$.
\textup{(b)}~If $\rho$ is law invariant and nondecreasing in $\le_{\mathrm{st}}$, then $\rho(\mathcal S_\phi(P))\subseteq[\rho(\eta^-),\rho(\eta^+)]$ and both endpoints are attained; for nonincreasing $\rho$ the endpoints swap.
The mean, the exceedance probability $\eta((x,\infty))$, the CDF value $F_\eta(x)$, which is nonincreasing, $\mathrm{VaR}_\beta(\eta)=\min\{x:F_\eta(x)\ge\beta\}$, and $\mathrm{CVaR}_\beta$ for $\beta\in(0,1)$ are such functionals.
\textup{(c)}~For the mean, the exceedance probability, the CDF value, and $\mathrm{CVaR}_\beta$, every value between the two endpoints is attained.
\end{proposition}
\begin{proof}[Proof sketch]
Point masses at feasible endpoints bound each continuation CDF, proving (a); CDF and quantile ordering give (b) \citep[Eqs.~(1.A.2), (1.A.7)]{shaked2007stochastic}.
For (c), mix $\eta^-$ and $\eta^+$ within $\mathcal S_\phi(P)$. Means and event probabilities are affine; finite support makes the CVaR formula \citep[author's version, Theorem~10, Eqs.~(27)--(28)]{rockafellar2002cvar} a finite minimum of affine functions and also gives its monotonicity in (b). Continuity and the intermediate value theorem finish.
\end{proof}

\begin{proof}[Proof of Proposition~\ref{prop:replay-cont}]
Combine Propositions~\ref{prop:app-identified} and~\ref{prop:app-order}, using full recording and (S) for equality; VaR takes values in the finite feasible set.
\end{proof}

\subsection{Logging Characterization and Proof of \texorpdfstring{Proposition~\ref{thm:logging}}{the Logging Proposition}}
\label{app:logging-scope}

\paragraph{Scope of the Necessity Direction.}
Necessity in Proposition~\ref{thm:logging} requires variation across all baseline laws on $\Omega^0$ via point-mass witnesses.
In restricted classes where coordinates have fixed laws, necessity can fail, whereas sufficiency holds for every baseline class.

\paragraph{Proof.}
In the tree setting, the contribution $C_\kappa$ of Proposition~\ref{thm:logging} is $\delta_{\ell(\omega)}$ on $\mathsf A_\kappa$, $b_\kappa(\cdot\mid h^+)$ on $[h]$ for terminal $h$, and $0$ on $[h]$ for continuing $h$; $\mathcal P_r$ is the set $\mathcal P$ above, $s_\kappa=s_{\kappa,r}$, and $\Omega^0_r=\Omega^0\cap\Omega_r$.
For a baseline law, $s_\kappa(P)=(\nu_\kappa,\varepsilon_\kappa)$, where $\nu_\kappa$ is the measure $\nu$ of Proposition~\ref{prop:app-identified} with $\phi$ the identity and $\varepsilon_\kappa=\sum_hw_h$.
A logging map is a function $g$ on $\Omega$, and it determines the record event when $\Omega_r=g^{-1}(Y_r)$ for a set $Y_r$ of log values.

\begin{proof}[Proof of Proposition~\ref{thm:logging}]
Fix $\kappa$ and assume $\Omega_r=g^{-1}(Y_r)$.
For sufficiency, fibre constancy makes $(C_\kappa,u_\kappa)$ a function of $g$ on $\Omega^0_r$, so its conditional expectation is determined by $g_\#P$.
For necessity, suppose $\omega,\omega'\in\Omega^0_r$ satisfy $g(\omega)=g(\omega')$ and $(C_\kappa,u_\kappa)(\omega)\neq(C_\kappa,u_\kappa)(\omega')$.
The point masses $\delta_\omega$ and $\delta_{\omega'}$ belong to $\mathcal P$ and have the same log law.
Their summaries are $(C_\kappa,u_\kappa)(\omega)$ and $(C_\kappa,u_\kappa)(\omega')$, which differ, so no function of $g_\#P$ equals $s_\kappa(P)$ on $\mathcal P$.
Applying both directions to each $\kappa$ in the family proves the equivalence.
\end{proof}

\begin{remark}[Status Need Not Be Logged]\label{rem:app-status}
A coarsened log need not record deviation status when its fibres pool paths with equal signatures (check T1-4, \texttt{checks/verify\_replay.py}). Condition (R) refers to the full record, which contains the gate input; the log need not.
\end{remark}

\subsection{Coarse Evidence and Garbling}

This corollary supports the coarse-log claim of Section~\ref{sec:logging}.

\begin{corollary}[Garbling]\label{cor:garbling}
For a fixed model class, replacement, and query conditioned on one record event measurable from $g'$, if $g'=G\circ g$ for some map $G$, the identified set under $g$ is contained in the set under $g'$.
\end{corollary}
\begin{proof}[Proof sketch]
Immediate from nesting the compatible model classes under $g$ and $G\circ g$ \citep{blackwell1951comparison,blackwell1953equivalent}.
\end{proof}

\subsection{Long Horizons}
\label{app:horizon}

This proves the horizon claims of Section~\ref{sec:horizons}.

\begin{corollary}[Long horizons]\label{cor:horizon}
Assume \textup{(R)}, \textup{(A1)}, and \textup{(A2)}. For $j=0,\dots,T$ let $\eta_j(r)=P^0(\sigma_\kappa=j\mid\sigma_\kappa\ge j,\ R_t=r)$, set to $0$ when $P^0(\sigma_\kappa\ge j\mid r)=0$; by Lemma~\ref{lem:app-prefix}(b) the same value holds under $P^\kappa$. Fix $\eta\in[0,1]$.
\textup{(i)} The agreement mass $P^\kappa(\mathsf A_\kappa\mid r)=\prod_{j=0}^{T}(1-\eta_j(r))$ is identified from the full record; if $\eta_j(r)\ge\eta$ at $k$ steps, it is at most $(1-\eta)^k$.
\textup{(ii)} If every deviation is terminal, \textup{(A3)} holds, and a logging map $g$ that determines the event $\{R_t=r\}$, as in Proposition~\ref{thm:logging}, determines the contribution $C_\kappa$ on $\Omega^0\cap\{R_t=r\}$, then Proposition~\ref{prop:replay}\textup{(i)} and \textup{(ii)} and Proposition~\ref{thm:logging} identify the whole law of $L$ given $r$, at every horizon.
\textup{(iii)} If every deviation is continuing, then under the hypotheses of Proposition~\ref{prop:replay-cont} the identified interval of the mean of $X$ has width $\int(\max\mathcal X_h-\min\mathcal X_h)\,P^0(dh,\text{continuing}\mid r)$. When every $\mathcal X_h$ has spread at least $\delta$, the width is at least $\delta\bigl[1-\prod_{j}(1-\eta_j(r))\bigr]$.
\end{corollary}
\begin{proof}[Proof sketch]
Part (i) follows from the chain rule and Lemma~\ref{lem:app-prefix}; the bound follows by comparing factors.
For (ii), apply Propositions~\ref{prop:replay} and~\ref{thm:logging}, since agreement and terminal deviations exhaust the mass.
For (iii), subtract the mean endpoints in Proposition~\ref{prop:app-order}; the lower bound uses continuing mass equal to one minus the agreement mass in (i).
\end{proof}

\section{Proofs: Supported Continuation, Unsupported Inputs, and Model Updates}
\label{app:supported}

This appendix proves the continuation results of Section~\ref{sec:supported}: Proposition~\ref{prop:supported}, Proposition~\ref{prop:r4-unsupported}, and Corollary~\ref{cor:model-update} in a finite coordinate model, using the sequential g-formula \citep{robins1986new,pearl1995sequential} and partial identification \citep{manski2003partial}. It also proves that section's execution-budget and snapshot-design results.

\subsection{Finite Coordinate Model}
\label{app:supported-model}

This is the model for the proof of Proposition~\ref{prop:supported}, summarized in Section~\ref{sec:supported}.

Coordinates in generation order are $V_1,\dots,V_n$ with finite ranges $\mathcal V_1,\dots,\mathcal V_n$, and $\Omega=\prod_i\mathcal V_i$. For $S\subseteq[n]=\{1,\dots,n\}$, $V_S=(V_i)_{i\in S}$, $\mathcal V_S=\prod_{i\in S}\mathcal V_i$, and $v_S$ is the restriction of $v\in\Omega$ to $S$. Kernel $i$ has declared input set $S_i\subseteq[i-1]$ and conditional law $k_i(\cdot\mid v_{S_i})$ on $\mathcal V_i$. The baseline law is
\begin{equation}
P^0(v)=\prod_{i=1}^{n}k_i(v_i\mid v_{S_i}),\qquad v\in\Omega .
\label{eq:p2-baseline}
\end{equation}
A declared set $\mathcal K_i\subseteq\mathcal V_{S_i}$ collects inputs where $k_i$ is known by declaration (default $\mathcal K_i=\emptyset$).

Tree mechanisms of Proposition~\ref{prop:replay} embed into coordinates by allocating one slot per history where call order or kernel identity varies. Inactive slots emit a dummy value. Replaced slots are those where replaced and baseline kernels differ. The post-abort path is a single valuation slot carrying the law of $L$ given the prefix (declared as $b_\kappa$ after terminal deviations by (A3)). Allowed sets $\mathcal E_{j,z}$ exclude the dummy at active inputs, and equal $\{\ell(\omega):\omega\text{ extends }p\}$ at undeclared abort prefixes $p$.

\paragraph{Replacement.}
Replacement $\kappa$ acts at boundary $s$, replacing kernels $D_\kappa\subseteq\{s,\dots,n\}$ by known kernels $k^\kappa_j(\cdot\mid v_{S^\kappa_j})$ with $S^\kappa_j\subseteq[j-1]$ (with $k^\kappa_j=\delta_{f^\kappa_j(v_{S^\kappa_j})}$ for deterministic replacements). For $i\notin D_\kappa$, $k^\kappa_i=k_i$ and $S^\kappa_i=S_i$. Invariance and prefix preservation ((A1) and (A2)) give
\begin{equation}
P^\kappa(v)=\prod_{i=1}^{n}k^\kappa_i(v_i\mid v_{S^\kappa_i}).
\label{eq:p2-replaced}
\end{equation}
Here $P^0=P_{E,K}$ and $P^\kappa=P_{E^\kappa,K^\kappa}$ on $\Omega$.

\paragraph{Loss and Records.}
Loss is $L=\ell(V)=\bar\ell(V_{S_L})$ for $S_L\subseteq[n]$ and $\bar\ell\colon\mathcal V_{S_L}\to\lossset$. Records are $V_{\mathcal O}$ for $\mathcal O\subseteq[n]$. The pre-intervention record is $R_t=V_{S_r}$ with $S_r\subseteq\mathcal O\cap[s-1]$ and $P^0(R_t=r)>0$ ($t=-1, S_r=\emptyset$ for unconditional queries). A value $v$ is $r$-compatible when $v_{S_r}=r$. The continuation query is $Q^\kappa_t(B\mid r)=P^\kappa(L\in B\mid R_t=r)$ for $B\subseteq\lossset$.

\begin{definition}[Downstream and Relevant Sets]
\label{def:p2-sets}
The downstream set $\Delta_\kappa\subseteq[n]$ is defined recursively in index order: $i\in\Delta_\kappa$ if $i\in D_\kappa$, or if $i\notin D_\kappa$ and $S_i\cap\Delta_\kappa\neq\emptyset$. Its complement $\bar\Delta_\kappa=[n]\setminus\Delta_\kappa$ is the unaffected set. The relevant set $J_\kappa$ is the smallest set that contains $S_L\cup S_r$ and satisfies $S^\kappa_j\subseteq J_\kappa$ for every $j\in J_\kappa\cap\Delta_\kappa$. A set $K\subseteq[n]$ is $\kappa$-closed when $S^\kappa_j\subseteq K$ for every $j\in K\cap\Delta_\kappa$.
\end{definition}

The recursion is well defined because $S_i\subseteq[i-1]$; the smallest set exists by intersection of closed sets.

\subsection{Assumptions and Their Deployment Reading}
\label{app:supported-assumptions}

These are the assumptions for Proposition~\ref{prop:supported} (Section~\ref{sec:supported}).

\begin{itemize}
\item[(G1)] \emph{Support.} For every $j\in J_\kappa\cap(\Delta_\kappa\setminus D_\kappa)$ and every $z\in\mathcal V_{S_j}\setminus\mathcal K_j$, $P^\kappa(V_{S_j}=z,\ R_t=r)>0$ implies $P^0(V_{S_j}=z)>0$.
\item[(G2)] \emph{Recorded downstream inputs.} $S_j\subseteq\mathcal O$ for every $j\in J_\kappa\cap(\Delta_\kappa\setminus D_\kappa)$.
\item[(G3)] \emph{Recorded loss inputs.} $S_L\subseteq\mathcal O$.
\item[(G4)] \emph{Recorded assignment.} $S^\kappa_j\subseteq\mathcal O$ and $k^\kappa_j$ is known for every $j\in J_\kappa\cap D_\kappa$.
\end{itemize}
For an unchanged agent reading full text context, (G1) fails if a refusal message or regenerated text has zero baseline probability at that position. Coarsening that context creates support only under a sufficient-input assumption that logs alone cannot confirm.

\subsection{Structural Lemmas}
\label{app:supported-lemmas}

These lemmas support the proof of Proposition~\ref{prop:supported} and its extension to unsupported inputs (Section~\ref{sec:supported}).

Every $i\in\bar\Delta_\kappa$ has $i\notin D_\kappa$ and $S_i\subseteq\bar\Delta_\kappa$, so $\Delta_\kappa\subseteq\{s,\dots,n\}$ and $S_r\subseteq\bar\Delta_\kappa$; minimality and (G2) to (G4) give $J_\kappa\subseteq\mathcal O$.

\begin{lemma}[Truncated Factorization]
\label{lem:p2-truncation}
Under both laws, the marginal of $V_{\bar\Delta_\kappa}$ is $\prod_{i\in\bar\Delta_\kappa}k_i(v_i\mid v_{S_i})$; in particular $P^\kappa$ and $P^0$ agree on $V_{\bar\Delta_\kappa}$ and on $R_t$. For every $\kappa$-closed $K$ and $w\in\mathcal V_K$,
\begin{equation}
\begin{aligned}
&P^\kappa(V_K=w)\\
&=P^0\!\left(V_{K\cap\bar\Delta_\kappa}=w_{K\cap\bar\Delta_\kappa}\right)\\
&\qquad\prod_{j\in K\cap\Delta_\kappa}k^\kappa_j\!\left(w_j\mid w_{S^\kappa_j}\right).
\end{aligned}
\label{eq:p2-truncation}
\end{equation}
\end{lemma}
\begin{proof}[Proof sketch]
Apply the causal Markov factorization \citep[Theorem~1, Eq.~(15)]{pearl2009overview} to the finite independent-seed model of $P^\kappa$, in whose replacement graph $\bar\Delta_\kappa$ and $K\cup\bar\Delta_\kappa$ are ancestral.
Marginalizing the latter to $K$ gives \eqref{eq:p2-truncation} by $\kappa$-closure.
\end{proof}

\begin{lemma}[Kernel Recovery]
\label{lem:p2-recovery}
Let $j\in\Delta_\kappa$, and let $K\subseteq\bar\Delta_\kappa\cup(\Delta_\kappa\cap[j-1])$ contain $S_j$. If $P^0(V_K=u)>0$, then $P^0(V_j=a\mid V_K=u)=k_j(a\mid u_{S_j})$ for every $a\in\mathcal V_j$, where $k_j$ is the baseline kernel.
\end{lemma}
\begin{proof}
Because $K$ contains only nondescendants of $j$ along with its inputs $S_j$, the claim follows from the parental Markov condition for \eqref{eq:p2-baseline} \citep[Theorem~1.2.7]{pearl2009causality}.
\end{proof}

\subsection{Full Statement and Proof of \texorpdfstring{Proposition~\ref{prop:supported}}{the Supported-Continuation Proposition}}
\label{app:supported-proof}

Enumerate $J_\kappa\cap\Delta_\kappa=\{j_1<\dots<j_q\}$ and set $K_m=(J_\kappa\cap\bar\Delta_\kappa)\cup\{j_1,\dots,j_m\}$ for $0\leq m\leq q$ ($K_q=J_\kappa$). For unchanged $j$, let $\mathcal Z_j\subseteq\mathcal V_{S_j}\setminus\mathcal K_j$ be inputs, and let $\lambda$ assign $\lambda_{j,z}\in\Delta(\mathcal V_j)$ to each $z\in\mathcal Z_j$. For $r$-compatible $v\in\mathcal V_{J_\kappa}$, define $G^\lambda_0(v)=P^0(V_{K_0}=v_{K_0})$ and $G^\lambda_m(v)=G^\lambda_{m-1}(v)\,c^\lambda_{j_m}(v)$ ($0$ if $G^\lambda_{m-1}(v)=0$), where
\begin{equation}
\begin{gathered}
c^\lambda_j(v)=\\
\begin{cases}
k_j\!\left(v_j\mid v_{S_j}\right),\quad j\notin D_\kappa,\ v_{S_j}\in\mathcal K_j,\\
\begin{aligned}
&P^0\!\left(V_j=v_j\mid V_{S_j}=v_{S_j}\right),\\
&\quad j\notin D_\kappa,\ v_{S_j}\notin\mathcal K_j\cup\mathcal Z_j,
\end{aligned}\\
k^\kappa_j\!\left(v_j\mid v_{S^\kappa_j}\right),\quad j\in D_\kappa,\\
\lambda_{j,v_{S_j}}(v_j),\quad j\notin D_\kappa,\ v_{S_j}\in\mathcal Z_j .
\end{cases}
\end{gathered}
\label{eq:p2-factors}
\end{equation}
Here every $\mathcal Z_j=\emptyset$, so the fourth case never applies; write $c_j$ and $G=G_q$. Section~\ref{app:r4-unsupported} takes $\mathcal Z_j$ to be the null inputs.

\paragraph{Full Statement.}
In the finite coordinate model, let $P^0(R_t=r)>0$.
\begin{enumerate}
\item[(i)] \emph{g-formula.} Under (G1), (G2), (G3), and (G4), every $c_j$ used in $G$ is well defined, and $P^\kappa(V_{J_\kappa}=v)=G(v)$ for every $r$-compatible $v\in\mathcal V_{J_\kappa}$. Hence, for every $B\subseteq\lossset$,
\begin{equation}
Q^\kappa_t(B\mid r)=\frac{1}{P^0(R_t=r)}\sum_{\begin{subarray}{c}v\in\mathcal V_{J_\kappa}:\\\ v_{S_r}=r,\\\ \bar\ell(v_{S_L})\in B\end{subarray}}G(v).
\label{eq:p2-gformula}
\end{equation}
The right side depends only on the baseline record law, the declared inputs, the declared kernel values, $\bar\ell$, and the known replacement kernels. The vector law $Q^\kappa_t(\cdot\mid r)$ and every query $\rho(\phi_\#Q^\kappa_t(\cdot\mid r))$ are therefore identified within mechanisms that share the declared inputs and declared kernel values.
\item[(ii)] \emph{Support is checkable.} Under (G2), (G3), and (G4), whether (G1) holds is determined by the baseline record law and the declared kernel values.
\item[(iii)] \emph{Zero agreement.} There exists a mechanism satisfying (G1) to (G4) in which the baseline and the replacement are deterministic given a recorded tape. Its agreement mass is $0$, and its replacement risk is identified as $1/2$.
\item[(iv)] \emph{Composition with Proposition~\ref{prop:replay}.} Let the coordinate list embed a tree mechanism of Proposition~\ref{prop:replay} as in Section~\ref{app:supported-model}, with (R), (A1), (A2), and (A3) of that proposition, and a conditioning record generated before every replaced slot. Then (G1) holds at every input that the replacement reaches on an agreement path, and every downstream input reached after a terminal deviation is declared. Under (G1) to (G4), Proposition~\ref{prop:replay}(i) and (ii) give the restrictions of $Q^\kappa_t(\cdot\mid r)$ to the agreement set and to terminal deviations. The continuing part is therefore identified as the remainder.
\item[(v)] \emph{Unsupported inputs.} Without (G1), the identified set is the image described in Proposition~\ref{prop:r4-unsupported}.
\end{enumerate}
Conditions (G2) and (G4) cannot be dropped, and (G3) cannot be weakened to recording $L$ alone (\texttt{verify\_supported.py}, Appendix~\ref{app:checks}).

\begin{proof}[Proof sketch]
Part (i) is the sequential g-formula \citep{robins1986new,pearl1995sequential}, established here in Proposition~\ref{prop:r4-unsupported}(a),(c).
For (ii), $J_\kappa\subseteq\mathcal O$ makes the recursion observable up to any first reached undeclared null input, which detects failure of (G1).
For (iii), record all coordinates of $U\sim\mathrm{Bernoulli}(1/2)$, baseline $A=U$, replacement $A=1-U$, and $Y$ whose kernel reads only $A$, with $P(Y=1\mid A=0)=1/4$ and $P(Y=1\mid A=1)=3/4$; take $L=Y$ and no conditioning record. Both outcome inputs have baseline support, all decisions disagree, and the g-formula gives replacement risk $1/2$.
For (iv), Lemma~\ref{lem:app-prefix} gives support on agreement paths, and (A3) declares terminal valuations; subtract the two restrictions in Proposition~\ref{prop:replay}.
Part (v) is Proposition~\ref{prop:r4-unsupported}.
\end{proof}

\subsection{Continuation from Records with Unsupported Inputs}
\label{app:r4-unsupported}

To handle unsupported inputs in Section~\ref{sec:supported}, we keep (A1), (A2), and (G2) to (G4), drop (G1), and vary kernel laws at unsupported inputs within allowed sets.

In the coordinate model of Section~\ref{app:supported-model}, let $\mathcal U_\kappa=J_\kappa\cap(\Delta_\kappa\setminus D_\kappa)$. For $j\in\mathcal U_\kappa$ and $z\in\mathcal V_{S_j}\setminus\mathcal K_j$, kernel $j$ emits values in a nonempty allowed set $\mathcal E_{j,z}\subseteq\mathcal V_j$ (default $\mathcal V_j$). Members of $\mathcal M_0$ share the declared inputs, declared kernel values, known replacement kernels, and allowed sets. Each coordinate has its own kernel; restrictions tying slots are addressed in Remark~\ref{rem:r4-scope}. The population identified set is
\begin{equation}
\begin{aligned}
&\Theta_r=\bigl\{Q^\kappa_{t,M}(\cdot\mid r):\ M\in\mathcal M_0,\\
&\ V_{\mathcal O}\text{ has the observed record law under }P^0_M\bigr\}.
\end{aligned}
\label{eq:r4-identified}
\end{equation}
Under (G2), an input $z\in\mathcal V_{S_j}\setminus\mathcal K_j$ with $P^0(V_{S_j}=z)=0$ is a null input; collect them in $\mathcal Z_j$. The null blocks are $\mathcal B=\{(j,z):j\in\mathcal U_\kappa,\ z\in\mathcal Z_j\}$, and free families form $\Lambda=\prod_{(j,z)\in\mathcal B}\Delta(\mathcal E_{j,z})$. For a member $M$ with the observed record law, define its own null-input family by $\lambda_M=(k^M_j(\cdot\mid z))_{(j,z)\in\mathcal B}$. For $\lambda\in\Lambda$, the recursion of Section~\ref{app:supported-proof} defines $G^\lambda$ via \eqref{eq:p2-factors}, and
\begin{equation}
Q^\lambda(B\mid r)=\frac{1}{P^0(R_t=r)}\sum_{\begin{subarray}{c}v\in\mathcal V_{J_\kappa}:\\\ v_{S_r}=r,\\\ \bar\ell(v_{S_L})\in B\end{subarray}}G^\lambda(v).
\label{eq:r4-gformula}
\end{equation}
The class $\mathcal M_0$ is \emph{closed at null inputs} if for each member $M$ with the observed record law and each $\lambda\in\Lambda$, setting $k_j(\cdot\mid z)=\lambda_{j,z}$ on $\mathcal B$ yields a member $M^\lambda\in\mathcal M_0$.

\begin{lemma}[Zero-Support Modification in the Coordinate Model]
\label{lem:r4-zero}
Let $M$ and $M'$ have the same declared inputs and the same kernels, except kernel $j$ at inputs in a set $\mathcal Z\subseteq\mathcal V_{S_j}$. If $P^0_M(V_{S_j}\in\mathcal Z)=0$, then $P^0_{M'}=P^0_M$.
\end{lemma}
\begin{proof}
The products \eqref{eq:p2-baseline} of $M$ and $M'$ differ only where $v_{S_j}\in\mathcal Z$. Since $S_j\subseteq[j-1]$, the probability of $\{V_{S_j}\in\mathcal Z\}$ depends only on the shared kernels $1,\dots,j-1$, so it is $0$ under both and both laws vanish there.
\end{proof}

\begin{proposition}[Continuation from Records with Unsupported Inputs]
\label{prop:r4-unsupported}
In the finite coordinate model, assume \textup{(A1)} and \textup{(A2)} in the form \eqref{eq:p2-replaced}, together with \textup{(G2)}, \textup{(G3)}, and \textup{(G4)}. Condition \textup{(G1)} is not assumed.
\begin{enumerate}
\item[(a)] \emph{Outer set.} Let $M$ have the declared inputs, the declared kernel values, and the known replacement kernels, with $P^0_M(R_t=r)>0$. Formula \eqref{eq:r4-gformula}, computed from the record law of $M$ with $\lambda=\lambda_M$, equals $Q^\kappa_{t,M}(\cdot\mid r)$. Hence $\Theta_r\subseteq\{Q^\lambda(\cdot\mid r):\lambda\in\Lambda\}$, where each $Q^\lambda$ is computed from the observed record law.
\item[(b)] \emph{Sharpness.} If $\mathcal M_0$ is closed at null inputs, then $\Theta_r=\{Q^\lambda(\cdot\mid r):\lambda\in\Lambda\}$.
\item[(c)] \emph{Supported extreme.} Suppose \textup{(G1)} holds; by Proposition~\ref{prop:supported}(ii), this is a property of the observed record law and the declared kernel values. Then $Q^\lambda(\cdot\mid r)$ equals the right side of \eqref{eq:p2-gformula} for every $\lambda\in\Lambda$, and the image is the single law of Proposition~\ref{prop:supported}(i).
\end{enumerate}
\end{proposition}
\begin{proof}[Proof of (a)]
Fix $M$ and $\lambda=\lambda_M$. Each $K_m$ is $\kappa$-closed, so Lemma~\ref{lem:p2-truncation} gives
\begin{equation}
\begin{aligned}
\Pi_m(v)&:=P^\kappa_M\!\left(V_{K_m}=v_{K_m}\right)\\
&=P^0_M\!\left(V_{K_0}=v_{K_0}\right)\\
&\qquad\prod_{l\leq m}k^\kappa_{j_l}\!\left(v_{j_l}\mid v_{S^\kappa_{j_l}}\right).
\end{aligned}
\label{eq:p2-prefix}
\end{equation}
Lemma~\ref{lem:p2-recovery} and the definitions of declared, replacement, and null factors give $G^{\lambda_M}_m=\Pi_m$, including zero prefixes.
Summing and conditioning on $r$ gives \eqref{eq:r4-gformula}; $J_\kappa\subseteq\mathcal O$ makes all terms except $\lambda$ observable, proving the inclusion.
\end{proof}

\begin{proof}[Proof of (b)]
For $\lambda\in\Lambda$ and a compatible $M$, closure gives $M^\lambda\in\mathcal M_0$. Resetting the null blocks one at a time, Lemma~\ref{lem:r4-zero} keeps the baseline law at each step, so $P^0_{M^\lambda}=P^0_M$ and $\lambda_{M^\lambda}=\lambda$. Part~(a) for $M^\lambda$ gives $Q^\lambda(\cdot\mid r)=Q^\kappa_{t,M^\lambda}(\cdot\mid r)\in\Theta_r$.
\end{proof}

\begin{proof}[Proof of (c)]
By (G1), no first undeclared null input can follow a positive $r$-compatible prefix in (a); zero prefixes stay zero.
Thus no free factor is used, so $G^\lambda=G$ for every $\lambda$, giving \eqref{eq:p2-gformula} by (a).
\end{proof}

\begin{remark}[What Finite Logs Show]
\label{rem:r4-logs}
Under (G2) to (G4), the population record law determines null inputs. Absence from a finite log does not prove an input is null. Treating unobserved inputs as null still yields a valid outer set. Setting free arguments to the true conditional law returns \eqref{eq:p2-factors}, so the enlarged image contains the true identified set. Finite-sample inference is a separate claim (Limitations).
\end{remark}

\begin{remark}[Scope of Sharpness]
\label{rem:r4-scope}
Part (b) uses that each coordinate's kernel is free at its own null inputs. In the slot embedding of Section~\ref{app:supported-model}, several slots can host one named kernel. A class that ties slots, such as one law per kernel and step when the embedding needs more slots than kernel and step pairs, is covered by the outer-set clause of Proposition~\ref{prop:supported}. It is a subclass with the same evidence, so part (a) still gives an outer set for it. Its identified set can be smaller.
\end{remark}

\subsection{Model Updates: Statement and Proof}
\label{app:model-update}

This is the model-release result of Section~\ref{sec:supported}.

\begin{corollary}[Model updates]\label{cor:model-update}
Assume \textup{(A1)} and \textup{(A2)} for every replaced decision, and let $\kappa$ replace the agent policy at its decisions from the boundary onward, under an interface whose output contains every action and dynamic state that later kernels or $\ell$ read; a model update applied from the episode start replaces every decision. Suppose the baseline policy is deterministic given its decision input and a random tape, and that decision inputs, tapes, and baseline outputs are recorded. If the new policy is seed-addressable, that is, evaluable deterministically at each logged input and tape, then the agreement mass $\mathsf m_\kappa(r)=P^\kappa(\mathsf A_\kappa\mid r)$ is identified. It is estimated by the empirical agreement frequency over independent baseline episodes and is at most the first-decision agreement probability. With the loss value recorded, the agreement contribution is identified as well.
\end{corollary}

\begin{proof}[Proof sketch]
Including tapes in the inputs, the stated determinism and recording give (R); Lemma~\ref{lem:app-prefix} identifies agreement mass and its logged loss contribution.
Seed-addressable evaluation determines the agreement indicators, whose empirical frequency is consistent by the strong law.
The bound follows since agreement throughout implies first-decision agreement.
\end{proof}

\subsection{Budgeted Targeted Execution: Statement and Proof}
\label{app:budget}

This subsection proves the sampling and restoration bounds used in Section~\ref{sec:supported}. Scripts in \texttt{checks/} reproduce the E1 and E7 bounds and check finite instances; Appendix~\ref{app:checks} and \texttt{checks/README.md} map each script to its claim.

Replay supplies the agree and stop terms of display~\eqref{eq:split}; a uniform sample of unresolved continuations bounds the rest. Hypergeometric and binomial inversion \citep{clopper1934use}, a union bound, and error spending over declared looks \citep{lan1983discrete} determine how many executions a decision needs.

\paragraph{Setting.} Let harm be binary and $q$ the changed system's P(harm), the target $q^\kappa_t(r)$ with $X$ the harm indicator. A pass-all log holds $n$ episodes. Under (R), (A1), (A2), and (A3), replay resolves an episode's outcome under the change when it agrees, when harm occurred before its first deviation, or when it stops and the abort valuation fixes the harm indicator (a point mass, as when a stop ends the episode with the harm accrued before it); $b$ resolved episodes are harmful. A nondegenerate abort valuation would need an independent draw for each stopped episode. The remaining $N$ continuing episodes are unresolved, partitioned into strata $s$ of sizes $N_s$ fixed by the log (for example, by the step of first deviation). Executing an unresolved episode's continuation from its snapshot gives its outcome under the change, by (T). The completed cohort assigns each resolved episode its replayed outcome and each unresolved episode the outcome of its continuation; $S$ is its harm count.

\paragraph{Hypothesis (H).} The completed outcomes are independent Bernoulli variables with common probability $q$; the random order in which unresolved episodes are executed is drawn independently of them, and the looks are fixed in advance. Independently sampled tasks and fresh seeds support (H); (T) links each continuation to its live law but does not by itself make outcomes independent across episodes, so dependence through shared serving or environment state lies outside it.

\paragraph{Bounds.} In stratum $s$, execute a uniform random sample without replacement of $k_s$ unresolved episodes and observe $x_s$ harms. With $\mathrm{H}(N_s,D,k_s)$ the hypergeometric law of harms in such a sample when $D$ of the $N_s$ are harmful, let
\[
\begin{aligned}
&D_s^U(x_s,k_s;e)\\
&=\max\bigl\{D\in\{x_s,\ldots,N_s-k_s+x_s\}:\\
&\qquad\ \Pr_{\mathrm{H}(N_s,D,k_s)}(X\le x_s)>e\bigr\},
\end{aligned}
\]
with $D_s^U=N_s$ when $k_s=0$; at $k_s=N_s$ it returns $x_s$. Let $S^U=b+\sum_s D_s^U$ and let $u_{\mathrm{CP}}(z,n;a)$ be the one-sided Clopper-Pearson upper limit for $z$ successes in $n$ trials at level $1-a$, with $u_{\mathrm{CP}}(n,n;a)=1$.

\begin{proposition}[Budgeted targeted execution]\label{prop:budget}
\textup{(i)} Given the completed cohort, $\Pr(S\le S^U)\ge 1-\sum_s e_s$, where stratum $s$ uses error $e_s$.
\textup{(ii)} Under \textup{(H)}, $\Pr\bigl(q\le u_{\mathrm{CP}}(S^U,n;a_L)\bigr)\ge 1-a_L-\sum_s e_s$.
\textup{(iii)} Let samples be nested over declared looks $j=1,\ldots,J$ with sizes $k_{s,j}$ and errors $e_{s,j}$, $\sum_{s,j}e_{s,j}\le a_E$, and let $\overline U_j=\min_{l\le j}u_{\mathrm{CP}}(S^U_l,n;a_L)$. Under \textup{(H)}, $\Pr(q\le\overline U_j\text{ for all }j)\ge 1-a_L-a_E$. Stopping at the first look whose bound clears a target, or at look $J$, keeps this coverage, for every target at once.
\end{proposition}

\begin{proof}
(i) If $D>D_s^U(x_s,k_s;e_s)$ then $\Pr_{\mathrm{H}(N_s,D,k_s)}(X\le x_s)\le e_s$, so in each stratum the failure event is a lower tail of probability at most $e_s$; a union bound gives the claim.
(ii) Combine (i) with Clopper-Pearson coverage for $S\sim\mathrm{Bin}(n,q)$ under (H) and monotonicity in the count \citep{clopper1934use}.
(iii) A union bound over strata and declared looks, intersected with the single Clopper-Pearson event for $S$, gives simultaneous coverage of the running minima. This also covers any selected look and every target.
\end{proof}

The guarantees above concern sampling error under exact restoration \textup{(T)}. Lemma~\ref{lem:restore} prices restoration error, and Remark~\ref{rem:restore} incorporates it into certified bounds.

\begin{lemma}[Restoration Error]\label{lem:restore}
Assume \textup{(R)} and \textup{(A1)} to \textup{(A3)}, and let $\nu$, $w_h$, and $Q_h=\phi_\#P^\kappa(L\in\cdot\mid[h])$ be as in Proposition~\ref{prop:app-identified}. Let $\hat Q_h$ be the law of $X=\phi(L)$ in the run restored at $h^+$, so that \textup{(T)} says $\hat Q_h=Q_h$, and let $\gamma_h=\max_{B}|\hat Q_h(B)-Q_h(B)|$. Let $\hat P=\nu+\sum_hw_h\hat Q_h$ be the law that display~\eqref{eq:split} gives when its continue term uses the restored laws, with mean $\hat\mu$, and let $\varepsilon_\kappa=\sum_hw_h$ be the continuing mass. Then for every event $B$ of $X$,
\begin{gather*}
\begin{aligned}
&|\hat P(B)-P^\kappa(X\in B\mid r)|\\
&\qquad\le\sum_hw_h\gamma_h\le\varepsilon_\kappa\max_h\gamma_h,
\end{aligned}\\
\begin{aligned}
&|\hat\mu-E^\kappa[X\mid r]|\\
&\qquad\le(\max\mathcal X-\min\mathcal X)\sum_hw_h\gamma_h .
\end{aligned}
\end{gather*}
\end{lemma}
\begin{proof}
Both event bounds follow from the decomposition in Proposition~\ref{prop:app-identified} and the triangle inequality; integrating tail differences over $[\min\mathcal X,\max\mathcal X]$ gives the mean bound.
\end{proof}
\begin{remark}[Restoration Allowance]\label{rem:restore}
Proposition~\ref{prop:budget} bounds the harm probability $\hat q$ of $\hat P$ when the completed restored outcomes satisfy hypothesis (H): independent Bernoulli with a common probability. Let $u$ be its certified bound and $\xi=\sum_hw_h\gamma_h$. Lemma~\ref{lem:restore} gives $|q-\hat q|\le\xi$, so with the probability of the proposition $q\le u+\xi$, and a target $q^\star$ is cleared when $u+\xi\le q^\star$; exact \textup{(T)} is the case $\xi=0$. For a tolerance $\xi_0\ge\xi$ declared in advance, stopping at the first look with $\overline U_j+\xi_0\le q^\star$ keeps the coverage of part (iii). The continuing mass $\varepsilon_\kappa$ is estimated from the log, so a population statement uses an upper confidence limit for it, whose error adds to that of the proposition, or the bound $\varepsilon_\kappa\le1$. At prefixes with harm before the deviation, $Q_h=\hat Q_h$ for harm, so they drop out of the harm allowance; task success can still depend on the continuation, so a decision with a success floor needs its own allowance from the same lemma. Both bounds of the lemma are attained: with $Q_h=\delta_{\min\mathcal X}$ and $\hat Q_h=(1-\gamma)\delta_{\min\mathcal X}+\gamma\delta_{\max\mathcal X}$ for every $h$, the errors equal $\varepsilon_\kappa\gamma$ and $(\max\mathcal X-\min\mathcal X)\varepsilon_\kappa\gamma$. A paired live run measures $q-\hat q$ itself and does not bound $\gamma_h$, since errors at different prefixes can offset.
\end{remark}

\subsection{Necessity of the Execution Budget}
\label{app:budget-limits}

This subsection proves the execution limits and achievability claims of Section~\ref{sec:supported}.

\paragraph{Setting.} Let harm be binary, and assume \textup{(R)}, \textup{(A1)} to \textup{(A3)}, and \textup{(S)} with feasible set $\{0,1\}$ at every unresolved prefix. Episodes are independent. Each is resolved and harmful with probability $q_0$, unresolved with probability $m>0$, and resolved and harmless otherwise. An unresolved continuation is harmful with probability $p\in(0,1)$. Write $q=q_0+mp$, and assume $\zeta=q^\star-q>0$ and $q^\star<q_0+m$. A \emph{world} gives every unresolved prefix a live harm probability $p_0$ and a restored one $\hat p$ with $|p_0-\hat p|\le\gamma$. Condition \textup{(S)} frees every live $p_0$ for a mechanism with the observed baseline law. We also let every restored $\hat p$ within $\gamma$ of $p_0$ occur, as in the attainment part of Remark~\ref{rem:restore}. The log then has one law across worlds, and each restored cohort satisfies \textup{(H)} because episodes are independent. A \emph{procedure} reads the log, may randomize internally, and executes unresolved episodes one at a time, each at most once and chosen adaptively. An execution reveals the record of one restored continuation, independent of the other executions given the log. Worlds may give these records any law with the stated harm probabilities, so a world can change a harm probability and keep the conditional law of the record given harm. It stops after $K\le n$ executions and certifies or not. Fix $\alpha,\beta\in(0,1)$ with $\alpha+\beta<1$ in this subsection; probabilities average over the log and any randomization. The \emph{size} of a procedure is its largest certification probability over worlds with $q_0+mp_0>q^\star$. Its \emph{power} is its certification probability in the world $W_1$ with $p_0=\hat p=p$. It is \emph{valid} if its size is at most $\alpha$ and \emph{powerful} if its power is at least $1-\beta$. Let $\mathrm{kl}$ be the binary relative entropy and $\bar p=p+(\zeta-m\gamma)/m$.

\begin{proposition}[Execution Budget: Necessity]\label{prop:budget-limits}
In the setting above, the following hold.
\begin{enumerate}
\item[(i)] If $m\gamma\ge\zeta$, every procedure has power at most its size, so no valid procedure is powerful, for any $n$.
\item[(ii)] If $m\gamma<\zeta$, every valid and powerful procedure has $\mathbb E_{W_1}[K]\ge\mathrm{kl}(1-\beta,\alpha)/\mathrm{kl}(p,\bar p)\ge\mathrm{kl}(1-\beta,\alpha)\,m^{2}\bar p(1-\bar p)/(\zeta-m\gamma)^{2}$. The first bound diverges as $m\gamma\uparrow\zeta$.
\item[(iii)] Consider a sequential test of independent live episodes whose stopping time $K_L$ has finite mean at live harm $q$. If it certifies with probability at most $\alpha$ under every episode law whose harm probability exceeds $q^\star$, and at least $1-\beta$ at harm $q$, then $\mathbb E_q[K_L]\ge\mathrm{kl}(1-\beta,\alpha)/\mathrm{kl}(q,q^\star)\ge\mathrm{kl}(1-\beta,\alpha)\,q^\star(1-q^\star)/\zeta^{2}$.
\end{enumerate}
\end{proposition}

\begin{proof}
For $\varepsilon>0$, let $W_0^\varepsilon$ have $p_0=p+(\zeta+\varepsilon)/m$ and $\hat p_\varepsilon=\max\{p_0-\gamma,p\}$. For small $\varepsilon$ it is a world, since $p_0<1$ when $q^\star<q_0+m$, and its live harm is $q^\star+\varepsilon$. At every prefix, $W_0^\varepsilon$ keeps the conditional law of the restored record given harm from $W_1$. Let $E$ be the event that the procedure certifies, with laws $\mathbb P_1$ in $W_1$ and $\mathbb P_0$ in $W_0^\varepsilon$. The log has the same law in both. Since $W_0^\varepsilon$ has live harm above $q^\star$, $\mathbb P_0(E)$ is at most the size, and at most $\alpha$ for a valid procedure.
\par(i) Let $m\gamma\ge\zeta$. Then $\hat p_\varepsilon-p\le\varepsilon/m$. Draw one uniform variable per unresolved episode, and let the episode be harmful when the variable falls below $p$ in $W_1$ and below $\hat p_\varepsilon$ in $W_0^\varepsilon$. Draw the rest of its record from the common conditional law given harm. The two outcomes differ with probability at most $\varepsilon/m$ per episode. The procedure acts identically unless one of the at most $n$ unresolved outcomes differs, so $\mathbb P_1(E)\le\mathbb P_0(E)+n\varepsilon/m$. Letting $\varepsilon\downarrow0$ bounds the power by the size. A valid procedure has size at most $\alpha<1-\beta$, so it is not powerful.
\par(ii) Let $m\gamma<\zeta$, so $\hat p_\varepsilon=\bar p+\varepsilon/m\in(p,1)$ for small $\varepsilon$, since $\bar p\le(q^\star-q_0)/m<1$. Conditional on the common log and internal randomization $\mathcal G$, unused episodes are independent arms. Their record laws have common conditional laws given harm, hence are mutually absolutely continuous with divergence $\mathrm{kl}(p,\hat p_\varepsilon)$. The stopping time satisfies $K\le n$. Apply \citet[Lemma~1]{kaufmann2016complexity} conditionally and average using joint convexity of $\mathrm{kl}$:
\[
\begin{aligned}
\mathbb E_1[K]\,\mathrm{kl}(p,\hat p_\varepsilon)&\ge\mathrm{kl}(\mathbb P_1(E),\mathbb P_0(E))\\
&\ge\mathrm{kl}(1-\beta,\alpha),
\end{aligned}
\]
by validity, power, and binary-KL monotonicity. Let $\varepsilon\downarrow0$. The standard binary-KL bound $\mathrm{kl}(p,\bar p)\le(\bar p-p)^2/\{\bar p(1-\bar p)\}$ gives the second bound. As $m\gamma\uparrow\zeta$, $\bar p\downarrow p$ and the first bound diverges.
\par(iii) Give the unsafe live law harm probability $q^\star+\varepsilon$, keeping the conditional record law given harm from $W_1$. For small $\varepsilon$, $0<q<q^\star+\varepsilon<1$. Apply the same lemma with one arm, stopping at $\min\{K_L,t\}$ and event $E\cap\{K_L\le t\}$. Let $t\to\infty$, using $\mathbb E_q[K_L]<\infty$, and then $\varepsilon\downarrow0$. The same binary-KL bound gives the second inequality.
\end{proof}

\paragraph{Achievability.} Sequential probability ratio tests \citep[SPRTs;][]{wald1945sequential} approach the bounds of Proposition~\ref{prop:budget-limits}(ii) and (iii) as error levels fall and the log grows, with a finite-pool allowance; Remark~\ref{rem:estimate-log} treats estimation of $q_0$ and $m$.

Let $m\gamma<\zeta$, and put $\mathrm{up}(a,c)=\log\{(1-a)/(1-c)\}$ for $0<a<c<1$. The log has one law in every world, so a procedure may use $q_0$ and $m$. The \emph{execution test} with levels $\alpha,\beta'\in(0,1)$ uses $\bar p=(q^\star-q_0)/m-\gamma$. It executes unresolved episodes in any order and reads only their harm bits. Starting at $\Xi_0=0$, it adds $\log(p/\bar p)$ after a harm and $\mathrm{up}(p,\bar p)$ after a harmless outcome. It certifies when $\Xi_t\ge\log(1/\alpha)$, and stops without certifying when $\Xi_t\le-\log(1/\beta')$ or all $N$ unresolved episodes are used. The \emph{live test} applies the same rule to live harm bits, with $(q,q^\star)$ in place of $(p,\bar p)$ and no pool. A test with \emph{thresholds} $A,B>0$ uses $A$ and $B$ in place of $\log(1/\alpha)$ and $\log(1/\beta')$. Both tests look after every outcome, and Ville's inequality covers all looks.

\begin{proposition}[Execution Budget: Achievability]\label{prop:budget-achievability}
Assume the setting of Proposition~\ref{prop:budget-limits} with $m\gamma<\zeta$. Let $\tau$ be the stopping time of the execution test with levels $\alpha,\beta'\in(0,1)$ and an unlimited pool. Put $\bar K_{\mathrm{ex}}=\{\log(1/\alpha)+\mathrm{up}(p,\bar p)\}/\mathrm{kl}(p,\bar p)$ and $\chi_n=2\bar K_{\mathrm{ex}}/(mn)+e^{-m^2n/2}$.
\begin{enumerate}
\item[(i)] The size of the test is at most $\alpha$. For any thresholds, certification is nonincreasing in the restored harm rate, so the size is the certification probability at $\hat p=\bar p$. This holds in every world with $q_0+mp_0>q^\star$, for every $\hat p$ within $\gamma$ of $p_0$ and every record law given harm.
\item[(ii)] The power is at least $1-\beta'-\chi_n$. Also $\mathbb E_{W_1}[K]\le\mathbb E_p[\tau]\le\bar K_{\mathrm{ex}}$, where $\mathbb E_p$ refers to independent Bernoulli$(p)$ bits.
\item[(iii)] Let the live episodes be independent. The live test certifies with probability at most $\alpha$ when every episode has harm probability at least $q^\star$, and with probability at least $1-\beta'$ at harm $q$. Also $\mathbb E_q[K_L]\le\bar K_{\mathrm{lv}}=\{\log(1/\alpha)+\mathrm{up}(q,q^\star)\}/\mathrm{kl}(q,q^\star)$.
\end{enumerate}
\end{proposition}

\begin{proof}
For the Bernoulli SPRT score $Z$ \citep{wald1945sequential}, a direct computation gives $\mathbb E_\vartheta e^Z=1-(\vartheta-c)(c-a)/\{c(1-c)\}$, $\mathbb E_ae^{-Z}=1$, and $\mathbb E_aZ=\mathrm{kl}(a,c)$ for $0<a<c<1$; take $(a,c)=(p,\bar p)$.
\par(i) Given the log, executed bits are independent Bernoulli$(\hat p)$ with $\hat p\ge p_0-\gamma>\bar p$, whatever the record law. Thus $e^{\Xi_t}$, stopped at the pool, is a nonnegative supermartingale starting at one. Ville's inequality \citep[Chapter~V, Theorem~1, Eq.~(15)]{ville1939etude}, applied after adding the predictable decreases and approaching the threshold from below, bounds certification by $\alpha$. Coupling by uniforms makes certification nonincreasing in $\hat p$; finite-pool continuity gives the size at $\bar p$.
\par(ii) At $W_1$, Ville's inequality for $e^{-\Xi_t}$ bounds lower exit by $\beta'$. A run of $\lceil\{\log(1/\alpha)+\log(1/\beta')\}/\mathrm{up}(p,\bar p)\rceil$ harmless outcomes leaves the strip, so $\tau$ has a geometric tail. Wald's identity \citep{wald1944cumulative}, with upper overshoot at most $\mathrm{up}(p,\bar p)$, gives $\mathbb E_p[\tau]\le\bar K_{\mathrm{ex}}$. Since $K=\min\{\tau,N\}\le\tau$, the expected-cost bound follows. Failure requires a lower exit or pool exhaustion; Markov's and Hoeffding's inequalities with $N\sim\mathrm{Bin}(n,m)$ give $\Pr(\tau>N)\le\chi_n$, hence power at least $1-\beta'-\chi_n$.
\par(iii) The same argument with $(a,c)=(q,q^\star)$ and no pool gives the live bounds. Size requires only conditional harm probability at least $q^\star$; independent Bernoulli$(q)$ bits give the power and mean bounds.
\end{proof}

\begin{corollary}[Cost Bounds and Ratio]\label{cor:cost-ratio}
Let $m\gamma<\zeta$, $\alpha+\beta<1$, and $\chi_n<\beta$. Let $C_{\mathrm{ex}}$ be the infimum of $\mathbb E_{W_1}[K]$ over valid and powerful procedures. Let $C_{\mathrm{lv}}$ be the infimum of $\mathbb E_q[K_L]$ over the live tests of Proposition~\ref{prop:budget-limits}\textup{(iii)}. Write $\mathrm{kl}_{\mathrm{ex}}=\mathrm{kl}(p,\bar p)$, $\mathrm{kl}_{\mathrm{lv}}=\mathrm{kl}(q,q^\star)$, $\mathrm{up}_{\mathrm{ex}}=\mathrm{up}(p,\bar p)$, and $\mathrm{up}_{\mathrm{lv}}=\mathrm{up}(q,q^\star)$. Then
\[
\begin{aligned}
\frac{\mathrm{kl}(1-\beta,\alpha)}{\mathrm{kl}_{\mathrm{ex}}}&\le C_{\mathrm{ex}}\le\frac{\log(1/\alpha)+\mathrm{up}_{\mathrm{ex}}}{\mathrm{kl}_{\mathrm{ex}}},\\
\frac{\mathrm{kl}(1-\beta,\alpha)}{\mathrm{kl}_{\mathrm{lv}}}&\le C_{\mathrm{lv}}\le\frac{\log(1/\alpha)+\mathrm{up}_{\mathrm{lv}}}{\mathrm{kl}_{\mathrm{lv}}}.
\end{aligned}
\]
Each upper bound exceeds its lower bound by at most $[\mathrm{up}+\beta\{\log(1/\alpha)+\log(1/\beta)+1\}]/\mathrm{kl}$ with matching subscripts. Moreover
\[
\begin{aligned}
\frac{C_{\mathrm{ex}}}{C_{\mathrm{lv}}}&\ge\frac{\mathrm{kl}(1-\beta,\alpha)}{\log(1/\alpha)+\mathrm{up}_{\mathrm{lv}}}\cdot\frac{\mathrm{kl}_{\mathrm{lv}}}{\mathrm{kl}_{\mathrm{ex}}},\\
\frac{C_{\mathrm{ex}}}{C_{\mathrm{lv}}}&\le\frac{\log(1/\alpha)+\mathrm{up}_{\mathrm{ex}}}{\mathrm{kl}(1-\beta,\alpha)}\cdot\frac{\mathrm{kl}_{\mathrm{lv}}}{\mathrm{kl}_{\mathrm{ex}}}.
\end{aligned}
\]
Both outer factors tend to $1$ as $\alpha,\beta\downarrow0$ with $n$ large enough for $\chi_n<\beta$, so $C_{\mathrm{ex}}/C_{\mathrm{lv}}\to\mathrm{kl}_{\mathrm{lv}}/\mathrm{kl}_{\mathrm{ex}}$.
\end{corollary}

\begin{proof}
Combine Propositions~\ref{prop:budget-limits} and~\ref{prop:budget-achievability}, taking $\beta'=\beta-\chi_n$ for executions and $\beta'=\beta$ for live episodes; the ratio bounds divide the cost bounds, and expanding $\mathrm{kl}(1-\beta,\alpha)$ with $-(1-\beta)\log(1-\beta)\le\beta$ and $\beta\log(1-\alpha)\le0$ gives the gaps and limiting factors.
\end{proof}

With Proposition~\ref{prop:budget-limits}(i), this makes $\gamma=\zeta/m$ the sharp threshold as $n$ grows. For every $\gamma<\zeta/m$ there is an $n_0$, depending also on $\alpha$ and $\beta$, such that for $n\ge n_0$ a valid and powerful procedure exists. For $\gamma\ge\zeta/m$ none exists at any $n$.

\begin{remark}[Estimating $q_0$ and $m$]\label{rem:estimate-log}
A finite log gives only $b/n$ and $N/n$, and $\bar p$ decreases in $q_0$ and in $m$. Let $q_0^U\ge q_0$ and $m^U\ge m$ be Clopper--Pearson upper limits at level $\alpha_{\log}/2$ each, so that both hold with probability at least $1-\alpha_{\log}$. Run the execution test with $\bar p^U=(q^\star-q_0^U)/m^U-\gamma$ in place of $\bar p$ and the realized $N$ as pool, and require $p<\bar p^U<1$ before evaluating the test, never certifying when this condition fails. On the event of the limits, $\bar p^U\le\bar p<\hat p$, so the size is at most $\alpha+\alpha_{\log}$. Given a log with $p<\bar p^U<1$, the argument of (ii) at $\bar p^U$ gives $\mathbb E[K\mid\mathrm{log}]\le\bar K^U$, and Markov's inequality bounds pool exhaustion by $\bar K^U/N$, so the failure probability is at most $\beta'+\bar K^U/N$, where $\bar K^U$ is $\bar K_{\mathrm{ex}}$ at $\bar p^U$. Unconditionally, the power loses at most $\mathbb P(\bar p^U\le p\text{ or }\bar p^U\ge1)+\mathbb E[\min\{1,\bar K^U/N\};p<\bar p^U<1]$. Both terms vanish as $n$ grows, because $\bar p^U\to\bar p\in(p,1)$ and $N\to\infty$ in probability.
\end{remark}

\subsection{Execution Across Strata}
\label{app:budget-strata}

This extends the execution bounds of Section~\ref{sec:supported} to strata.

\paragraph{Setting.} We extend the framework of Appendix~\ref{app:budget-limits} to strata $s=1,\dots,S$, such as the step where the gate first blocks. An episode is unresolved in stratum $s$ with probability $w_s>0$, so $\sum_sw_s=m$. In stratum $s$, continuations are harmful with probability $p_s\in(0,1)$, restoration error is at most $\gamma_s\in[0,1]$, and each execution costs $c_s>0$ (for example, in agent calls); live episodes cost $c_L>0$. A world assigns stratum $s$ live harm probability $p_{0s}$ and restored rate $\hat p_s$ with $|p_{0s}-\hat p_s|\le\gamma_s$. An adaptive procedure executes unused unresolved episodes in chosen strata, incurring cost $C=\sum_sc_sK_s$. Let $q=q_0+\sum_sw_sp_s$, $\zeta=q^\star-q>0$, and $q^\star<q_0+m$. Power is evaluated at world $W_1$ ($p_{0s}=\hat p_s=p_s$), and size is the maximum certification probability over worlds with $q_0+\sum_sw_sp_{0s}>q^\star$.

A restored vector can come from a world of the null only if the live harm it allows, at most $q_0+\sum_sw_s\min\{1,\hat p_s+\gamma_s\}$, exceeds $q^\star$. So the restored vectors of the null lie in the convex set
\[
\begin{aligned}
\mathcal A=\Bigl\{&r\in[0,1]^S:\\
&\ q_0+\sum_sw_s\min\{1,r_s+\gamma_s\}\ge q^\star\Bigr\},
\end{aligned}
\]
and every $r\in(0,1)^S$ with strict inequality is one. Let $\xi=\sum_sw_s\min\{\gamma_s,1-p_s\}$ cap each restoration error at $1-p_s$ (Remark~\ref{rem:restore}), put $\zeta'=\zeta-\xi$, and let $J=\{s:p_s+\gamma_s<1\}$. For $s\notin J$, null worlds can keep restored rate $p_s$ while setting live harm to one, so the design below does not sample these strata. For $s\in J$ and $t\ge0$, let $z_s(t)\in[p_s,1)$ solve $\mathrm{kl}(p_s,z_s(t))=c_st$, and set $b_s(t)=\min\{1-\gamma_s,z_s(t)\}$. When $\zeta'>0$, let $t_*$ be the unique positive root of
\begin{equation}\label{eq:strata-root}
\textstyle\sum_{s\in J}w_s\{b_s(t)-p_s\}=\zeta',
\end{equation}
We define the \emph{design point}: $\bar p_s=b_s(t_*)$ on $J$ and $\bar p_s=p_s$ off $J$; active strata $J_*=\{s\in J:\bar p_s<1-\gamma_s\}$; slopes $g_s=(\bar p_s-p_s)/\{\bar p_s(1-\bar p_s)\}$; execution shares $a_s\propto w_s/g_s$ on $J_*$ ($a_s=0$ elsewhere); and mean cost $\bar c=\sum_sa_sc_s$. With $S=1$, $\bar p_1$ is the $\bar p$ of Appendix~\ref{app:budget-limits}.

\begin{proposition}[Execution across Strata: Necessity and Allocation]\label{prop:strata-limits}
In the setting above, the following hold.
\begin{enumerate}
\item[(i)] If $\xi\ge\zeta$, every procedure has power at most its size. If every $p_s+\gamma_s\le1$, then $\xi=\sum_sw_s\gamma_s$; otherwise $\sum_sw_s\gamma_s\ge\zeta$ does not suffice.
\item[(ii)] If $\zeta'>0$, then~\eqref{eq:strata-root} has a unique root $t_*>0$, $J_*$ is nonempty, and every valid and powerful procedure has $\mathbb E_{W_1}[C]\ge\mathrm{kl}(1-\beta,\alpha)/t_*$.
\item[(iii)] Let $\zeta'>0$. Then $t_*=\max_\lambda\inf_{r\in\mathcal A}\sum_s\lambda_s\mathrm{kl}(p_s,r_s)/c_s=\min_{r\in\mathcal A}\max_s\mathrm{kl}(p_s,r_s)/c_s$, with $\lambda$ ranging over cost shares ($\lambda_s\ge0$, $\sum_s\lambda_s=1$). The cost shares $\lambda^*_s=a_sc_s/\bar c$ and the vector $\bar p$ form a saddle point, every stratum in $J_*$ has $\mathrm{kl}(p_s,\bar p_s)=c_st_*$, and every $r\in\mathcal A$ has $\sum_{s\in J_*}w_s(r_s-\bar p_s)\ge0$.
\item[(iv)] Let every $p_s+\gamma_s<1$ and $q^\star\downarrow q+\xi$, and put $V=\sum_sw_s\sqrt{p_s(1-p_s)c_s}$. Then $1/t_*=2V^2\zeta'^{-2}\{1+O(\zeta')\}$, and $a_s$ tends to $w_s\sqrt{p_s(1-p_s)/c_s}/\sum_jw_j\sqrt{p_j(1-p_j)/c_j}$, the cost-weighted allocation of \citet{neyman1934two}. With $S=1$ and $c_1=1$, (i) and (ii) are Proposition~\ref{prop:budget-limits}(i) and the first bound of (ii). If $\sum_sw_s\gamma_s<\zeta$, then $1/t_*\le c_{\mathrm{pool}}/\mathrm{kl}(p_{\mathrm{pool}},p_{\mathrm{pool}}+(\zeta-\sum_sw_s\gamma_s)/m)$ with $p_{\mathrm{pool}}=\sum_sw_sp_s/m$ and $c_{\mathrm{pool}}=\sum_sw_sc_s/m$: the strata never raise the bound of Proposition~\ref{prop:budget-limits}(ii) applied to the pooled rate at the mean cost.
\item[(v)] Replace the strata of a set $I$ by one stratum with mass $\sum_{s\in I}w_s$, the mass-weighted means of $p_s$ and $\gamma_s$, and a cost at least the mass-weighted mean of $c_s$. Then $t_*$ does not increase, where $t_*=0$ when $\zeta'\le0$, the max-min value of (iii) in that case because then $p\in\mathcal A$.
\end{enumerate}
\end{proposition}

\begin{proof}
(i) Since $\min\{1,p_s+\gamma_s\}=p_s+\min\{\gamma_s,1-p_s\}$, the live harm allowed by $\hat p=p$ is $q+\xi$. If $\xi>\zeta$, the world with $\hat p=p$, $p_{0s}=\min\{1,p_s+\gamma_s\}$, and the record law of $W_1$ given harm is in the null and gives the log and every executed record the laws they have in $W_1$, so the power is at most the size. If $\xi=\zeta$, then $q+\xi=q^\star<q_0+m$, so some stratum $s'$ has $p_{s'}+\gamma_{s'}<1$. For small $\varepsilon>0$, raise only $\hat p_{s'}$ to $p_{s'}+\varepsilon$ and keep $p_{0s}=\min\{1,\hat p_s+\gamma_s\}$ and the record law of $W_1$ given harm; the live harm is $q^\star+w_{s'}\varepsilon$. The coupling of Proposition~\ref{prop:budget-limits}(i) changes at most $n$ outcomes, each with probability at most $\varepsilon$, and $\varepsilon\downarrow0$ bounds the power by the size. For the second claim, let $q_0=0$, $w=(1/2,1/2)$, $p=(0.95,0.05)$, $\gamma=(0.2,0)$, and $q^\star=0.58$. Then $\sum_sw_s\gamma_s=0.10>\zeta=0.08$, but $\xi=0.025$, and every world of the null has $\hat p_2>0.16$ because $p_{01}\le1$. With $n=4{,}000$, a test that executes 1,000 episodes of stratum~2 when the log holds that many, and certifies when at most 105 are harmful, has size below $e^{-2\cdot1000\cdot0.055^2}<0.0024$ and failure probability below $e^{-2\cdot1000\cdot0.056^2}+e^{-500}<0.0024$ (Hoeffding).
\par(iii) and (ii) Setting $r_s=p_s$ off $J$ and replacing $r_s$ by $\min\{1-\gamma_s,\max\{p_s,r_s\}\}$ on $J$ keep $r$ in $\mathcal A$ and raise no objective. So the inner problems run over $p_s\le r_s\le1-\gamma_s$ on $J$ with $\sum_{s\in J}w_s(r_s-p_s)\ge\zeta'$. Each $b_s(t)-p_s$ is continuous and nondecreasing from $0$, and strictly increasing until it reaches $1-\gamma_s-p_s$. The total capacity $\sum_{s\in J}w_s(1-\gamma_s-p_s)$ equals $q_0+m-q-\xi$ and exceeds $\zeta'$. So $t_*$ exists, is unique, and leaves $J_*$ nonempty, and $\bar p$ meets the constraint with equality. Let $r\in\mathcal A$. Strata outside $J_*$ contribute $w_s$ to the live harm at $\bar p$, the most they can, so $\sum_{s\in J_*}w_s[\min\{1,r_s+\gamma_s\}-(\bar p_s+\gamma_s)]\ge0$, and the bracket is at most $r_s-\bar p_s$; this is the last claim of (iii). Every stratum has $\mathrm{kl}(p_s,\bar p_s)\le c_st_*$, with equality on $J_*$, so $\max_s\mathrm{kl}(p_s,\bar p_s)/c_s=t_*$ and every $\lambda$ has inner value at most $t_*$. Convexity of $\mathrm{kl}(p_s,\cdot)$ and $\lambda^*_sg_s/c_s=a_sg_s/\bar c\propto w_s$ on $J_*$ give, for $r\in\mathcal A$,
\[
\begin{aligned}
&\textstyle\sum_s\frac{\lambda^*_s}{c_s}\mathrm{kl}(p_s,r_s)\\
&\ \ge\ \sum_{s\in J_*}\frac{a_s}{\bar c}\{c_st_*+g_s(r_s-\bar p_s)\}\\
&\ \ge\ t_*,
\end{aligned}
\]
so $\lambda^*$ has inner value $t_*$, and $\max_s\mathrm{kl}(p_s,r_s)/c_s\ge t_*$ on $\mathcal A$. This proves (iii) without exchanging the maximum and the infimum. For (ii), take $r$ with strict inequality in $\mathcal A$ and $0<r_s<1$, and the world of the null with $\hat p=r$, $p_{0s}=\min\{1,r_s+\gamma_s\}$, and the record law of $W_1$ given harm. An execution in stratum $s$ contributes relative entropy $\mathrm{kl}(p_s,r_s)$, and Lemma~1 of \citet{kaufmann2016complexity}, applied as in Proposition~\ref{prop:budget-limits}(ii), gives $\sum_s\mathbb E_{W_1}[K_s]\,\mathrm{kl}(p_s,r_s)\ge\mathrm{kl}(1-\beta,\alpha)$. With $B=\mathbb E_{W_1}[C]$ and $\lambda_s=c_s\mathbb E_{W_1}[K_s]/B$, this reads $B\sum_s\lambda_s\mathrm{kl}(p_s,r_s)/c_s\ge\mathrm{kl}(1-\beta,\alpha)$. Points $\bar p+\delta\mathbf 1_{J_*}$ are such $r$ for small $\delta>0$, so letting $\delta\downarrow0$ gives $B\max_s\mathrm{kl}(p_s,\bar p_s)/c_s\ge\mathrm{kl}(1-\beta,\alpha)$, that is, $B\,t_*\ge\mathrm{kl}(1-\beta,\alpha)$.
\par(iv) For small $\zeta'$, no coordinate reaches $1-\gamma_s$; Taylor expansion of binary relative entropy in \eqref{eq:strata-root} and $a_s\propto w_s/g_s$ gives both limits. With $S=1$ and $c_1=1$, the setting's $q^\star<q_0+m$ gives $\zeta<m(1-p)$, so $\xi=m\min\{\gamma,1-p\}\ge\zeta$ exactly when $m\gamma\ge\zeta$. If $m\gamma<\zeta$, then $\xi=m\gamma$, the root of \eqref{eq:strata-root} is $\bar p_1=p+(\zeta-m\gamma)/m$, and $t_*=\mathrm{kl}(p,\bar p)$. For pooling, choose $\lambda_s\propto w_sc_s$ in (iii). Joint convexity gives $\sum_s(w_s/m)\mathrm{kl}(p_s,r_s)\ge\mathrm{kl}(p_{\mathrm{pool}},\sum_sw_sr_s/m)$, where feasibility ensures $\sum_sw_sr_s/m\ge p_{\mathrm{pool}}+(\zeta-\sum_sw_s\gamma_s)/m>p_{\mathrm{pool}}$; monotonicity gives the bound.
\par(v) The claim is immediate if the merged $t_*=0$. Otherwise split the optimal share of merged stratum $\iota$ as $\lambda_s=\lambda'_\iota w_sc_s/\sum_{j\in I}w_jc_j$. Map feasible $r$ to $r'$ by mass-weighted averaging on $I$; concavity of $\min\{1,\cdot\}$ preserves feasibility. Joint convexity and the cost condition give $\sum_{s\in I}\lambda_s\mathrm{kl}(p_s,r_s)/c_s\ge\lambda'_\iota\mathrm{kl}(p_\iota,r'_\iota)/c_\iota$, so (iii) bounds the split value below by the merged value.
\end{proof}

\paragraph{Achievability.} The \emph{stratified execution test} with levels $\alpha,\beta'\in(0,1)$ draws stratum $s\in J_*$ with probability $a_s$, independently of the past, executes an unused unresolved episode, and reads its harm bit $Y$. Like the test of Proposition~\ref{prop:budget-achievability}, it uses $p$ and the log's masses. Starting at $\Xi_0=0$, each draw adds score $Y\log(p_s/\bar p_s)+(1-Y)\,\mathrm{up}(p_s,\bar p_s)$. The test certifies when $\Xi_t\ge\log(1/\alpha)$, and stops without certifying when $\Xi_t\le-\log(1/\beta')$ or the drawn stratum pool is exhausted. Set $\bar u=\max_{s\in J_*}\mathrm{up}(p_s,\bar p_s)$, $\bar K=\{\log(1/\alpha)+\bar u\}/(t_*\bar c)$, and $\chi_n=\sum_{s\in J_*}\{2a_s\bar K/(w_sn)+e^{-w_s^2n/2}\}$.

\begin{proposition}[Execution across Strata: Achievability and Live Crossover]\label{prop:strata-achieve}
Assume the setting of Proposition~\ref{prop:strata-limits} with $\zeta'>0$.
\begin{enumerate}
\item[(i)] For every log, every restored vector in $\mathcal A$, and every record law given harm, the test certifies with probability at most $\alpha$. This also holds when the design point is computed from design values $\tilde p\in(0,1)^S$ in place of $p$, provided $\zeta'>0$ at $\tilde p$, since $\mathcal A$ does not involve $p$.
\item[(ii)] With unlimited pools, at $W_1$ the test stops at the lower threshold with probability at most $\beta'$, uses $\mathbb E_{W_1}[\tau]\le\bar K$ draws, and costs $\mathbb E_{W_1}[C]\le\{\log(1/\alpha)+\bar u\}/t_*$. With the pools of a log of $n$ episodes, its power is at least $1-\beta'-\chi_n$ and the cost bound holds.
\item[(iii)] Let $C_{\mathrm{ex}}$ be the least $\mathbb E_{W_1}[C]$ over valid and powerful procedures, and $C_{\mathrm{lv}}$ the least expected cost, at $c_L$ per episode, of the live tests of Proposition~\ref{prop:budget-limits}\textup{(iii)}. If $\chi_n<\beta$, then $\mathrm{kl}(1-\beta,\alpha)/t_*\le C_{\mathrm{ex}}\le\{\log(1/\alpha)+\bar u\}/t_*$. As $\alpha,\beta\downarrow0$ with $\chi_n<\beta$, $C_{\mathrm{ex}}/C_{\mathrm{lv}}\to\mathrm{kl}(q,q^\star)/(c_Lt_*)$. So executing the unresolved continuations is cheaper to first order exactly when $t_*>t_L=\mathrm{kl}(q,q^\star)/c_L$, that is, when the left side of~\eqref{eq:strata-root} at $t_L$ is below $\zeta'$.
\end{enumerate}
\end{proposition}

\begin{proof}
Use the score identities and SPRT argument of Proposition~\ref{prop:budget-achievability}, with $(a,c)=(p_s,\bar p_s)$ in stratum $s$.
\par(i) Given the log, past, and stratum $s$, the bit is Bernoulli$(\hat p_s)$ whatever its record law. For $\hat p\in\mathcal A$, Proposition~\ref{prop:strata-limits}(iii) and $a_sg_s\propto w_s$ give $\mathbb E[e^{Z_t}\mid\text{past}]=1-\sum_{s\in J_*}a_sg_s(\hat p_s-\bar p_s)\le1$. Extend each pool by independent Bernoulli$(\hat p_s)$ bits; finite-pool certification requires the unlimited supermartingale $e^{\Xi_t}$ to cross $1/\alpha$. Ville's inequality gives size at most $\alpha$, also for $\tilde p$, since the argument uses only the design point's properties relative to $\mathcal A$.
\par(ii) At $W_1$, the mean score is $t_*\bar c$ and upper overshoot is at most $\bar u$. The same SPRT argument gives lower-exit probability at most $\beta'$, $\mathbb E[\tau]\le\bar K$, and mean cost $\bar c\mathbb E[\tau]$. Wald's identity gives mean draws $a_s\mathbb E[\tau]\le a_s\bar K$ in stratum $s$. The same Markov and Hoeffding bounds, summed over strata, bound exhaustion by $\chi_n$; together with lower exit this gives the power, while exhaustion only reduces cost.
\par(iii) Combine Proposition~\ref{prop:strata-limits}(ii) with (i) and (ii), taking $\beta'=\beta-\chi_n$, for the execution bounds. The live bounds of Propositions~\ref{prop:budget-limits}(iii) and~\ref{prop:budget-achievability}(iii), and $\mathrm{kl}(1-\beta,\alpha)/\log(1/\alpha)\to1$, give the cost ratio. Strict increase of the left side of~\eqref{eq:strata-root} on $[0,t_*]$ gives the crossover equivalence.
\end{proof}

\begin{remark}[Allocation, Pools, and Snapshots]\label{rem:strata-scope}
The shares $a$ are part of the test's validity, not only of its cost. In the instance of \texttt{checks/verify\_budget\_strata.py}, the same increments with draws in proportion to $w_s$ have a per-draw mean of $e^Z$ above one at a boundary point of $\mathcal A$, and their size there is about 0.15 at $\alpha=0.05$. Design values $\tilde p\in(0,1)^S$ from a pilot on separate episodes, such as smoothed proportions, keep the size at $\alpha$ by part~(i), with no certification when $\zeta'\le0$ at $\tilde p$; we do not analyze the cost of such an adaptive rule \citep[compare][]{garivier2016optimal}. The bounds concern population harm under independent outcomes, with power at exact restoration; fixed-cohort harm fractions require a sampling design such as Proposition~\ref{prop:budget}. Power at every approximately restored law is not guaranteed. With $q_0$ and $w$ estimated, replacing them in $\mathcal A$ by simultaneous upper confidence limits of level $\alpha_{\log}$ enlarges $\mathcal A$ on the event of the limits. Running the test only when $q_0^U+\sum_sw_s^U\min\{1,\tilde p_s+\gamma_s\}<q^\star<q_0^U+\sum_sw_s^U$ for these limits $q_0^U$ and $w^U$, and otherwise not certifying, keeps the size at most $\alpha+\alpha_{\log}$, as in Remark~\ref{rem:estimate-log}.

Runs from an earlier snapshot re-execute the agreeing steps (Remark~\ref{rem:design-evidence}). Suppose they reach the first-deviation strata of a set $I$ with probabilities $w_s/\sum_{j\in I}w_j$ in every world, with the same restored harm laws and admissible rates; that their records can be simulated from the stratum and one restored harm bit by a randomization that does not depend on the world; and that they cost at least the mean cost of runs started at those deviations. Then data processing and the share splitting of Proposition~\ref{prop:strata-limits}(v) show that restoring at the first deviations of Proposition~\ref{prop:snapshot-design} never raises the bound of part~(ii). The comparison does not cover earlier runs that observe more, restore differently, or route differently, and choosing coarser snapshots trades storage against the bound, which we do not solve.
\end{remark}

\paragraph{Related Tests.} With unit costs and no restoration error, $\mathcal A$ is a half-space of mean vectors, and the saddle point of Proposition~\ref{prop:strata-limits}(iii) is the half-space case of \citet[Theorem~5]{juneja2019sample}, whose least favorable alternative also equalizes the divergences. Multi-fidelity best-arm identification pairs cheaper samples of known maximal bias with their costs \citep{poiani2024multifidelity}. Stratified sequential audits permit arbitrary stratum selection \citep{spertus2022sweeter,spertus2024sequential}; this likelihood-ratio test requires its design shares.

\subsection{Snapshot Design for a Family of Gates}
\label{app:design}

This proves the snapshot-design claim of Section~\ref{sec:supported}.

\paragraph{Setting and Evidence.} We adopt the framework of Appendix~\ref{app:replay-setting} with query $X=\phi(L)$ and pre-intervention record $r$, where $f^0$ executes each call. Decisions represent steps in the demonstrations; gates acting on each call treat calls as decisions (via the refined clock of Appendix~\ref{app:formal-spec}). Let $m(h)=m_{t(h)}$ be the score in $I(h)$, and $\bar m(h)$ the maximal score over previous decision histories $h'\prec h$ with $t(h')\ge s$ for boundary step $s$ ($-\infty$ if none). The retry gate $\kappa_\theta$ of Section~\ref{sec:running} replaces the call by a refusal $a_\theta(i)\ne f^0(i)$ when $t\ge s$ and $m(i)>\theta$. A refusal at final step $T$ constitutes an abort, as no decision history follows (Appendix~\ref{app:replay-setting}). At each step $t<T$, a decision history follows the refusal.

Write $w_h=P^0([h]\mid r)$, $\mathcal D^+_\kappa=\mathcal D^{\mathrm{cont}}_\kappa(r,P^0)$, and $Q^\kappa_h=\phi_\#P^\kappa(L\in\cdot\mid[h])$ for $h\in\mathcal D^+_\kappa$. Feasible set $\mathcal X^\kappa_h$ follows Appendix~\ref{app:replay-cont} for gate $\kappa$, depending on $\kappa$ via subsequent refusals. A history $h\in\mathcal D^+_\kappa$ is \emph{open for $\kappa$} when $|\mathcal X^\kappa_h|\ge2$ and closed otherwise (for instance, when $X$ is the harm indicator and harm accrued before $h$). Let $\mathcal O_\kappa$ collect histories open for $\kappa$.

A set $\mathcal S$ of logged decision histories $h$ with $[h]\subseteq\Omega_r$ and $P^0([h])>0$ stores \emph{first-deviation snapshots}. For $h\in\mathcal S$ and gate $\kappa$ with $h\in\mathcal D^+_\kappa$, the evaluator restores at $h^+_\kappa$ and executes the continuation. Assumption \textup{(T)} equates this continuation law with $Q^\kappa_h$, estimated by finite executions (Proposition~\ref{prop:budget}). Earlier runs re-execute agreeing steps as partial live runs (Remark~\ref{rem:design-evidence}). Condition \textup{(U)} of Section~\ref{sec:supported} admits arbitrary response laws at inputs of zero baseline probability, taking model class $\mathcal M_{\mathrm{full}}$ from Lemma~\ref{lem:app-unrestricted}. Its mechanisms satisfy \textup{(R)} and \textup{(A1)} to \textup{(A3)}, with unreplaced kernels reading the full history, so \textup{(G1)} fails at refusal inputs. The proof of \textup{(c)} does not require realizability.

\begin{proposition}[Snapshots for a Threshold Family]\label{prop:snapshot-design}
In \textup{(a)} and \textup{(b)}, let $h$ be a decision history with $s\le t(h)<T$, $[h]\subseteq\Omega_r$, and $P^0([h])>0$.
\par\textup{(a)}~$h\in\mathcal D^+_{\kappa_\theta}$ exactly when $\bar m(h)\le\theta<m(h)$, so $h$ is a first deviation for some $\theta\in\Theta$ exactly when $\Theta\cap[\bar m(h),m(h))\ne\emptyset$.
\par\textup{(b)}~For $\Theta=[\theta_{\min},1)$, this means $m(h)>\max\{\theta_{\min},\bar m(h)\}$, a \emph{running maximum above $\theta_{\min}$}, and the gates $\theta\in[\max\{\theta_{\min},\bar m(h)\},m(h))$ deviate at $h$. Ties and a score equal to $\theta_{\min}$ give no running maximum, and a score of $1$ is one exactly when all earlier scores since $s$ are below $1$. For a finite $\Theta=\{\theta_1<\dots<\theta_n\}$, the condition is $B(m(h))>B(\bar m(h))$ with $B(x)=\#\{i:\theta_i<x\}$, and an episode has at most $n$ first deviations.
\par\textup{(c)}~Assume \textup{(R)}, \textup{(A1)} to \textup{(A3)}, \textup{(U)}, and \textup{(T)}. Given first-deviation snapshots $\mathcal S$, the identified set of $\phi_\#P^\kappa(L\in\cdot\mid r)$ is $\{\nu+\sum_{u\in\mathcal D^+_\kappa\cap\mathcal S}w_uQ^\kappa_u+\sum_{u\in\mathcal D^+_\kappa\setminus\mathcal S}w_uQ_u:Q_u\in\Delta(\mathcal X^\kappa_u)\}$ ($\nu$ as in Proposition~\ref{prop:app-identified}). It is a singleton exactly when $\mathcal O_\kappa\subseteq\mathcal S$; otherwise, for binary $X$, the set for $P^\kappa(X=1\mid r)$ has width $\sum_{u\in\mathcal O_\kappa\setminus\mathcal S}w_u$. So $\mathcal S$ identifies every $\kappa_\theta$, $\theta\in\Theta$, exactly when it contains $\bigcup_\theta\mathcal O_{\kappa_\theta}$, the least such set. For $[\theta_{\min},1)$, these are the running maxima above $\theta_{\min}$ at steps $t<T$ that are open for some gate deviating there.
\end{proposition}

\begin{proof}
(a) This is the first-deviation condition by definition, with continuation because $t(h)<T$.

(b) Intersecting $[\bar m(h),m(h))$ with $\Theta$ gives the stated cases. For finite $\Theta$, $B(m)-B(\bar m)$ counts the available thresholds, and each threshold has at most one first deviation per episode.

(c) By Proposition~\ref{prop:app-identified}, a compatible model has law $\nu+\sum_uw_uQ^M_u$ with $Q^M_u\in\Delta(\mathcal X^\kappa_u)$, and \textup{(T)} fixes $Q^M_u=Q^\kappa_u$ for $u\in\mathcal D^+_\kappa\cap\mathcal S$. Conversely, let $M^*\in\mathcal M_{\mathrm{full}}$ generate the evidence and fix $Q_u\in\Delta(\mathcal X^\kappa_u)$ for $u\in\mathcal D^+_\kappa\setminus\mathcal S$. For each $x\in\mathcal X^\kappa_u$, take a replacement-consistent extension $\omega_x$ of $u^+_\kappa$ with $\phi(\ell(\omega_x))=x$. By \textup{(U)}, set the kernels inside $[u^+_\kappa]$ to the conditional laws of $\sum_xQ_u(x)\delta_{\omega_x}$.

These cylinders are $P^0$-null, so kernels set inside them do not enter the baseline law, which the modification therefore preserves. They are disjoint from $[g^+_{\kappa'}]$ for every gate $\kappa'$ of the family and every $g\in\mathcal S\cap\mathcal D^+_{\kappa'}$ or terminal $g$ of positive mass. If $u\ne g$ and they meet, one of $u,g$ extends a deviating output of the other, which has baseline probability zero. A terminal $g=u$ has an abort output, whereas $u^+_\kappa$ continues. Thus the modification also preserves the stored laws and the valuations of \textup{(A3)}, and attains the chosen law. The set is a constant plus $w_u\Delta(\mathcal X^\kappa_u)$ for each $u\notin\mathcal S$, which gives the singleton criterion and the width.
\end{proof}

\begin{remark}[Broader Evidence]\label{rem:design-evidence}
The least-set claim concerns restoration at the first deviation of each gate. Suppose instead that $\mathcal S$ holds logged decision histories $g$ with $[g]\subseteq\Omega_r$ and $P^0([g])>0$. Assume that \textup{(T)} links a run of $\kappa$ from $g$ to the changed system only for gates that reach $g$, those that agree with $f^0$ at every earlier decision. Such a run re-executes the agreeing steps below $g$, and its law determines only the sum $\sum_{h\succeq g}w_hQ^\kappa_h$ over the first deviations $h$ of $\kappa$ below $g$, up to known terms. Under the other assumptions of \textup{(c)}, $\kappa$ is then identified exactly when each $h\in\mathcal O_\kappa$ has an ancestor-or-self in $\mathcal S$, by the proof of (c). Minimal sets are then not unique. With the trivial record and a decision at step $0$ on every path, storing every step-$0$ history (the live-run cover) identifies every gate with no snapshot at a running maximum. Re-executing every agreeing step is the price. The claim above is the least set of first-deviation snapshots.
\end{remark}

\section{Proofs: Component Evidence and Composition}
\label{app:p3-proof}

This appendix proves Proposition~\ref{prop:component-joint}, used in Sections~\ref{sec:joint} and~\ref{sec:cues}.

\subsection{Statement}
\label{app:p3-full}

Section~\ref{sec:joint} states this result in words. Fix a target composition and scheduler, with shared state $S\sim P_S$, component losses $L_i$ with conditional laws $F_i(\cdot\mid s)$, and declared joint loss $L=h(S,L_1,\ldots,L_n)$; write $p_i(s)=F_i(B_i\mid s)$. Section~\ref{app:p3-setting} makes these objects precise.

\begin{proposition}[Component evidence to joint risk]\label{prop:component-joint}
\textup{(i)} If $L_1,\dots,L_n$ are mutually conditionally independent given $S$ \textup{(CI)}, the law of $(S,L_1,\dots,L_n)$ is $P_S(ds)\prod_iF_i(dl_i\mid s)$, and the law of $L$ is its image under $h$.
\textup{(ii)} Let the evidence be $P_S$ and the $F_i$, as from separate component logs that each carry $S$, with no restriction on dependence given $S$ \textup{(NR)}. Then $P(L_1\in B_1,L_2\in B_2)$ takes exactly the values in $\bigl[\int(p_1+p_2-1)_+\,dP_S,\ \int\min(p_1,p_2)\,dP_S\bigr]$.
\textup{(iii)} Records linking $S$ and every $L_i$ within each episode identify the law of $(S,L_1,\dots,L_n)$, hence every query of $L$. Under \textup{(NR)}, separate logs of each $(S,L_i)$ identify that law if and only if, at each state of positive probability, at most one $F_i(\cdot\mid s)$ is not a point mass. Records of $S$ alone identify only $P_S$ in the family of Lemma~\ref{lem:p3-realize}.
\textup{(iv)} Standalone runs of component $i$ identify $F_i(\cdot\mid s)$ when the composition's state law is known \textup{(Known $P_S$)}, the conditional law of $L_i$ given $S$ is the same in standalone runs and in the composition \textup{(Inv$_i$)}, and every composed state occurs in standalone runs \textup{(Supp$_i$)}. In a family that does not tie the composed component to its standalone run, the standalone law of $(S,L_i)$ places no restriction on $F_i$. Without \textup{(Supp$_i$)}, let component $i$ read only $S$ and private randomness, and let $U_i$ hold the composed states never seen in isolation. For $\emptyset\neq B\subsetneq\lossset_i$, the identified set of $P(L_i\in B)$ is then an interval of width $P_S(U_i)$, and other restrictions on component $i$ can narrow it.
\end{proposition}

Parts (i) to (iii) restate conditional independence, Fr\'echet bounds given $S$ \citep{frechet1951,nelsen2006copulas}, and linked-record identification; part (iv) gives transportability with overlapping support \citep{pearl2011transportability}.

\subsection{Finite Setting and Composition Interfaces}
\label{app:p3-setting}
\label{sec:coupling}

This setting supplies the objects and assumptions for Proposition~\ref{prop:component-joint} (Section~\ref{sec:joint}).

A composition declares interfaces, a scheduler, compatible kernels, and joint loss map $h:\mathcal S\times\lossset_{1:n}\to\lossset\subset\mathbb R^m$.
This defines joint specification $\riskmodel_{\mathrm{joint}}=(E_{\mathrm{joint}},\ell_{\mathrm{joint}},K_{\mathrm{joint}},\interventions_{\mathrm{joint}})$ with law $P$ on finite trajectory space $\Omega$.

Shared state is a declared map $S:\Omega\to\mathcal S$, with finite $\mathcal S$.
Component $i\in\{1,\dots,n\}$ has loss map $L_i:\Omega\to\lossset_i$ with finite $\lossset_i$.
Write $\mathbf L=(L_1,\dots,L_n)\in\lossset_{1:n}$, consequence map $\ell_{\mathrm{joint}}=h\circ(S,\mathbf L)$, and full-episode loss $L=h(S,\mathbf L)$.
Let $\Lambda_P(s,\mathbf l)=P(S=s,\mathbf L=\mathbf l)$ be the law of $(S,\mathbf L)$, with state marginal $P_S(s)=\sum_{\mathbf l}\Lambda_P(s,\mathbf l)$, support $\mathcal S_+=\{s:P_S(s)>0\}$, and conditional laws $F_i(l\mid s)=P(L_i=l\mid S=s)$ for $s\in\mathcal S_+$.
For events $B_i\subseteq\lossset_i$, write $E_i=\{L_i\in B_i\}$ and $p_i(s)=F_i(B_i\mid s)$.
The product law is
\begin{equation}
\Lambda^{\otimes}(s,\mathbf l)=P_S(s)\prod_{i=1}^nF_i(l_i\mid s).
\label{eq:p3-product}
\end{equation}
For component $i$, standalone deployment is a mechanism with law $P^{\mathrm{sa}}_i$ yielding evidence $Q^{(i)}=\Law_{P^{\mathrm{sa}}_i}(S,L_i)$, state marginal $Q^{(i)}_S$, and support $\mathcal S^{\mathrm{sa}}_i=\{s:Q^{(i)}_S(s)>0\}$.
The model family $\mathcal M_0$ contains finite composition mechanisms under the scheduler, with true mechanism $M^\star\in\mathcal M_0$ having $P_{M^\star}=P$ and $\Lambda_{M^\star}=\Lambda_P$.
Identified sets are $\Theta_e(\theta)=\{\theta(M):M\in\mathcal M_0,\ e(M)=e(M^\star)\}$.
Evidence maps are linked records $e_{\mathrm{link}}(M)=\Law_{P_M}(S,\mathbf L)$, paired records $e_{\mathrm{pair}}(M)=(\Law_{P_M}(S,L_1),\dots,\Law_{P_M}(S,L_n))$, and state records $e_S(M)=\Law_{P_M}(S)$.

\begin{definition}[Coupling set]\label{def:p3-coupling}
$\Gamma=\Gamma(P_S;F_1,\dots,F_n)$ is the set of probability laws $\Lambda$ on $\mathcal S\times\lossset_{1:n}$ such that, for every $s\in\mathcal S$, every $i$, and every $l\in\lossset_i$,
\begin{equation*}
\sum_{\mathbf l:\,l_i=l}\Lambda(s,\mathbf l)=P_S(s)\,F_i(l\mid s).
\end{equation*}
\end{definition}

Summing over $l$ shows every $\Lambda\in\Gamma$ has state marginal $P_S$, and for each $s\in\mathcal S_+$, $\pi_s=\Lambda(s,\cdot)/P_S(s)$ couples $F_1(\cdot\mid s),\dots,F_n(\cdot\mid s)$; thus $\Lambda^{\otimes},\Lambda_P\in\Gamma$.

\begin{definition}[Assumptions]\label{def:p3-assumptions}
\par\textup{(CI)} $L_1,\dots,L_n$ are mutually conditionally independent given $S$ under $P$: $P(\mathbf L=\mathbf l\mid S=s)=\prod_iF_i(l_i\mid s)$ for every $s\in\mathcal S_+$ and every $\mathbf l$.
\par\textup{(NR)} No dependence restriction: $\Gamma(P_S;F_1,\dots,F_n)\subseteq\{\Lambda_M:M\in\mathcal M_0\}$.
\par\textup{(Known $P_S$)} The shared-state law of the target composition is available, from logged $S$ in that composition or from its declared scheduler and upstream kernels.
\par\textup{(Inv$_i$)} There is a kernel $G_i$ from $\mathcal S$ to $\lossset_i$ with $F_i(\cdot\mid s)=G_i(\cdot\mid s)$ for every $s\in\mathcal S_+$ and $Q^{(i)}(L_i\in\cdot\mid S=s)=G_i(\cdot\mid s)$ for every $s\in\mathcal S^{\mathrm{sa}}_i$.
\par\textup{(Supp$_i$)} $\mathcal S_+\subseteq\mathcal S^{\mathrm{sa}}_i$.
\end{definition}

\subsection{Lemmas}

These lemmas support the proof of Proposition~\ref{prop:component-joint} (Section~\ref{sec:joint}).

\begin{lemma}[Paired evidence fixes exactly the coupling set]\label{lem:p3-paired}
For $M\in\mathcal M_0$, $e_{\mathrm{pair}}(M)=e_{\mathrm{pair}}(M^\star)$ if and only if $\Lambda_M\in\Gamma(P_S;F_1,\dots,F_n)$, where $P_S$ and the $F_i$ are those of $M^\star$.
\end{lemma}
\begin{proof}
The marginal conditions $\Law_{P_M}(S,L_i)(s,l)=\sum_{\mathbf l:\,l_i=l}\Lambda_M(s,\mathbf l)=P_S(s)F_i(l\mid s)$ for all $i,s,l$ define $\Gamma$.
\end{proof}

\begin{lemma}[Realization by an unlogged shared input]\label{lem:p3-realize}
Let $\mathcal M_0^W$ be the family of mechanisms in which a shared-state kernel draws $S$, an unlogged exogenous input $V$ on a finite set is drawn from a kernel given $S$, and each component $i$, in the scheduled order, outputs $L_i=k_i(S,V)$ for a deterministic map $k_i$. Every law $\Lambda$ on $\mathcal S\times\lossset_{1:n}$ is $\Lambda_M$ for some $M\in\mathcal M_0^W$. In particular $\mathcal M_0^W$ satisfies (NR) at every $(P_S,F_1,\dots,F_n)$.
\end{lemma}
\begin{proof}
Draw $S\sim\Lambda_S$, draw $V\sim\Lambda(\cdot\mid S)$, and set $k_i(s,\mathbf v)=v_i$.
\end{proof}

\begin{lemma}[When the coupling set is a point]\label{lem:p3-singleton}
$\Gamma$ is a singleton if and only if, for every $s\in\mathcal S_+$, at most one of $F_1(\cdot\mid s),\dots,F_n(\cdot\mid s)$ is not a point mass.
\end{lemma}
\begin{proof}
If at most one margin is non-degenerate at each $s\in\mathcal S_+$, the marginal constraints uniquely fix $\Lambda(s,\cdot)$. Otherwise, choose two distinct positive-mass values in each of two non-degenerate margins at some $s\in\mathcal S_+$, and fix positive-mass values in all other margins. The four resulting cells have positive mass under $\prod_iF_i(\cdot\mid s)$. Add $\epsilon>0$ to the two diagonal cells and subtract it from the other two, with $\epsilon$ smaller than their masses. This preserves every margin and non-negativity; multiplying by $P_S(s)$ and leaving other states unchanged gives a second member of $\Gamma$.
\end{proof}

\subsection{Proofs of Proposition~\ref{prop:component-joint} Parts (i) to (iv)}

\paragraph{Proof of Part (i).}
For $s\in\mathcal S_+$ and $\mathbf l$, conditional probability and (CI) give $\Lambda_P(s,\mathbf l)=P_S(s)\prod_i F_i(l_i\mid s)=\Lambda^{\otimes}(s,\mathbf l)$, vanishing outside $\mathcal S_+$.
Because $L=h(S,\mathbf L)$, $\Law_P(L)=h_{\#}\Lambda^{\otimes}$, so every query of $L$ is a function of $(P_S,F_1,\dots,F_n)$.
Paired records cannot test (CI) because $\Lambda^{\otimes}\in\Gamma$ and part (iii)(b) shows they exclude no element of $\Gamma$ under (NR).

\begin{remark}[Sub-parts of part (ii)]\label{rem:p3-part-ii}
Let $\beta^-(s)=(p_1(s)+p_2(s)-1)_+$ and $\beta^+(s)=\min(p_1(s),p_2(s))$ for $s\in\mathcal S_+$. For a function $\beta$ on $\mathcal S_+$, write $\bar\beta=\sum_sP_S(s)\beta(s)$; in particular $\bar\beta^-=\sum_sP_S(s)\beta^-(s)$ and $\bar\beta^+=\sum_sP_S(s)\beta^+(s)$.
(a) Every $\Lambda\in\Gamma$, in particular $\Lambda_P$, satisfies $\beta^-(s)\le\Lambda(E_1\cap E_2\mid S=s)\le\beta^+(s)$ for $s\in\mathcal S_+$, and $\bar\beta^-\le\Lambda(E_1\cap E_2)\le\bar\beta^+$.
(b) Under (NR), $\Theta_{e_{\mathrm{pair}}}(P(E_1\cap E_2))=[\bar\beta^-,\bar\beta^+]$.
(c) $(P(E_1)+P(E_2)-1)_+\le\bar\beta^-$ and $\bar\beta^+\le\min(P(E_1),P(E_2))$.
(d) Under (NR), the identified set for the law of $L$ under paired records is $\mathcal H=\{h_{\#}\Lambda:\Lambda\in\Gamma\}$, and that of a query $(\phi,\rho)$ is $\{\rho(\phi_{\#}\nu):\nu\in\mathcal H\}$. If $n=2$ and $h(s,l_1,l_2)=\bar h(\mathbf 1_{B_1}(l_1),\mathbf 1_{B_2}(l_2))$ does not depend on $s$, each element $h_\#\Lambda$ of $\mathcal H$ depends on $\Lambda$ only through $\bar\beta=\Lambda(E_1\cap E_2)$, which ranges over $[\bar\beta^-,\bar\beta^+]$.
\end{remark}

\paragraph{Proof of Part (ii).}
(a) At each $s\in\mathcal S_+$, apply the Fr\'echet bounds \citep{frechet1951}, in \citet[Theorem~2.2.3 and Eq.~(2.5.1)]{nelsen2006copulas}, to $1-\mathbf 1_{B_1}(L_1)$ and $1-\mathbf 1_{B_2}(L_2)$ conditionally on $S=s$, evaluating their joint CDF at $(0,0)$. These finite conditional laws have CDF values $p_1(s),p_2(s)$ there, giving $\beta^\pm(s)$; averaging gives $\bar\beta^\pm$. (c) follows by summing $(x)_+\ge x,0$ and $\min(x,y)\le x,y$ against $P_S$.
(b) The bounds in Nelsen's Eq.~(2.5.1) are themselves joint CDFs, including for discrete margins. Mix the two endpoint indicator couplings with a common weight $\lambda\in[0,1]$ at every $s$. Lift each positive indicator cell to the full losses by taking the product of the $F_i(\cdot\mid s)$ conditioned on that cell's indicators; positive cell mass ensures these conditional laws exist. Give components $i\ge3$ their independent margins. This yields a member of $\Gamma$ with intersection mass $(1-\lambda)\bar\beta^-+\lambda\bar\beta^+$. By (NR) and Lemma~\ref{lem:p3-paired}, each such coupling has a compatible mechanism, proving sharpness and every intermediate value.
(d) By Lemma~\ref{lem:p3-paired} and (NR), the consistent $(S,\mathbf L)$-laws form exactly $\Gamma$, and $\Law(L)=h_{\#}\Lambda$. Let $n=2$ and let $\bar h$ be an indicator map that does not depend on $s$. The law of $(\mathbf 1_{B_1}(L_1),\mathbf 1_{B_2}(L_2))$ under $\Lambda\in\Gamma$ has the fixed margins $P(E_1)$ and $P(E_2)$ and joint cell $\bar\beta=\Lambda(E_1\cap E_2)$. Hence $h_\#\Lambda$ depends on $\Lambda$ only through $\bar\beta$, which ranges over $[\bar\beta^-,\bar\beta^+]$ by (b).

\paragraph{Proof of Part (iii).}
(a) Linked records log $(S,\mathbf L)$ jointly, so in every family $e_{\mathrm{link}}(M)=e_{\mathrm{link}}(M^\star)$ implies $\Lambda_M=\Lambda_P$, which identifies every query of $L$ through $\Law_P(L)=h_{\#}\Lambda_P$. (b) By Lemma~\ref{lem:p3-paired}, $\Theta_{e_{\mathrm{pair}}}(\Lambda)=\{\Lambda_M:M\in\mathcal M_0\}\cap\Gamma$, which equals $\Gamma$ under (NR). (c) $e_S(M)$ is the state marginal of each component of $e_{\mathrm{pair}}(M)$, so $\Theta_{e_{\mathrm{pair}}}(\Lambda)\subseteq\Theta_{e_S}(\Lambda)$. The family $\mathcal M_0^W$ of Lemma~\ref{lem:p3-realize} realizes every law with state marginal $P_S$ and satisfies (NR). In it, $e_S(M)=e_S(M^\star)$ holds exactly when $\Lambda_M$ has state marginal $P_S$, and two elements $\mathbf l\neq\mathbf l'$ of $\lossset_{1:n}$ give the distinct members $\sum_sP_S(s)\delta_{(s,\mathbf l)}$ and $\sum_sP_S(s)\delta_{(s,\mathbf l')}$, so records of $S$ alone identify only $P_S$. (d) Lemma~\ref{lem:p3-singleton} combined with (b) shows that, under (NR), paired records identify $\Lambda_P$ exactly when, for every $s\in\mathcal S_+$, at most one $F_i(\cdot\mid s)$ is not a point mass.

\begin{remark}[Sub-parts of part (iv)]\label{rem:p3-part-iv}
Assume (Known $P_S$). (a) Under (Inv$_i$) and (Supp$_i$), $F_i(\cdot\mid s)=Q^{(i)}(L_i\in\cdot\mid S=s)$ for every $s\in\mathcal S_+$. If this holds for every $i$, then under (CI) the law of $L$ is $h_{\#}\Lambda^{\otimes}$ computed from standalone evidence and $P_S$, and without (CI) the bounds of Remark~\ref{rem:p3-part-ii}(a) hold with these $F_i$.
(b) Let $U_i=\mathcal S_+\setminus\mathcal S^{\mathrm{sa}}_i$, and let $\mathcal M_0^{\mathrm{tr},i}$ be the family of pairs of a composition and a standalone deployment in which component $i$ is one kernel $G_i(\cdot\mid S)$, reading only $S$ and private randomness in both, with $G_i$ arbitrary, composition state law $P_S$, and standalone state law $Q^{(i)}_S$. In this family (Inv$_i$) holds, and for every $B$ with $\emptyset\neq B\subsetneq\lossset_i$, the identified set of $P(L_i\in B)$ given $Q^{(i)}$ and $P_S$ is
\begin{equation*}
\begin{gathered}
\bigl[c_B,\ c_B+P_S(U_i)\bigr],\\
\qquad c_B=\sum_{s\in\mathcal S_+\setminus U_i}P_S(s)\,Q^{(i)}(L_i\in B\mid S=s).
\end{gathered}
\end{equation*}
It is a singleton if and only if (Supp$_i$) holds.
(c) Without (Inv$_i$), $Q^{(i)}$ places no restriction on $F_i$: for every family of laws $(F'_i(\cdot\mid s))_{s\in\mathcal S_+}$ on $\lossset_i$, some composition with state law $P_S$ has these conditional laws while its standalone deployment has evidence $Q^{(i)}$.
\end{remark}

\paragraph{Proof of Part (iv).}
(a) Under (Supp$_i$), each $s\in\mathcal S_+$ lies in $\mathcal S^{\mathrm{sa}}_i$, so (Inv$_i$) gives $F_i(\cdot\mid s)=G_i(\cdot\mid s)=Q^{(i)}(L_i\in\cdot\mid S=s)$. With (Known $P_S$), $\Lambda^{\otimes}$ is computable, part (i) gives $\Law_P(L)=h_{\#}\Lambda^{\otimes}$ under (CI), and Remark~\ref{rem:p3-part-ii}(a) applies without (CI) because it uses only membership in $\Gamma$.
(b) Component $i$ reads only $S$ and private randomness in both deployments, so (Inv$_i$) holds in $\mathcal M_0^{\mathrm{tr},i}$. For $s\in\mathcal S_+\setminus U_i$, (Inv$_i$) fixes $F_i(B\mid s)=Q^{(i)}(L_i\in B\mid S=s)$, contributing $c_B$, while $G_i(B\mid s)$ is unconstrained in $[0,1]$ on $U_i$, so $P(L_i\in B)=c_B+\sum_{s\in U_i}P_S(s)G_i(B\mid s)\in[c_B,c_B+P_S(U_i)]$. For attainment, fix $x_{\mathrm{in}}\in B$ and $x_{\mathrm{out}}\notin B$, set $G_i(\cdot\mid s)=Q^{(i)}(L_i\in\cdot\mid S=s)$ on $\mathcal S^{\mathrm{sa}}_i$ and $\lambda\delta_{x_{\mathrm{in}}}+(1-\lambda)\delta_{x_{\mathrm{out}}}$ on $U_i$; the standalone deployment that draws $S$ from $Q^{(i)}_S$ then has record law $Q^{(i)}$, and the composition gives $c_B+\lambda P_S(U_i)$ for $\lambda\in[0,1]$. Since $P_S>0$ on $U_i$, the interval is a singleton if and only if $U_i=\emptyset$, which is (Supp$_i$).
(c) A standalone deployment that draws $S$ from $Q^{(i)}_S$ and $L_i$ from $Q^{(i)}(L_i\in\cdot\mid S)$, paired with a composition that produces $S$ with law $P_S$ and draws $L_i$ from $F'_i(\cdot\mid S)$, has evidence $Q^{(i)}$ and conditional laws $F'_i$. In a deployment, different kernels at the same logged $S$ arise when an unlogged input differs between the two deployments.

\begin{remark}[Oversight cues]\label{rem:p3-eval}
Let the composition be deployment and the standalone deployment be testing, with one component, the agent, and the cue $Z$ as $S$. When $Z$ is recorded and the agent's kernel reads only $Z$ and private randomness, Remark~\ref{rem:p3-part-iv}(b) gives the interval $[c_B,c_B+P_{\mathrm{dep}}(U)]$ for a deployment event, where $U$ holds the cue values never seen in testing.
\end{remark}

\section{Verification Index}\label{app:checks}

Scripts supporting Sections~\ref{sec:framework} and~\ref{sec:results} evaluate finite instances using exact rational arithmetic (\texttt{fractions.Fraction}); seeded Monte Carlo cases, logarithms, and dynamic programs use floating point as disclosed in their headers. These calculations verify finite instances and counterexamples, whereas Appendices~\ref{app:formal} to~\ref{app:p3-proof} prove the general theorems. The companion file \texttt{checks/README.md} provides the full script-to-claim map, execution commands, and worked examples.

\section{AI Use Statement}
\label{app:ai-use}

This disclosure supplements the Ethical Considerations.

The authors used AI coding and writing assistants to support implementation, analysis tooling, proof checking, figure preparation, and manuscript drafting and revision. The reported empirical results are computed by evaluation code over logged runs and saved experimental artifacts. The authors take full responsibility for the methods, interpretation, and final content of the paper.

\end{document}